\documentclass{article}

\usepackage[preprint]{neurips_2026}

\usepackage[T1]{fontenc}
\usepackage{microtype}
\usepackage{amsmath,amssymb,amsthm}
\usepackage{graphicx}
\usepackage{xcolor}
\usepackage[hidelinks,hypertexnames=false]{hyperref}
\usepackage{tikz}
\usetikzlibrary{positioning}
\usepackage{algorithm}
\usepackage{algorithmic}
\usepackage{balance}
\usepackage{mathtools,bm}
\usepackage{booktabs}
\usepackage{multirow}
\usepackage{enumitem}
\usepackage{float}

\newcommand{\Description}[1]{}

\newcommand{\A}{\mathcal{A}}                 
\newcommand{\St}{\mathcal{S}}                
\newcommand{\Buf}{\mathcal{B}}               
\newcommand{\pgood}{P_{\mathrm{good}}}       
\newcommand{\pcurr}{P_{\mathrm{curr}}}       
\newcommand{\E}{\mathbb{E}}
\newcommand{\KL}{D_{\mathrm{KL}}}
\newcommand{\TV}{D_{\mathrm{TV}}}
\newcommand{\dpi}{d^{\pi}}                    
\newcommand{\rocc}{\rho^{A}}                 
\newcommand{\Adv}{A^{\pi}}                   
\newcommand{\rew}{r}                         
\newcommand{\srew}{\tilde r}                 
\newcommand{\bonus}{b}                       
\newcommand{\dratio}{\delta}                 
\newcommand{\sdratio}{\tilde\delta}          
\newcommand{\ssc}{\beta}                     
\newcommand{\dec}{\lambda_{e}}               
\newcommand{\Edec}{E_{\mathrm{decay}}}       
\newcommand{\rstd}{m_{|r|}}                   
\newcommand{\Nbuf}{N_{\mathrm{buf}}}         
\newcommand{\Kep}{K_{\mathrm{ep}}}           
\newcommand{\pot}{\phi}                      
\newcommand{\Rmax}{R_{\max}}
\DeclareRobustCommand{\reviewblue}[1]{#1}
\newcommand{\Align}{\Phi}                    
\DeclareMathOperator*{\sign}{sign}
\DeclareMathOperator*{\argmax}{arg\,max}
\DeclareMathOperator{\clip}{clip}
\DeclareMathOperator{\Var}{Var}
\DeclareMathOperator{\Cov}{Cov}

\theoremstyle{plain}
\newtheorem{theorem}{Theorem}[section]
\newtheorem{lemma}[theorem]{Lemma}
\newtheorem{proposition}[theorem]{Proposition}
\newtheorem{corollary}[theorem]{Corollary}
\theoremstyle{definition}
\newtheorem{definition}[theorem]{Definition}
\newtheorem{assumption}[theorem]{Assumption}
\theoremstyle{remark}
\newtheorem{remark}[theorem]{Remark}

\title{Optimistic Policy Regularization: Elite-Statistic Reward Shaping}

\author{Mai Pham \qquad Vikrant Vaze \qquad Peter Chin\\
Thayer School of Engineering, Dartmouth College\\
\texttt{mai.p.pham.th@dartmouth.edu}}
\date{}
\hypersetup{
  pdftitle={Optimistic Policy Regularization: Elite-Statistic Reward Shaping},
  pdfauthor={Mai Pham, Vikrant Vaze, Peter Chin}
}

\newcommand{\BibTeX}{\rm B\kern-.05em{\sc i\kern-.025em b}\kern-.08em\TeX}

\begin{document}
\maketitle


\begin{abstract}
On-policy agents discard trajectories after each update, including rare high-return episodes. We propose Optimistic Policy Regularization (OPR), which turns those episodes into a closed-form reward bonus: we estimate an elite action prior from top-quartile episodes and add a \reviewblue{bounded transform of the log-ratio} of the prior to the policy's action distribution. OPR needs no auxiliary network, value gate, off-policy correction, or adversarial loop. It is a one-line modification of Proximal Policy Optimization (PPO)'s reward; yet, on 49 Atari games at 10M steps, $5\times$ fewer agent decisions than the 200M-frame baselines, it reaches a 148.4\% median human-normalized score (HNS; 196.9\% interquartile mean): higher scores than published PPO on $36/49$ games ($p{=}0.0014$) and a higher aggregate median than self-imitation learning, which falls well below OPR when frames and backbone are matched. We pair the method with a theory of when it works. The underlying log-ratio has an exact forward-KL interpretation; its pre-advantage placement propagates the signal through Generalized Advantage Estimation; and, under an idealized local update, its first-order effect depends on a prior-advantage alignment whose sign indicates help or harm. Since alignment varies across tasks and runs, we also study an online gating extension. On a ten-game set, it outperforms the backbone by $+17.4$ median HNS on average across seeds. Controlled experiments test temporal propagation and the tradeoffs of state conditioning and gating, supporting OPR as a simple regularizer with an analysis of where to shape.
\end{abstract}

\smallskip
\noindent\textbf{Keywords:} RL, reward shaping, self-imitation, policy regularization
\medskip


\section{Introduction}
On-policy actor-critic methods such as Proximal Policy Optimization (PPO)~\citep{schulman2017ppo} collect a batch of rollouts, take a few gradient steps, and discard the data. Occasionally, a rollout contains a rare, high-return episode - a lucky exploit or a well-executed skill - whose behavior the agent has no direct mechanism to reinforce beyond the noisy advantage signal it produces, and that signal is especially weak in the many environments where reward is sparse. Existing remedies re-use good experience, but each adds machinery: experience replay~\citep{mnih2015dqn,schaul2016per} requires off-policy correction, self-imitation learning (SIL)~\citep{oh2018sil} adds an auxiliary loss and a value gate, and curiosity bonuses~\citep{pathak2017icm,burda2019rnd} add auxiliary predictive models. Re-using an agent's own elite experience remains an active direction, from recent self-imitation variants for agentic reinforcement learning (RL)~\citep{qin2025spear} to closed-form, policy-ratio reward signals in KL-regularized fine-tuning~\citep{cui2025prime}, which we connect to reward shaping in \S\ref{sec:kl-interp}.

We propose \emph{Optimistic Policy Regularization} (OPR), which re-uses good episodes by replaying their \emph{statistics} rather than their transitions. From a buffer of the agent's own highest-return episodes, we estimate an \emph{elite prior} $\pgood$ - the distribution of actions those episodes take, conditioned on \reviewblue{$K$ state clusters (\S\ref{sec:prior})} - and compare it with $\pcurr$, the policy's own recent action frequencies. For simplicity, all main experiments use the same state-agnostic setting $K=1$, without per-game tuning. The per-step reward bonus is a bounded transform of the log-ratio of the two, positive for actions the elite episodes favor. OPR needs no auxiliary network, no value gate, and no off-policy correction. It is simply a one-line modification of the reward.

The minimality is principled: before bounding, the log ratio is, in expectation, exactly the negative \reviewblue{Kullback--Leibler (KL)} divergence from the policy's action distribution to the elite prior; the bounded transform preserves its ordering while controlling its scale. Its pre-advantage placement allows ordinary Generalized Advantage Estimation (GAE) to broadcast the elite signal densely, while post-hoc imitation losses act sparsely. On a $49$-game Atari benchmark, OPR reaches a $148.4\%$ median human-normalized score (HNS; $196.9\%$ interquartile mean, IQM; $32/49$ games above human) from $10$M agent steps - $5\times$ fewer decisions than the $200$M-frame ($50$M-step) baselines it is compared against, at a similar raw-frame budget (\S\ref{sec:setup}; full per-game results and budget accounting in the supplement).

While 49 games provide breadth, we use a ten-game set for analysis with baseline reimplementations, exploratory extensions, and ablations. The results show that OPR outperforms the benchmarks on average, although the performance has high variation across runs. We characterize \emph{when OPR works}: we compute a scalar \emph{alignment} $\Align(g)$ from a single checkpoint, with no additional training, whose sign predicts whether shaping will help or hurt in a given game. 
As an additional result, under the idealized KL-proximal assumptions, gating the bonus on positive local alignment is no worse than the same raw update to first order; the checkpoint-based oracle gate stays above the same-seed backbone at every available seed (\S\ref{sec:gated}). The theory includes $K>1$ for state-aliased games such as Pong, where a coarse prior aggregates away relevant state information. Larger $K$ trades bias against variance rather than guaranteeing improvement; robust \reviewblue{state-cluster construction} remains future work. Full runs and protocols are in the supplement.

\paragraph{Contributions.}
\begin{itemize}
\item \textbf{Method and results.} OPR: a closed-form density-ratio reward bonus from an elite-episode buffer (Algorithm~\ref{alg:opr}), tested on $49$ Atari games with one shared hyperparameter setting and no per-game tuning. At matched frames and backbone, SIL falls below OPR. Controlled experiments show how pre-GAE injection (\S\ref{sec:synthetic}) propagates an informative signal to earlier decisions.
\item \textbf{Theory.} The underlying log-ratio connects OPR to forward-KL regularization, discriminator-free adversarial imitation, regularized MDPs, and advantage-weighted regression (\S\ref{sec:theory}). Pre-advantage injection propagates densely through GAE, whereas post-hoc imitation updates are sparse (\S\ref{sec:th-prop}); and the alignment $\tilde\Align$ gives the first-order effect of adding shaping to an otherwise identical update, with an impossibility result for state-aliased games (\S\ref{sec:predictive}).
\item \textbf{Prediction and control.} From a single checkpoint, $\Align$ has the correct sign on \reviewblue{all seven meaningful-effect games after the disclosed Jamesbond exclusion} (\S\ref{sec:phi-exp}). Under the idealized update, an oracle sign gate has a nonnegative first-order incremental effect; online gating gains a seed-averaged $+17.4$ median HNS over the backbone (\S\ref{sec:gated}).
\end{itemize}.

\section{Preliminaries}
\reviewblue{A discounted Markov decision process} $(\St,\A,P,\rew,\gamma,\rho_0)$ \reviewblue{(MDP) has} states $\St$, a finite action set $\A$, a transition kernel $P(s'\mid s,a)$, a bounded reward $\rew:\St\times\A\to\mathbb{R}$, discount $\gamma\in(0,1)$, and initial distribution $\rho_0$. A stationary policy $\pi_\theta(a\mid s)$ induces the value function $V^\pi(s)=\E_\pi[\sum_{k\ge0}\gamma^k \rew_{t+k}\mid s_t=s]$, the advantage $\Adv(s,a)=Q^\pi(s,a)-V^\pi(s)$, the discounted state-visitation distribution $\dpi$, and the discounted action marginal $\rocc_\pi(a)=(1-\gamma)\sum_{t\ge0}\gamma^t\Pr_\pi(a_t=a)$, summing out state. The objective is $J(\pi)=\E_{s_0\sim\rho_0}[V^\pi(s_0)]$. The supplementary table collects the notation.

\paragraph{PPO and GAE.}
PPO~\citep{schulman2017ppo} collects on-policy rollouts of horizon $T$ in $N$ parallel environments and updates $\pi_\theta$ using the clipped surrogate objective with an entropy bonus and a value-function loss for the critic $V$. Advantages are formed by GAE~\citep{schulman2016gae},
\begin{equation}\label{eq:gae}
\hat A_t=\sum_{l=0}^{T-t-1}(\gamma\lambda)^l\,\delta^{\mathrm{TD}}_{t+l},
\quad
\delta^{\mathrm{TD}}_t=\rew_t+\gamma V(s_{t+1})-V(s_t),
\end{equation}
with GAE parameter $\lambda\in[0,1]$. The reward enters through the \reviewblue{temporal-difference (TD)} residual and propagates into \emph{every} preceding advantage with weight $(\gamma\lambda)^l$. Modifying reward \emph{before} GAE inherits this propagation; adding a loss \emph{after} GAE does not.

\paragraph{Self-imitation learning.}
SIL~\citep{oh2018sil} stores past transitions $(s,a,R)$, \reviewblue{with return-to-go $R$}, \reviewblue{and uses the loss} $\mathcal{L}_{\mathrm{SIL}}=-\E[\log\pi(a\mid s)\,(R-V(s))_+]$ after the PPO update; its value gate is sensitive to critic error (\S\ref{sec:theory}; supplementary value-gate analysis). OPR instead distills elite episodes into an action distribution and emits a \emph{reward} before GAE, with the alignment diagnostic developed in \S\ref{sec:predictive}.

\paragraph{Evaluation protocol.}
We report $\mathrm{HNS}(g)=\frac{\text{score}(g)-\text{random}(g)}{\text{human}(g)-\text{random}(g)}\times100\%$ and take its median across games as the primary aggregate metric, following the robustness recommendations of \citet{agarwal2021rliable}; the median is insensitive to games with an extreme HNS (e.g., Jamesbond, whose human baseline is only $302.8$). Unless noted, every run uses $10$M agent steps.

\section{Optimistic Policy Regularization}\label{sec:method}

\begin{figure}[t]
\centering
\begin{tikzpicture}
  \node[anchor=south west, inner sep=0] (img)
    {\includegraphics[width=0.45\columnwidth]{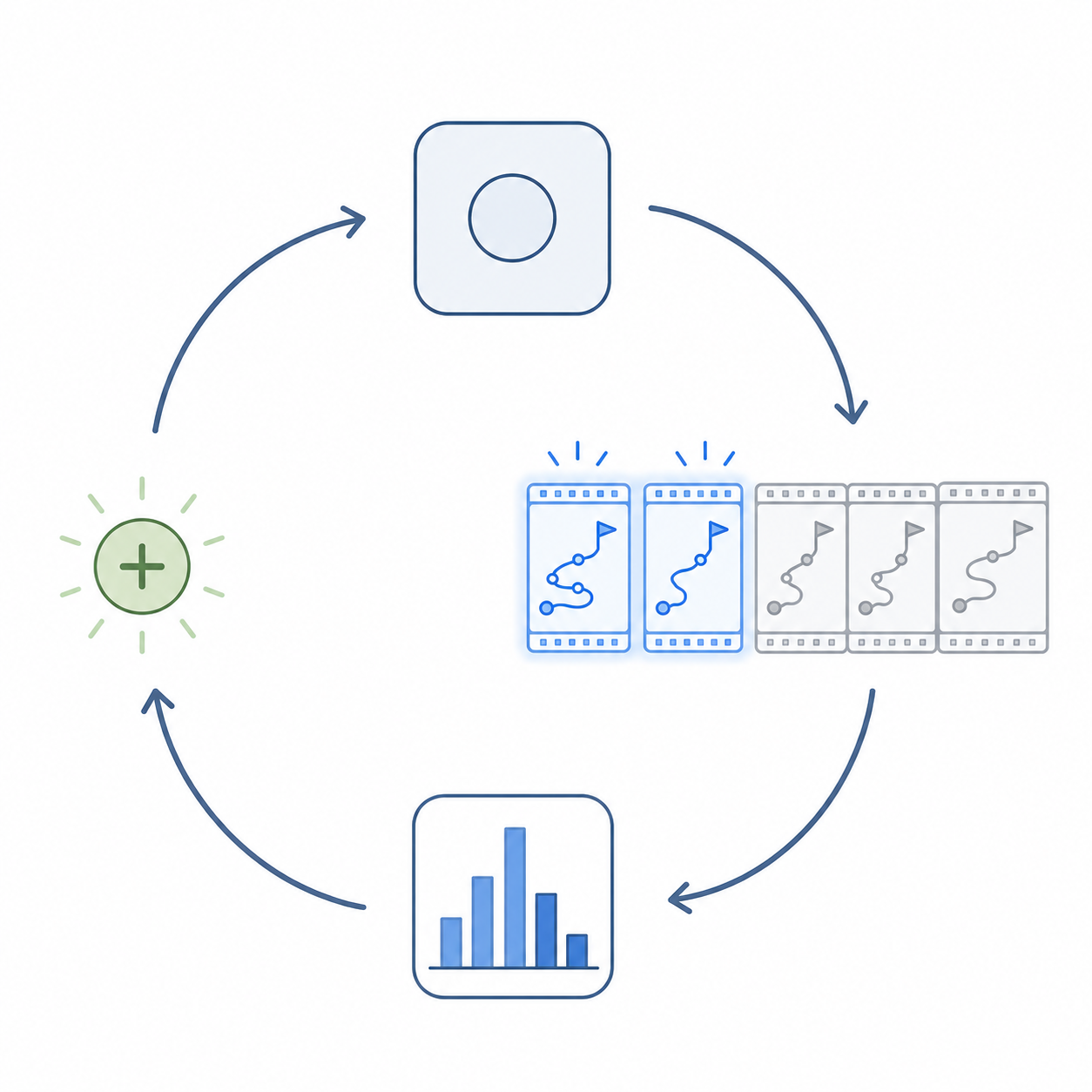}};
  \begin{scope}[x={(img.south east)}, y={(img.north west)}]
    \tikzset{lbl/.style={font=\tiny\itshape, text=black!80,
             fill=white, fill opacity=0.85, text opacity=1,
             inner sep=1.2pt, rounded corners=1pt, align=center}}
    \node[lbl] at (0.47,0.93) {policy $\pi_\theta$};
    \node[lbl] at (0.72,0.36) {best episodes};
    \node[lbl] at (0.53,0.04) {action preference $\pgood$};
    \node[lbl] at (0.135,0.33) {reward bonus for\\ matching $\pgood$};
  \end{scope}
\end{tikzpicture}
\caption{OPR is optimistic self-reinforcement. The agent distills its highest-return episodes' action statistics into an elite prior $\pgood$ and adds a reward bonus for matching it, pulling the policy toward its best observed behavior.}
\label{fig:concept}
\end{figure}

OPR augments PPO with a scalar reward bonus derived from the agent's own high-return episodes. The method has three parts -- a good-episode buffer (\S\ref{sec:buffer}), an elite action prior estimated from it (\S\ref{sec:prior}), and a density-ratio bonus injected before GAE (\S\ref{sec:shaping}). Algorithm~\ref{alg:opr} states the full procedure; hyperparameters and preprocessing are in the supplement.

\subsection{Good-Episode Buffer}
\label{sec:buffer}

We score each completed episode $\tau$ by its undiscounted return $R(\tau)=\sum_t \rew_t$. After each collection batch, we compute its $75$th-percentile episode return $R^\star$ and insert episodes with $R(\tau)\ge R^\star$ into a \reviewblue{first-in, first-out (FIFO)} buffer $\Buf$ of capacity $1000$. Retaining the top quartile keeps diverse high-return behavior, while FIFO forgetting tracks the improving policy. Shaping is disabled until $\Buf$ holds more than $20$ episodes, allowing data collection before the prior is used.

\subsection{Elite Action Prior}
\label{sec:prior}

We form an \emph{elite prior} over actions from only the $\Kep$ highest-return buffered episodes ($\Kep{=}10$ throughout), concentrating on the best observed behavior. \reviewblue{For state clusters $z:\St\to\{1,\dots,K\}$, let $N_{\mathrm{good}}(a\mid k)$ count action $a$ in $k$, with $N_{\mathrm{good}}(k)=\sum_a N_{\mathrm{good}}(a\mid k)$. Laplace smoothing gives}
\begin{equation}\label{eq:pgood}
\pgood(a\mid k)=\frac{N_{\mathrm{good}}(a\mid k)+\eta}{N_{\mathrm{good}}(k)+\eta\,|\A|},\qquad \eta=0.1,
\end{equation}
with a current prior $\pcurr(a\mid k)$ initialized from the first batch and updated as a Laplace-smoothed exponential moving average \reviewblue{(EMA)} (rate $\alpha=0.01$) of rollout-batch action counts. \textbf{Our deployed setting is $K{=}1$}, so $\pgood(a\mid k)$ reduces to $\pgood(a)$. This is a bias-variance choice: \S\ref{sec:th-l1} bounds the estimation cost of larger $K$, while \S\ref{sec:predictive} characterizes the bias from state aliasing. The bonus is zero where $\pcurr=\pgood$; the conditional notation below also covers $K>1$.

\subsection{Density-Ratio Reward Shaping}
\label{sec:shaping}

For $k_t=z(s_t)$, the core quantity is the log-ratio of the two conditional priors at the taken action:
\begin{equation}\label{eq:delta}
\dratio(a_t\mid k_t)=\log\pgood(a_t\mid k_t)-\log\pcurr(a_t\mid k_t).
\end{equation}
$\dratio(a_t\mid k_t)$ is positive when $a_t$ is over-represented among elite episodes relative to the policy's own frequency in cluster $k_t$. We pass $\dratio$ through a smooth squashing process as follows.
\begin{equation}\label{eq:sdelta}
\begin{aligned}
\sdratio(a_t\mid k_t)&=2\tanh\!\big(\dratio(a_t\mid k_t)/2\big)\\
&=\frac{2\,(\pgood-\pcurr)}{\pgood+\pcurr}(a_t\mid k_t)\;\in(-2,2).
\end{aligned}
\end{equation}
This is not merely an ad-hoc stabilizer: it is a scaled \emph{centered Bayes posterior} for distinguishing elite from current actions under a balanced mixture -- \reviewblue{twice the centered least-squares discriminator in closed form} --so its bound is intrinsic rather than imposed by clipping. (proof in the supplement). \S\ref{sec:kl-interp} gives its exact expectation.
We scale it to the reward's own units as follows.
\begin{equation}\label{eq:bonus}
\bonus_t=\ssc\,\dec\,\rstd\,\sdratio(a_t \mid k_t)
\qquad
\dec=\max\!\big(0,\,1-e/\Edec\big)
\end{equation}
Here $\ssc>0$ is the shaping scale, $\rstd=\max(\frac1B\sum_{t=1}^B |r_t|,10^{-4})$ the implementation's batch reward scale \reviewblue{for $B$ transitions}, and $\dec$ a linear decay factor over training epoch $e$. The \emph{shaped reward} is $\srew_t=\rew_t+\bonus_t$. Because $|\sdratio|\le2$, the perturbation is uniformly bounded $|\bonus_t|\le 2\ssc\,\dec\,\rstd$; decay makes this instantaneous perturbation vanish without undoing earlier shaped updates (\S\ref{sec:theory}). Adding $\bonus_t$ before GAE~\eqref{eq:gae} propagates it into preceding advantage estimates (\S\ref{sec:th-prop}); both derivations are in the supplement.

\begin{algorithm}[t]
\caption{Optimistic Policy Regularization (one epoch)}
\label{alg:opr}
\begin{algorithmic}[1]
\STATE \textbf{Input:} policy $\pi_\theta$, critic $V$, buffer $\Buf$, EMA $\pcurr(\cdot\mid k)$, epoch $e$
\STATE Collect rollouts $\{(s_t,a_t,\rew_t)\}$ with $\pi_\theta$ ($N$ envs, $T$ steps)
\STATE Update $\pcurr(\cdot\mid k)$ (EMA of per-cluster batch action counts)
\IF{a prior from $|\Buf|>20$ earlier episodes is available}
  \STATE Compute $\sdratio(a_t\mid z(s_t))$ for all transitions \hfill(Eqs.~\ref{eq:delta}--\ref{eq:sdelta})
  \STATE $\dec\leftarrow\max(0,1-e/\Edec)$
  \FORALL{transitions $t$}
    \STATE $\bonus_t\leftarrow \ssc\,\dec\,\rstd\,\sdratio(a_t\mid z(s_t))$ \hfill(Eq.~\ref{eq:bonus})
    \STATE $\srew_t\leftarrow \rew_t+\bonus_t$
  \ENDFOR
\ELSE
  \STATE $\srew_t\leftarrow \rew_t$
\ENDIF
\STATE $\hat A_t\leftarrow\text{GAE}(\srew,V)$ \hfill(Eq.~\ref{eq:gae}; \emph{pre-advantage})
\STATE PPO update of $\pi_\theta,V$ on $\hat A_t$
\STATE $R^\star\!\leftarrow\!75$th-pct.\ episode return; add episodes with $R\!\ge\!R^\star$ to $\Buf$
\STATE Recompute $\pgood$ from the updated $\Buf$ for the next epoch
\end{algorithmic}
\end{algorithm}

\subsection{Variants and Ablations}
\label{sec:extensions}

The theory defines the prior for general $K$. Our $K{>}1$ experiments use batch $k$-means over frozen random-encoder features; larger $K$ can reduce aliasing but increases estimation error (\S\ref{sec:th-l1}). How best to form the \reviewblue{state clusters} remains future work. Orthogonally, return-quantile bins vary which episodes define the prior, while the gating extension uses a running estimate of bounded-score/raw-advantage alignment to turn shaping on or off (\S\ref{sec:predictive}). Proofs, a controlled study, and implementation details are in the supplement.

\section{Theoretical Analysis}\label{sec:theory}

The conceptual core is the \emph{dense propagation} lemma (\S\ref{sec:th-prop}), which distinguishes reward-channel shaping from post-hoc imitation losses, and the \emph{alignment} diagnostic (\S\ref{sec:predictive}), which characterizes when OPR helps. Full proofs are in the self-contained supplement.

\subsection{OPR is a forward-KL regularizer}
\label{sec:kl-interp}

\begin{proposition}[Raw and bounded score identities]\label{prop:score-identities}
For positive action distributions $p$ and $q$, let $\delta(a)=\log(p(a)/q(a))$ and $\tilde\delta(a)=2(p(a)-q(a))/(p(a)+q(a))$. Then
\begin{align}
\E_{a\sim q}[\delta(a)]&=-\KL(q\|p),\label{eq:raw-kl}\\
\E_{a\sim q}[\tilde\delta(a)]&=-\sum_a\frac{(p(a)-q(a))^2}{p(a)+q(a)}.\label{eq:bounded-tri}
\end{align}
Both expressions are non-positive and vanish iff $p=q$.
\end{proposition}
\begin{proof}
The first identity defines KL. For the second, sum $2q(p-q)/(p+q)=(p-q)-(p-q)^2/(p+q)$ over actions, using $\sum_a(p(a)-q(a))=0$.
\end{proof}
The raw score induces a forward-KL objective from the policy's action distribution to the elite prior. The bounded score instead induces triangular discrimination, with the same unique distributional optimum and a bounded reward.

\begin{lemma}[Closed-form discriminator]\label{lem:discriminator}
Under a balanced mixture of $p$ and $q$, the Bayes posterior that an action came from $p$ is $D^*(a)=p(a)/(p(a)+q(a))$, and
\begin{equation}
\delta(a)=\log\frac{D^*(a)}{1-D^*(a)},\qquad \tilde\delta(a)=2(2D^*(a)-1).
\end{equation}
\end{lemma}
\begin{proof}
Bayes' rule with equal class priors gives $D^*=p/(p+q)$; substitution gives both identities.
\end{proof}
Thus the raw ratio has the log-odds form used by \reviewblue{generative adversarial imitation learning (GAIL) and adversarial inverse reinforcement learning (AIRL)} \citep{ho2016gail,fu2018airl}, after projection to per-cluster action distributions, while the bounded score is twice the centered least-squares discriminator. OPR obtains both in closed form, with no discriminator network or min--max loop. \reviewblue{This action-level connection does not equate OPR with state--action GAIL/AIRL.} The same density ratio connects OPR to regularized MDPs \citep{geist2019regularized,vieillard2020munchausen}, while elite filtering is a hard-threshold analogue of advantage weighting \citep{peters2007rwr,peng2019awr,abdolmaleki2018mpo}.

\begin{proposition}[Uniqueness within a separable family]\label{prop:log-unique}
Suppose $b(a)=g(p(a))-g(q(a))$, where measurable $g:(0,1]\to\mathbb R$ satisfies information additivity, $g(xy)=g(x)+g(y)$. Then $g(x)=c\log x$ for some constant $c$.
\end{proposition}
\begin{proof}[Proof sketch]
Set $h(u)=g(e^{-u})$. The measurable Cauchy equation $h(u+v)=h(u)+h(v)$ forces $h$ to be linear.
\end{proof}
Thus, among separable information-additive bonuses, the log-ratio form is forced up to scale. The same closed-form, policy-ratio reward appears in KL-regularized fine-tuning of language models~\citep{cui2025prime}; OPR derives it instead from an agent's own elite episodes and injects it into the RL reward channel rather than a supervised objective. State re-enters through $k=z(s)$: $\pcurr$ and $\pgood$ are per-cluster action distributions, so $K$ is a state-conditioning dial.

\begin{proposition}[Raw-score KL direction]\label{prop:raw-direction}
At a reference action distribution $q$, the score $\log(p/q)$ is the negative gradient at $x=q$ of $x\mapsto\KL(x\|p)$, up to an action-independent constant that vanishes along directions tangent to the probability simplex.
\end{proposition}
\begin{proof}
The gradient is coordinatewise
\[
\left.\nabla_x\KL(x\|p)\right|_{x=q}=\log(q/p)+1.
\]
Negating it leaves $\log(p/q)$ plus a constant one vector.
\end{proof}
This is the mirror-descent connection for the idealized raw-score objective. The bounded transform, lagged EMA reference, and clipped neural update require the separate analysis below.

\subsection{Dense advantage propagation}
\label{sec:th-prop}

\begin{lemma}[Exact shaped-advantage decomposition]\label{lem:gae-propagation}
Holding the critic fixed, within an uninterrupted rollout segment,
\begin{equation}\label{eq:decomp}
\hat A_t^{\mathrm{OPR}}=\hat A_t+\sum_{l=0}^{T-t-1}(\gamma\lambda)^l \bonus_{t+l}.
\end{equation}
\end{lemma}
\begin{proof}
The shaped TD residual is the raw residual plus $\bonus_t$; substitution into GAE gives \eqref{eq:decomp}.
\end{proof}
\begin{corollary}[Dense propagation]\label{cor:dense-propagation}
A bonus $\bonus_\tau$ enters every earlier advantage in the same segment with weight $(\gamma\lambda)^{\tau-t}$; its total downstream weight approaches $1/(1-\gamma\lambda)$ as $\tau$ grows.
\end{corollary}
\begin{proof}
Sum the geometric weights in \eqref{eq:decomp} over $t\le\tau$.
\end{proof}
OPR shapes all $T\!\times\!N$ rollout transitions per epoch and propagates each through GAE; SIL applies auxiliary gradients to $n_{\rm steps}\!\times\!|\mathrm{batch}|$ sampled replay transitions. This propagates the \emph{new reward}, not the credit used to select an elite episode. Section~\ref{sec:synthetic} isolates temporal placement; \S\ref{sec:controlled} compares against corrected SIL. Implementation details are in the supplement.

\subsection{Statistical cost, robustness, and decay}
\label{sec:th-l1}

\begin{theorem}[Prior estimation]\label{thm:prior-estimation}
If $N_{\rm elite}$ actions are i.i.d. draws from the elite marginal, then the unsmoothed empirical prior satisfies
\begin{equation}
\E\|\hat\pgood-\pgood\|_1\le\sqrt{\frac{|\A|-1}{N_{\rm elite}}}.
\end{equation}
With $K$ comparably occupied clusters, the per-cluster rate increases by $O(\sqrt K)$.
\end{theorem}
\begin{proof}[Proof sketch]
Jensen's inequality and Cauchy--Schwarz give the bound; using $N_{\rm elite}/K$ samples per comparably occupied cluster gives the second statement.
\end{proof}
Trajectory dependence and uneven occupancy reduce the effective sample size. Thus larger $K$ can reduce state-aliasing bias but increases estimation error; the theorem does not make $K{=}1$ optimal.

\reviewblue{For illustration, with} $N_{\rm elite}=3\times10^5$ and $|\A|=18$, Theorem~\ref{thm:prior-estimation} gives $0.0075\approx0.008$ for $K{=}1$ and a four-times-larger bound for $K{=}16$. This quantifies estimation error, not return.

\begin{corollary}[Marginalization contracts drift]\label{cor:tv-contraction}
For any state--action distributions $\mu$ and $\nu$ and their action marginals $\mu_A,\nu_A$, $\TV(\mu_A,\nu_A)\le\TV(\mu,\nu)$.
\end{corollary}
\begin{proof}
Apply total-variation data processing to $(s,a)\mapsto a$.
\end{proof}

\begin{lemma}[SIL value-gate error]\label{lem:sil-gate}
The SIL decision $\mathbf 1\{R>\hat V(s)\}$ differs from the exact-critic decision $\mathbf 1\{R>V^\pi(s)\}$ exactly when $R$ lies between $\hat V(s)$ and $V^\pi(s)$ (apart from equality). Such an error requires $|\hat V(s)-V^\pi(s)|>|R-V^\pi(s)|$; the reverse implication need not hold.
\end{lemma}
\begin{proof}
The two indicators disagree exactly when $(R-\hat V(s))(R-V^\pi(s))<0$; this is precisely the between-ness condition and implies the stated magnitude inequality.
\end{proof}
Ungated OPR has no value-based inclusion gate, so this particular error does not apply, although prior estimation, state aliasing, and episode selection introduce their own errors.

\begin{proposition}[Effect of decay]\label{prop:decay}
Because $|\bonus_t|\le2\ssc\dec\rstd$, the instantaneous reward perturbation is zero after the decay completes. For constant shaping and a fixed normalization scale, the true-return loss of a shaped-reward optimum is at most $4\ssc\rstd/(1-\gamma)$. Decay does not undo earlier shaped updates or recover the parameters of an independently trained unshaped agent.
\end{proposition}
\begin{proof}[Proof sketch]
The first claim follows from $\dec=0$. For constant shaping, every policy's shaped and true returns differ by at most $\|\bonus\|_\infty/(1-\gamma)$; applying this bound to the true and shaped optima gives twice this quantity, and $\|\bonus\|_\infty\le2\ssc\rstd$.
\end{proof}
OPR is not potential-based, so the policy-invariance theorem of \citet{ng1999shaping} does not apply to constant shaping. Proposition~\ref{prop:decay} bounds the shift and identifies what decay removes.
\label{sec:th-sil}\label{sec:th-decay}

\subsection{Predictive per-game theory}
\label{sec:predictive}

\paragraph{\reviewblue{A scalar that predicts the effect.}} The original analysis defines the \emph{prior--advantage alignment}
\begin{equation}\label{eq:phi}
\Align(g)=\E_{s\sim\dpi}\Big[\textstyle\sum_a\pgood(a\mid k(s))\Adv(s,a)\Big].
\end{equation}
\begin{proposition}[Prior-directed checkpoint diagnostic]\label{prop:phi-diagnostic}
If an idealized OPR step satisfies $\pi_\beta(\cdot\mid s)=\pi(\cdot\mid s)+\beta\kappa[\pgood(\cdot\mid k(s))-\pi(\cdot\mid s)]+o(\beta)$ for $\kappa>0$, then
\begin{equation}
J(\pi_\beta)-J(\pi)=\frac{\beta\kappa}{1-\gamma}\Align(g)+o(\beta).
\end{equation}
\end{proposition}
\begin{proof}[Proof sketch]
Apply the Performance Difference Lemma \citep{kakade2002cpi}, insert the assumed direction and use $\sum_a\pi(a\mid s)\Adv(s,a)=0$; replacing $d^{\pi_\beta}$ by $\dpi$ changes only the higher-order term.
\end{proof}
Within this model, $\Align>0$, $\Align\approx0$, and $\Align<0$ define GAIN, FLAT, and HURT regimes. Moreover, $|\Align(g)|\le\|\Adv\|_\infty\E_s\|\pgood(\cdot\mid k(s))-\pi(\cdot\mid s)\|_1$, so a small prior--policy gap forces FLAT. The explicit direction assumption makes $\Align$ a checkpoint diagnostic, rather than a derived description of clipped PPO. We next remove that assumption by comparing shaped and unshaped KL-proximal updates from the same policy with the same raw score $h$. For $\pi_\epsilon^0\propto\pi e^{\epsilon h}$ and $\pi_\epsilon^b\propto\pi e^{\epsilon(h+\beta f)}$:
\begin{theorem}[First-order incremental shaping effect]\label{thm:local-gain}
In a finite discounted MDP, for a full-support policy and bounded scores,
\begin{align}
J(\pi_\epsilon^b)-J(\pi_\epsilon^0)&=\frac{\epsilon\beta}{1-\gamma}\widetilde\Align[f]+O(\epsilon^2),\label{eq:local-gain}\\
\widetilde\Align[f]&=\E_{s\sim\dpi,a\sim\pi}[f(s,a)\Adv(s,a)].
\end{align}
The common raw update $h$ cancels from the first-order difference.
\end{theorem}
\begin{proof}[Proof sketch]
The normalized exponential tilt satisfies $\pi_\epsilon^u=\pi\{1+\epsilon(u-\E_\pi u)\}+O(\epsilon^2)$. Subtracting the expansions for $u=h+\beta f$ and $u=h$ cancels $h$; the Performance Difference Lemma and centered advantage then give \eqref{eq:local-gain}.
\end{proof}
\begin{corollary}[Oracle local gate]\label{cor:local-gate}
Gating on $\widetilde\Align[f]>0$ replaces the leading coefficient by $\max\{\widetilde\Align[f],0\}\ge0$.
\end{corollary}
For the immediate OPR direction, $f=\sdratio$; pre-GAE injection replaces it by the expected geometrically propagated score. The checkpoint experiment estimates $\Align$, while the online covariance gate is a practical proxy for $\widetilde\Align$; the two are related but not identical (supplement). The theorem concerns this idealized local update, not the complete clipped neural-PPO training path.

\begin{corollary}[Gating with estimation error]\label{cor:selective-gate}
Let differentiable raw and shaped update paths $U_0(\eta)$ and $U_1(\eta)$ start at the same $\theta$, with differentiable $J$. Their local coefficient is
\begin{equation}
C=\nabla J(\theta)^\top[U_1'(0)-U_0'(0)].
\end{equation}
Suppose $\Pr(|\hat C-C|\le r)\ge1-\alpha$. On this event, the rule $g=\mathbf1\{\hat C>r\}$ activates only if $C>0$, and activates whenever $C>2r$. Every activated update therefore improves on the same raw step for sufficiently small positive $\eta$.
\end{corollary}
\begin{proof}[Proof sketch]
On the stated event, activation gives $C\ge\hat C-r>0$, while $C>2r$ gives $\hat C\ge C-r>r$. Differentiability gives $J(U_1(\eta))-J(U_0(\eta))=\eta C+o(\eta)$.
\end{proof}
For Theorem~\ref{thm:local-gain}, $C=\beta\widetilde\Align[f]/(1-\gamma)$. Thus the gate trades useful coverage against estimation error, provided it estimates the coefficient of the actual update. The supplement gives a valid held-out bounded-sample radius; \S\ref{sec:gated} tests empirical calibration and the effect of advantage normalization.

\begin{theorem}[\reviewblue{When $K{=}1$ is insufficient}]\label{thm:marginal-insufficiency}
There are finite MDPs in which an optimal and an arbitrarily suboptimal policy have the same action marginal. A $K{=}1$ elite/current ratio cannot distinguish them, while a state-conditioned prior can have strictly positive alignment at the suboptimal policy.
\end{theorem}
\begin{proof}[Proof sketch]
Let the initial state be uniform on $\{s_1,s_2\}$, with reward one for $(s_1,L)$ and $(s_2,R)$ and zero for the other actions. The optimal and reversed policies have returns one and zero but the same uniform action marginal; scaling the reward gives an arbitrary gap. Conditioning on the two states separates their action distributions and favors the positive-advantage action in each state.
\end{proof}
This is the state-aliasing regime: increasing $K$ can recover relevant information but also raises estimation error. Paddle games such as Pong motivate state conditioning; robust \reviewblue{state-cluster construction} remains future work.

\section{Experiments}
\label{sec:exp}

Our experiments address four research questions. \textbf{RQ1} (\emph{budgeted performance}): how does OPR compare with published methods at a fraction of their decision budget? \textbf{RQ2} (\emph{attribution and reliability}): with the backbone fixed, what does the shaping contribute, and how consistently across runs? \textbf{RQ3} (\emph{predictability}): is checkpoint alignment associated with where shaping helps? \textbf{RQ4} (\emph{exploitability}): can the algorithm use that signal? Ablations over the main design choices complete the study.

\subsection{Setup}
\label{sec:setup}

All agents use \reviewblue{PPO with the IMPALA residual-network encoder~\citep{espeholt2018impala}} (widths $[64,128,128]$, $512$-d hidden layer) shared by the policy and value heads, $8$ training / $10$ test environments, batch size $256$, and $100$ epochs at $10$M agent steps on \reviewblue{Arcade Learning Environment (ALE)} v5 with sticky actions. \textbf{Budget and protocol note.} The v5 frameskip ($4$) and skip-$4$ wrapper compose to an \emph{effective action-repeat of $16$}: $10$M agent steps $=160$M raw ALE frames. All our runs share this protocol; published baselines use $50$M decisions ($200$M frames, action-repeat $4$). Table~\ref{tab:broad} compares fewer decisions under coarser control, not matched sample efficiency. We use a pre-existing ten-game set for controlled analyses. Unless noted, OPR uses $\ssc=0.5$, a $75$th-percentile threshold, a $1000$-episode buffer (prior from its top $\Kep{=}10$ episodes), a $20$-episode warmup, $\eta=0.1$, and linear decay to zero over the run. \reviewblue{The supplement's Complete Results Inventory gives per-game scores and seed coverage.}

\subsection{Main Results}
\label{sec:main}

\reviewblue{We evaluate all $56$ Atari games; Table~\ref{tab:broad} compares the $49$ shared with published baselines.}

\begin{table}[t]
\centering
\small \setlength{\tabcolsep}{3.2pt}
\caption{Comparison on the $49$ games shared with published baselines. OPR uses one training seed; baselines are published scores from different systems. Steps are agent decisions. \reviewblue{DQN: deep Q-network; A2C: advantage actor--critic; ACPER: A2C with prioritized experience replay~\citep{oh2018sil}.} Median, interquartile mean (IQM), and optimality gap follow \citet{agarwal2021rliable}; $>$H counts games above human.}
\label{tab:broad}
\begin{tabular}{@{}lrrrrr@{}}
\toprule
Method & Steps & Median & IQM & Gap$\downarrow$ & $>$H \\
\midrule
DQN \citep{mnih2015dqn}     & 50M & 46.5\%  & 52.1\%  & .484 & 13/49 \\
ACPER                        & 50M & 46.5\%  & 64.7\%  & .451 & 18/49 \\
PPO \citep{schulman2017ppo} & 50M & 82.0\%  & 91.8\%  & .511 & 22/49 \\
A2C \citep{mnih2016a3c}     & 50M & 90.5\%  & 109.6\% & .369 & 24/49 \\
SIL \citep{oh2018sil}       & 50M & 140.0\% & 160.9\% & .213 & 29/49 \\
\midrule
OPR (ours)                   & 10M & \textbf{148.4\%} & \textbf{196.9\%} & \textbf{.171} & \textbf{32/49} \\
\bottomrule
\end{tabular}
\end{table}

OPR has the highest median HNS, IQM, and fraction above human, and the lowest optimality gap in Table~\ref{tab:broad}. Per-game sign tests compare score vectors, not matched training replications: OPR scores higher than published PPO on $36/49$ games ($p{=}0.0014$), ACPER on $38/49$ ($p{=}10^{-4}$), and DQN on $43/49$ ($p{<}10^{-4}$). Against SIL the sign test does not distinguish them ($26/49$, $p{=}0.78$; paired median difference $+1.0$ pp, $95\%$ bootstrap CI $[-57,+25]$). The supplement gives performance profiles. Different backbones, action repeats, and codebases leave attribution to the controlled set below.

\subsection{Fixed-Backbone Attribution and Reliability}
\label{sec:controlled}

To isolate the shaping's contribution, we hold the encoder, budget, and optimizer fixed and compare matched seeds. Two findings emerge. First, OPR's gains are large and structured---for example, Frostbite ($24{,}926$ vs $8{,}963$) and Qbert ($33{,}888$ vs $28{,}818$)---while Pong, Breakout, and Seaquest see little benefit at $K{=}1$. Second, ungated OPR's margin over the same-seed backbone spans $94$ pp ($+4.0/-43.2/+51.1$; mean $+4.0$): positive on average, but highly variable across runs.

\paragraph{Attribution: encoder, information, placement.} \reviewblue{We separate three sources of the headline result.} \emph{Encoder}: replacing the Nature-CNN with the IMPALA ResNet, both unshaped, moves median HNS from $56.4\%$ to $152.6\%$; this effect is held fixed thereafter. \emph{Information}: \reviewblue{marginal SIL, which restricts full (state-conditioned) SIL to OPR's $K{=}1$ action prior,} raises its median from $64.6\%$ to $98.5\%$ at the matched seed, so extra state information does not explain the comparison. \emph{Placement}: marginal SIL and OPR use the same encoder, budget, and elite signal, but introduce it through a post-hoc value-gated loss or a pre-GAE reward. \reviewblue{OPR outperforms marginal SIL on $9/10$ games at each matched seed (sign test $p{=}0.021$ each).} \reviewblue{This supports reward-channel placement, consistent with \S\ref{sec:th-prop}.}

\paragraph{Variability.} \reviewblue{The Intrinsic Curiosity Module (ICM)} has similar paired margins ($-10.6/-10.9$ pp), while elite-derived reward configurations vary by tens of points. This is consistent with policy $\to$ elite buffer $\to$ reward $\to$ returns $\to$ policy feedback in this setup, but does not establish causation. \reviewblue{See the supplementary Complete Results Inventory.}

\begin{table}[t]
\centering
\scriptsize \setlength{\tabcolsep}{2.2pt}
\caption{Architecture-controlled ten-game results \reviewblue{sharing the IMPALA encoder}: mean $\pm$ s.d.\ of median HNS across seeds, and worst- and best-seed medians. \reviewblue{IS: importance sampling.} $\dagger$: two seeds; $\ddagger$: one seed.}
\label{tab:selected}
\begin{tabular}{@{}lrrr@{}}
\toprule
Method & mean $\pm$ s.d. & worst & best \\
\midrule
\reviewblue{PPO backbone} & $142.7\pm17.0$ & 123.0 & 152.6 \\
\reviewblue{PPO + ICM}$^\dagger$ & $141.7\pm0.4$ & 141.4 & 142.0 \\
\midrule
OPR ($K{=}1$) & $146.6\pm33.6$ & 109.1 & 174.1 \\
OPR ($K{=}4$)$^\dagger$ & $125.9\pm39.0$ & 98.3 & 153.4 \\
marginal SIL$^\dagger$ & $99.2\pm1.0$ & 98.5 & 100.0 \\
full SIL$^\dagger$ & $64.0\pm0.8$ & 63.5 & 64.6 \\
\midrule
\reviewblue{gated OPR, checkpoint oracle}$^\dagger$ & $165.6\pm13.7$ & 155.9 & 175.3 \\
gated OPR, IS$^\ddagger$ & 151.2 & -- & -- \\
gated OPR, covariance & $\mathbf{160.1\pm8.2}$ & 151.1 & 167.3 \\
\bottomrule
\end{tabular}
\end{table}

\reviewblue{Ungated $K=1$ OPR and covariance-gated OPR exceed} the matched PPO backbone in average median HNS (Table~\ref{tab:selected}); gating also narrows the observed seed range.

\subsection{Controlled Mechanism Tests}
\label{sec:synthetic}

Controlled finite-MDP experiments test whether elite selection produces useful information, whether pre-GAE placement propagates it, and when state conditioning helps. Constructing the prior from top-quartile rollouts gives positive local alignment in $475/500$ random MDPs ($95.0\%$, Wilson $95\%$ CI $[92.7,96.6]\%$). \reviewblue{Additional horizon and elite-fraction checks give $94.5$--$96.5\%$: elite selection usually gives a favorable direction here.} Separately, across $5{,}000$ random six-state MDPs, the first-order residual scales as \reviewblue{$\epsilon^{1.9995}$ in step size $\epsilon$}, \reviewblue{matching Theorem~\ref{thm:local-gain} (protocols and finite-step checks in the supplement).}

\begin{figure*}[t]
\centering
\includegraphics[width=0.70\textwidth]{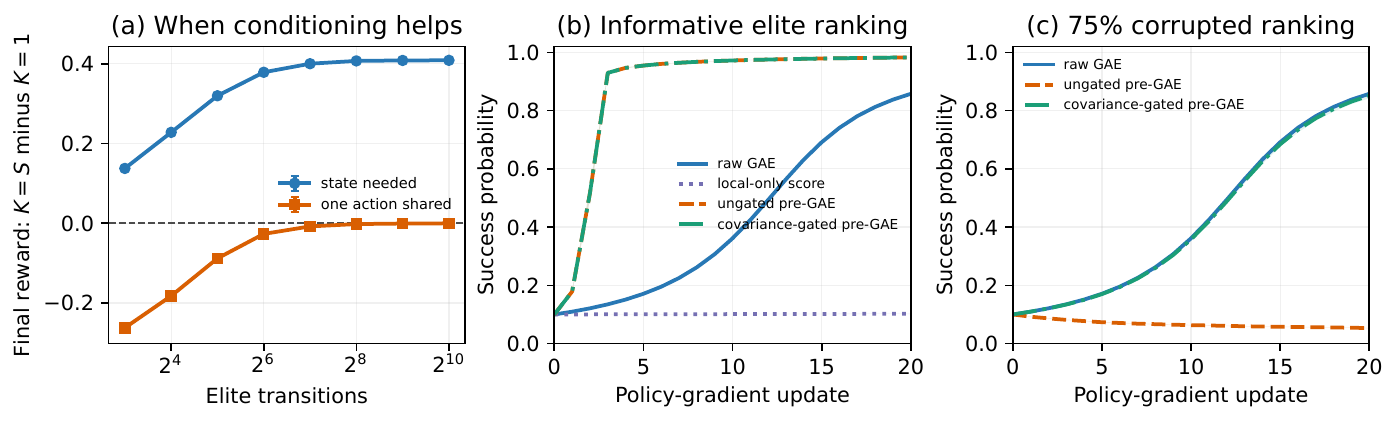}
\Description{Controlled experiments showing the bias-variance effect of state conditioning and learning in a delayed-decision task under informative and corrupted elite rankings.}
\caption{Controlled mechanism tests. (a) Conditioning helps when actions depend on state, but fragments scarce data when one action is shared ($400$ paired elite datasets). (b--c) Pre-GAE shaping propagates an informative delayed signal; covariance/EMA gating retains this gain and suppresses corrupted shaping ($200$ paired runs). Bands and bars are $95\%$ Monte Carlo intervals. The local-only control isolates placement, not SIL.}
\label{fig:toy-k-credit}
\end{figure*}

\paragraph{Temporal placement.} In a length-$10$ delayed-decision task, the first action determines terminal success and a later action-sequence signature. After $20$ sampled updates, \reviewblue{unshaped learning (raw GAE)} reaches success $0.858$, applying the score only at the initial transition reaches $0.102$, and pre-GAE shaping reaches $0.983$ (Figure~\ref{fig:toy-k-credit}). With $75\%$ corruption of the elite ranking, ungated shaping falls to $0.053$, while the \reviewblue{online covariance gate (EMA-smoothed)} reaches $0.851\pm0.009$ (paired difference from raw $-0.006\pm0.009$). \reviewblue{These paired controls isolate temporal propagation without PPO clipping or advantage normalization.}

\paragraph{\reviewblue{Action-relevant state clusters.}} Figure~\ref{fig:toy-k-credit}(a) shows both sides of the $K$ tradeoff: conditioning recovers state-dependent action preferences, whereas a shared action favors pooling scarce samples. A further four-context/four-action study varies label quality, merging, and over-splitting. With $2{,}048$ elite samples, the gain over $K=1$ is $0.570$, $0.379$, $0.180$, and $0$ under clean labels, $20\%$ noise, $40\%$ noise, and random labels, respectively. Action-preserving merges retain the useful distinction; merges of states requiring different actions lose it. \reviewblue{With a shared action and $32$ samples, unnecessary conditioning lowers reward by $0.022$, and over-splitting by $0.056$. These labels isolate useful cluster information; learning clusters remains future work.}

\paragraph{Randomized mechanism breadth.} \reviewblue{We randomize $1{,}000$ delayed-credit environments over horizon, $\gamma$, GAE $\lambda$, initial policy, elite fraction, shaping scale, and how reliably later actions identify the initial decision.} Table~\ref{tab:ranking-stress} compares clean and fully reversed elite rankings using $20$ exact population updates per environment. The immediate gate retains most of the average clean benefit and substantially reduces reversed-ranking damage. \reviewblue{Immediate and propagated alignment can disagree: an exact propagated-score gate eliminates harm in this family (corruption and estimation sweeps in the supplement).}

\begin{table}[t]
\centering
\small\setlength{\tabcolsep}{3pt}
\caption{Final-success difference from raw learning on $1{,}000$ randomized delayed-credit environments (paired mean $\pm$ $95\%$ CI half-width). Harm is the fraction with a negative difference. The gate uses \reviewblue{exact immediate covariance (without sampling error)}; reversed rankings stress-test corrupted elite selection.}
\label{tab:ranking-stress}
\begin{tabular}{@{}llrr@{}}
\toprule
Ranking & Shaping & $\Delta$ success & Harm \\
\midrule
Clean & Ungated & $+0.1296\pm0.0110$ & $22.8\%$ \\
 & Gated & $+0.1249\pm0.0110$ & $20.8\%$ \\
Reversed & Ungated & $-0.0569\pm0.0058$ & $77.2\%$ \\
 & Gated & $-0.0013\pm0.0007$ & $7.7\%$ \\
\bottomrule
\end{tabular}
\end{table}

\subsection{The Predictor Forecasts Per-Game Outcomes}
\label{sec:phi-exp}

\begin{figure}[t]
\centering
\includegraphics[width=0.67\columnwidth]{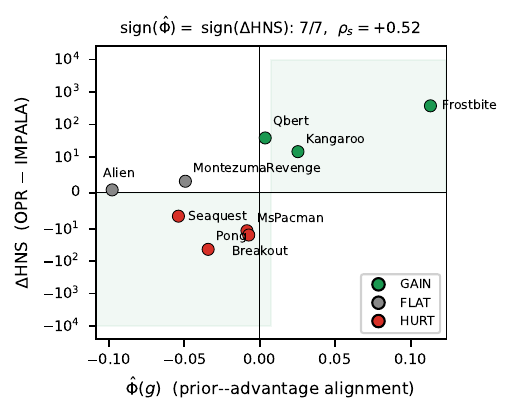}
\Description{Checkpoint alignment estimate versus the realized OPR-minus-IMPALA score gap on the ten-game controlled set.}
\caption{Checkpoint alignment estimate $\hat\Align(g)$ versus the realized OPR$-$IMPALA gap on the ten-game set. \reviewblue{Jamesbond is omitted because its small human-minus-random denominator makes the HNS gap extreme.} Shaded quadrants indicate sign agreement; all games with $|\Delta\mathrm{HNS}|>5$ fall inside ($7/7$; $\rho_{\mathrm{Spearman}}=+0.52$). The $y$-axis is symlog.}
\label{fig:phi}
\end{figure}

From each game's $5$M-step checkpoint, we form $\pgood$ from top-$25\%$ rollout episodes and $\pcurr$ from all episodes, and estimate \reviewblue{$\hat\Align(g)=\operatorname{mean}_t[(\clip(\pgood(a_t)/\pi(a_t\mid s_t),0,10)-1)\hat A_t]$ with batch-standardized raw-reward GAE from the saved value head}, without training. \reviewblue{After omitting Jamesbond for the reason in Figure~\ref{fig:phi},} its sign agrees with the realized OPR$-$IMPALA gap on all seven games with $|\Delta\mathrm{HNS}|>5$ ($\rho_{\mathrm{Spearman}}=+0.52$), and is stable \reviewblue{across the nominal $30$- and $100$-episode budgets}. This retrospective, in-sample learned-critic diagnostic checks the predicted help/harm distinction; it is not a held-out test. \reviewblue{The supplementary predictor protocol gives realized episode counts.}

\subsection{Extension: Self-Gated OPR}
\label{sec:gated}

\reviewblue{The gating extension shapes where estimated alignment is positive and otherwise uses the raw update.} Corollary~\ref{cor:local-gate} establishes a non-negative first-order contribution for the \reviewblue{theoretical oracle gate using the exact local update coefficient}. Empirically, the checkpoint-oracle gate based on $\Align$ reaches $165.6\pm13.7\%$ median HNS across its two available seeds ($175.3\%$ at its best), above the same-seed backbone in both. This is \reviewblue{a retrospective oracle, not an online gate.}

We train two online gates end to end. The importance-sampling gate reaches $151.2\%$ at seed 42 (backbone $152.6\%$). The \reviewblue{online covariance gate} reaches $167.3\%$, $151.1\%$, and $161.9\%$ across seeds, or $160.1\pm8.2\%$ versus $142.7\pm17.0\%$ for the backbone. Its same-seed margins are $+14.7$, $-1.2$, and $+38.9$ pp. At seed 42 it captures Frostbite upside ($655$ versus $208$) and avoids the ungated Pong loss, although Kangaroo simultaneously loses $95$ HNS; the improvement is an average across games and seeds.

\begin{figure}[t]
\centering
\includegraphics[width=0.70\columnwidth]{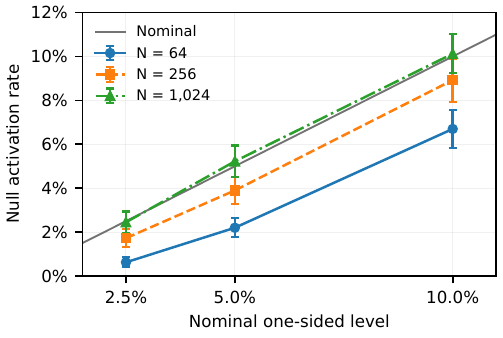}
\Description{Observed null activation versus the nominal one-sided level for three batch sizes in the held-out synthetic gate experiment.}
\caption{Selective-gate calibration. Four independent fold-pair estimates form a $t_3$ abstention rule; the diagonal is the nominal level. Bars are $95\%$ intervals clustered over $200$ environments ($20$ repetitions each). This is empirical calibration, not a neural-PPO certificate.}
\label{fig:gate-calibration}
\end{figure}

\paragraph{Estimation error and abstention.} On $200$ randomized delayed-credit environments with $20$ repetitions each, a held-out propagated-score estimate and four-fold $t$ threshold test the tradeoff in Corollary~\ref{cor:selective-gate}. At nominal $\alpha=0.05$, null activation is $2.20\%$, $3.90\%$, and $5.23\%$ for batches of $64$, $256$, and $1{,}024$, while \reviewblue{agreement with the exact gate's clean-case decisions} rises from $48.6\%$ to $92.2\%$ (Figure~\ref{fig:gate-calibration}). \reviewblue{This calibrated synthetic gate differs from both the theoretical bounded-sample rule and the Atari gate.}

\paragraph{Estimating the right update effect.} Reliability also depends on the target. With separately normalized raw and shaped advantages, covariance agrees in sign with the exact matched-update coefficient on $50.4\%$ of $800$ random MDPs at $\ssc=0.5$. In a separate sampled study, sharing the raw normalization scale raises agreement from $51.2\%$ to $88.0\%$ at batch $2{,}048$. Estimating the normalized update difference directly on held-out samples reaches $95.3\%$ agreement at $\ssc=0.1$; its one-sided rule activates on $17.1\%$ of decisions, with $0.76\%$ target conflict among activations. \reviewblue{Matching the update and controlling uncertainty address different errors (full grids and independent replications in the supplement).}

\subsection{Ablations}
\label{sec:ablation}

\paragraph{Shaping scale.} Relative to the IMPALA reference ($\ssc=0$, $152.6\%$ median HNS), OPR reaches $153.8\%$, $154.1\%$, and $156.6\%$ at $\ssc=0.1,0.25,0.5$, degrades to $143.9\%$ at $\ssc=1$, and falls to $99.5\%$ at $\ssc=2$ as shaping dominates the task reward. We use $\ssc=0.5$ throughout.

\paragraph{Cluster count $K$ and return-quantile bins.} \reviewblue{Atari conditioning has mixed effects: $K=4$ scores below $K=1$ at one matched seed ($98.3\%$ versus $156.6\%$) but above it at another ($153.4\%$ versus $109.1\%$; full results in the supplement).} In a paired $16$-state controlled study, state conditioning instead improves reward by $+0.408\pm0.001$ when actions are state-aliased, whereas with eight elite samples the marginal prior improves reward by $+0.263\pm0.006$ when one action is shared across states ($95\%$ CIs). Thus larger $K$ trades lower aliasing bias for noisier estimates, as predicted by \S\ref{sec:th-l1}. The return-quantile prior reaches $156.2\%$ at $Q=4$ and $107.7\%$ at $Q=2$ (both seed 43); since the top-half setting has more samples, sample count alone does not explain this ordering. Details are in the supplement.

\paragraph{Decay.} Disabling decay is close in median to the default ($154.9\%$ versus $156.6\%$); the large mean-HNS difference is due to one Jamesbond outlier. This Atari comparison does not identify decay as load-bearing at the tested horizon. In a controlled multi-state stress test with a reversed prior and $\ssc=2$, constant shaping converges to normalized progress $-0.805$, whereas decay over $40$ updates recovers $0.990$ versus $0.995$ for raw learning. Decay removes a persistent shaping force but does not undo transient damage; the full horizon sweep and independent replication are in the supplement.

\paragraph{Learned prior.} \reviewblue{Replacing the histogram with a trained behavioral-cloning prior reaches $63.1\%$ average median HNS with decay ($68.4\%$ without), well below OPR's $146.6\%$.} \reviewblue{This favors the histogram here without isolating why the learned prior underperforms.}

\subsection{Beyond Atari}
\label{sec:cage}

\paragraph{CAGE-2.} \reviewblue{On Cyber Autonomy Gym for Experimentation Challenge 2 (CAGE-2)~\citep{cage2022challenge},} we add OPR to the cyber-defense challenge winner's PPO agent, tuning only the shaping scale to the domain's reward magnitude ($\ssc=0.1$). Against the direct B\_line attacker, OPR improves mean $100$-step return from $-15.31$ to $-14.80$ with the same direction across tested seeds; against the exploratory Meander attacker it is approximately neutral. \reviewblue{We use ungated OPR here: the online covariance gate's near-zero estimated alignment does not reliably resolve its sign, and gating removes the B\_line gain.} Full results are in the supplement.

\paragraph{Crafter.} As a held-out transfer test \reviewblue{in this survival-and-crafting environment}, we evaluate a frozen configuration for one million interactions over ten paired seeds. The PPO backbone obtains the official Crafter score $4.775\pm0.282$, closely reproducing the published PPO reference $4.6\pm0.3$~\citep{hafner2022crafter}; marginal $K{=}1$ OPR obtains $4.873\pm0.157$ (paired difference $+0.097$, $95\%$ CI $[-0.129,+0.325]$, $6/10$ wins). This supports transfer of the method, although the score difference remains unresolved. \reviewblue{The supplement separates the frozen comparison from follow-up diagnostics.}

\section{Related Work}
\label{sec:related}

\paragraph{Reward shaping and curiosity.} Potential-based shaping preserves the optimal policy \citep{ng1999shaping}; OPR is not potential-based and instead decays its perturbation. Curiosity methods---ICM \citep{pathak2017icm}, \reviewblue{Random Network Distillation (RND)} \citep{burda2019rnd}, and \reviewblue{Never Give Up (NGU)} \citep{badia2020ngu}---reward novelty through learned signals, whereas OPR rewards resemblance to elite action statistics.

\paragraph{Self-imitation.} SIL \citep{oh2018sil} and generalized lower-bound $Q$-learning \citep{tang2020glbq} replay good transitions through a value-gated auxiliary loss. \reviewblue{RAPID (exploration via ranking episodes)} \citep{zha2021rapid} is closer still in its episode-level view: it ranks episodes by extrinsic and exploration scores, stores their state--action pairs, and behavior-clones the ranking buffer. OPR instead discards state from its deployed prior, performs no replay optimization, and routes a histogram ratio through the reward before GAE. Unlike return-conditioned policies \citep{schmidhuber2019udrl,chen2021dt}, OPR does not condition action generation on a target return; it extracts behavioral statistics from achieved high-return episodes. \reviewblue{The marginal prior trades away state information, motivating the raw-advantage alignment check.}

\paragraph{Adversarial imitation and regularized RL.} OPR shares density-ratio structure with adversarial imitation methods such as GAIL and AIRL \citep{ho2016gail,fu2018airl}, but it neither learns a discriminator nor estimates a state--action occupancy ratio.  \reviewblue{The raw action log-ratio has a KL interpretation; the bounded score has a triangular-discrimination identity (\S\ref{sec:theory}).}  \reviewblue{OPR also differs from regularized MDP objectives} \citep{geist2019regularized,neu2017unified}.  \reviewblue{These motivate proximal updates, not an exact mirror-descent interpretation of neural PPO} \citep{vieillard2020munchausen}.  The top-return filter is related to hard advantage weighting \citep{peters2007rwr,peng2019awr}, without an auxiliary optimization loop.

\reviewblue{Adversarially Guided Actor--Critic (AGAC)} \citep{fletberliac2021agac} also uses an action log-ratio in actor--critic learning \reviewblue{to reward novelty: an adversary predicts the actor, which learns to be difficult to predict.}  \reviewblue{OPR instead rewards resemblance to elite action statistics, without an auxiliary adversarial objective.}

\section{Conclusion}
\label{sec:conclusion}

\reviewblue{OPR turns elite statistics into a bounded pre-GAE reward without auxiliary networks or replay optimization.} \reviewblue{It is competitive across $49$ Atari games and exceeds matched PPO on average.} \reviewblue{Elite selection supplies useful information; pre-GAE placement propagates credit; state conditioning trades bias against variance; and gating trades harmful activation against missed benefit. Atari tests practical relevance, with CAGE-2 and Crafter as transfer checks.} The guarantees are local, with preliminary \reviewblue{state clusters} and a checkpoint-based alignment diagnostic. \reviewblue{This is a starting point for deciding when and where to learn from self-generated experience.} Future work should learn robust action-relevant \reviewblue{state clusters}. It should also develop \reviewblue{gates matched to the actual policy update} accounting for advantage normalization, critic error, and the complete PPO update path.

\bibliographystyle{plainnat}
\bibliography{references}

@misc{cage2022challenge,
  title={{CAGE Challenge 2}},
  author={{TTCP CAGE Working Group}},
  year={2022},
  howpublished={\url{https://github.com/cage-challenge/cage-challenge-2}}
}

@inproceedings{oh2018sil,
  title={Self-Imitation Learning},
  author={Oh, Junhyuk and Guo, Yijie and Singh, Satinder and Lee, Honglak},
  booktitle={Proceedings of the 35th International Conference on Machine Learning},
  pages={3878--3887},
  year={2018},
  organization={PMLR}
}

@article{schulman2017ppo,
  title={Proximal Policy Optimization Algorithms},
  author={Schulman, John and Wolski, Filip and Dhariwal, Prafulla and Radford, Alec and Klimov, Oleg},
  journal={arXiv preprint arXiv:1707.06347},
  year={2017}
}

@inproceedings{schulman2016gae,
  title={High-Dimensional Continuous Control Using Generalized Advantage Estimation},
  author={Schulman, John and Moritz, Philipp and Levine, Sergey and Jordan, Michael I. and Abbeel, Pieter},
  booktitle={International Conference on Learning Representations},
  year={2016}
}

@inproceedings{espeholt2018impala,
  title={{IMPALA}: Scalable Distributed Deep-{RL} with Importance Weighted Actor-Learner Architectures},
  author={Espeholt, Lasse and Soyer, Hubert and Munos, Remi and Simonyan, Karen and Mnih, Vlad and Ward, Tom and Doron, Yotam and Firoiu, Vlad and Harley, Tim and Dunning, Iain and Legg, Shane and Kavukcuoglu, Koray},
  booktitle={Proceedings of the 35th International Conference on Machine Learning},
  pages={1407--1416},
  year={2018},
  organization={PMLR}
}

@inproceedings{agarwal2021rliable,
  title={Deep Reinforcement Learning at the Edge of the Statistical Precipice},
  author={Agarwal, Rishabh and Schwarzer, Max and Castro, Pablo Samuel and Courville, Aaron C. and Bellemare, Marc G.},
  booktitle={Advances in Neural Information Processing Systems},
  volume={34},
  pages={29304--29320},
  year={2021}
}

@inproceedings{ng1999shaping,
  title={Policy Invariance Under Reward Transformations: Theory and Application to Reward Shaping},
  author={Ng, Andrew Y. and Harada, Daishi and Russell, Stuart},
  booktitle={Proceedings of the Sixteenth International Conference on Machine Learning},
  pages={278--287},
  year={1999}
}

@inproceedings{pathak2017icm,
  title={Curiosity-driven Exploration by Self-Supervised Prediction},
  author={Pathak, Deepak and Agrawal, Pulkit and Efros, Alexei A. and Darrell, Trevor},
  booktitle={Proceedings of the 34th International Conference on Machine Learning},
  pages={2778--2787},
  year={2017},
  organization={PMLR}
}

@inproceedings{burda2019rnd,
  title={Exploration by Random Network Distillation},
  author={Burda, Yuri and Edwards, Harrison and Storkey, Amos and Klimov, Oleg},
  booktitle={International Conference on Learning Representations},
  year={2019}
}

@inproceedings{badia2020ngu,
  title={Never Give Up: Learning Directed Exploration Strategies},
  author={Badia, Adri{\`a} Puigdom{\`e}nech and Sprechmann, Pablo and Vitvitskyi, Alex and Guo, Daniel and Piot, Bilal and Kapturowski, Steven and Tieleman, Olivier and Arjovsky, Mart{\'i}n and Pritzel, Alexander and Bolt, Andew and Blundell, Charles},
  booktitle={International Conference on Learning Representations},
  year={2020}
}

@inproceedings{tang2020glbq,
  title={Self-Imitation Learning via Generalized Lower Bound Q-Learning},
  author={Tang, Yunhao},
  booktitle={Advances in Neural Information Processing Systems},
  volume={33},
  pages={13964--13975},
  year={2020}
}

@article{schmidhuber2019udrl,
  title={Reinforcement Learning Upside Down: Don't Predict Rewards -- Just Map Them to Actions},
  author={Schmidhuber, J{\"u}rgen},
  journal={arXiv preprint arXiv:1912.02875},
  year={2019}
}

@inproceedings{chen2021dt,
  title={Decision Transformer: Reinforcement Learning via Sequence Modeling},
  author={Chen, Lili and Lu, Kevin and Rajeswaran, Aravind and Lee, Kimin and Grover, Aditya and Laskin, Misha and Abbeel, Pieter and Srinivas, Aravind and Mordatch, Igor},
  booktitle={Advances in Neural Information Processing Systems},
  volume={34},
  pages={15084--15097},
  year={2021}
}

@article{mnih2015dqn,
  title={Human-level control through deep reinforcement learning},
  author={Mnih, Volodymyr and Kavukcuoglu, Koray and Silver, David and Rusu, Andrei A. and Veness, Joel and Bellemare, Marc G. and Graves, Alex and Riedmiller, Martin and Fidjeland, Andreas K. and Ostrovski, Georg and Petersen, Stig and Beattie, Charles and Sadik, Amir and Antonoglou, Ioannis and King, Helen and Kumaran, Dharshan and Wierstra, Daan and Legg, Shane and Hassabis, Demis},
  journal={Nature},
  volume={518},
  pages={529--533},
  year={2015}
}

@inproceedings{mnih2016a3c,
  title={Asynchronous Methods for Deep Reinforcement Learning},
  author={Mnih, Volodymyr and Badia, Adri{\`a} Puigdom{\`e}nech and Mirza, Mehdi and Graves, Alex and Lillicrap, Timothy P. and Harley, Tim and Silver, David and Kavukcuoglu, Koray},
  booktitle={Proceedings of the 33rd International Conference on Machine Learning},
  pages={1928--1937},
  year={2016},
  organization={PMLR}
}

@inproceedings{ho2016gail,
  title={Generative Adversarial Imitation Learning},
  author={Ho, Jonathan and Ermon, Stefano},
  booktitle={Advances in Neural Information Processing Systems},
  volume={29},
  year={2016}
}

@inproceedings{fu2018airl,
  title={Learning Robust Rewards with Adversarial Inverse Reinforcement Learning},
  author={Fu, Justin and Luo, Katie and Levine, Sergey},
  booktitle={International Conference on Learning Representations},
  year={2018}
}

@inproceedings{geist2019regularized,
  title={A Theory of Regularized Markov Decision Processes},
  author={Geist, Matthieu and Scherrer, Bruno and Pietquin, Olivier},
  booktitle={Proceedings of the 36th International Conference on Machine Learning},
  pages={2160--2169},
  year={2019},
  organization={PMLR}
}

@article{neu2017unified,
  title={A Unified View of Entropy-Regularized {Markov} Decision Processes},
  author={Neu, Gergely and Jonsson, Anders and G{\'o}mez, Vicen{\c{c}}},
  journal={arXiv preprint arXiv:1705.07798},
  year={2017}
}

@inproceedings{vieillard2020munchausen,
  title={Munchausen Reinforcement Learning},
  author={Vieillard, Nino and Pietquin, Olivier and Geist, Matthieu},
  booktitle={Advances in Neural Information Processing Systems},
  volume={33},
  pages={4235--4246},
  year={2020}
}

@inproceedings{peters2007rwr,
  title={Reinforcement Learning by Reward-Weighted Regression for Operational Space Control},
  author={Peters, Jan and Schaal, Stefan},
  booktitle={Proceedings of the 24th International Conference on Machine Learning},
  pages={745--750},
  year={2007}
}

@article{peng2019awr,
  title={Advantage-Weighted Regression: Simple and Scalable Off-Policy Reinforcement Learning},
  author={Peng, Xue Bin and Kumar, Aviral and Zhang, Grace and Levine, Sergey},
  journal={arXiv preprint arXiv:1910.00177},
  year={2019}
}

@inproceedings{abdolmaleki2018mpo,
  title={Maximum a Posteriori Policy Optimisation},
  author={Abdolmaleki, Abbas and Springenberg, Jost Tobias and Tassa, Yuval and Munos, Remi and Heess, Nicolas and Riedmiller, Martin},
  booktitle={International Conference on Learning Representations},
  year={2018}
}

@inproceedings{kakade2002cpi,
  title={Approximately Optimal Approximate Reinforcement Learning},
  author={Kakade, Sham and Langford, John},
  booktitle={Proceedings of the 19th International Conference on Machine Learning},
  pages={267--274},
  year={2002}
}

@article{huang2022cleanrl,
  title={CleanRL: High-quality Single-file Implementations of Deep Reinforcement Learning Algorithms},
  author={Huang, Shengyi and Dossa, Rousslan Fernand Julien and Ye, Chang and Braga, Jeff and Chakraborty, Dipam and Mehta, Kinal and Ara{\'u}jo, Jo{\~a}o G.M.},
  journal={Journal of Machine Learning Research},
  volume={23},
  number={274},
  pages={1--18},
  year={2022}
}

@inproceedings{bellemare2016unifying,
  title={Unifying Count-Based Exploration and Intrinsic Motivation},
  author={Bellemare, Marc G. and Srinivasan, Sriram and Ostrovski, Georg and Schaul, Tom and Saxton, David and Munos, R{\'e}mi},
  booktitle={Advances in Neural Information Processing Systems},
  year={2016}
}

@inproceedings{schaul2016per,
  title={Prioritized Experience Replay},
  author={Schaul, Tom and Quan, John and Antonoglou, Ioannis and Silver, David},
  booktitle={International Conference on Learning Representations},
  year={2016}
}

@article{cui2025prime,
  title={Process Reinforcement through Implicit Rewards},
  author={Cui, Ganqu and Yuan, Lifan and Wang, Zefan and Wang, Hanbin and Zhang, Yuchen and Chen, Jiacheng and Li, Wendi and He, Bingxiang and Fan, Yuchen and Yu, Tianyu and Xu, Qixin and Chen, Weize and Yuan, Jiarui and Chen, Huayu and Zhang, Kaiyan and Lv, Xingtai and Wang, Shuo and Yao, Yuan and Han, Xu and Peng, Hao and Cheng, Yu and Liu, Zhiyuan and Sun, Maosong and Zhou, Bowen and Ding, Ning},
  journal={arXiv preprint arXiv:2502.01456},
  year={2025}
}

@article{qin2025spear,
  title={Learn the Ropes, Then Trust the Wins: Self-imitation with Progressive Exploration for Agentic Reinforcement Learning},
  author={Qin, Yulei and Tan, Xiaoyu and He, Zhengbao and Li, Gang and Lin, Haojia and Li, Zongyi and Xu, Zihan and Shi, Yuchen and Cai, Siqi and Rui, Renting and Cai, Shaofei and Cai, Yuzheng and Zhang, Xuan and Ye, Sheng and Li, Ke and Sun, Xing},
  journal={arXiv preprint arXiv:2509.22601},
  year={2025}
}

@inproceedings{zha2021rapid,
  title={Rank the Episodes: A Simple Approach for Exploration in Procedurally-Generated Environments},
  author={Zha, Daochen and Ma, Wenye and Yuan, Lei and Hu, Xia and Liu, Ji},
  booktitle={International Conference on Learning Representations},
  year={2021},
  url={https://openreview.net/forum?id=MtEE0CktZht}
}

@inproceedings{fletberliac2021agac,
  title={Adversarially Guided Actor-Critic},
  author={Flet-Berliac, Yannis and Ferret, Johan and Pietquin, Olivier and Preux, Philippe and Geist, Matthieu},
  booktitle={International Conference on Learning Representations},
  year={2021},
  url={https://arxiv.org/abs/2102.04376}
}

@inproceedings{hafner2022crafter,
  title={Benchmarking the Spectrum of Agent Capabilities},
  author={Hafner, Danijar},
  booktitle={International Conference on Learning Representations},
  year={2022},
  url={https://openreview.net/forum?id=1W0z96MFEoH}
}

\clearpage
\appendix
\raggedbottom
\emergencystretch=2em
\section*{Supplementary Material}
\addcontentsline{toc}{section}{Supplementary Material}

\section{Notation}
The notation below is used consistently in the main paper and throughout this
supplement.

\begin{table}[H]
\centering\small
\caption{Notation.}
\label{tab:notation}
\resizebox{\columnwidth}{!}{%
\begin{tabular}{@{}ll@{}}
\toprule
Symbol & Meaning \\
\midrule
$\Buf$ & good-episode buffer (top-$25\%$ returns) \\
$\pgood(a\mid k)$ & elite prior; marginal $\pgood(a)$ at $K{=}1$ (Eq.~\ref{eq:pgood}) \\
$\pcurr(a\mid k)$ & current-policy conditional action prior (EMA); marginal at $K{=}1$ \\
$\dratio(a\mid k)$ & raw log-ratio $\log\pgood(a\mid k)-\log\pcurr(a\mid k)$ \\
$\sdratio(a\mid k)$ & bounded transform of the conditional log-ratio (Eq.~\ref{app:eq:sdelta}) \\
$\bonus_t$ & per-step shaping bonus (Eq.~\ref{eq:bonus}) \\
$\srew_t$ & shaped reward $\rew_t+\bonus_t$ \\
$\ssc,\ \dec,\ \Edec$ & shaping scale, decay factor, decay horizon \\
$\rstd$ & batch mean absolute reward, floored at $10^{-4}$ \\
$\Nbuf,\ \eta$ & transition count in the selected elite episodes, Laplace constant \\
$K,\ \Kep$ & number of state clusters, number of top buffered episodes \\
$\hat A_t$ & GAE advantage; $\Adv(s,a)$ true advantage \\
$\rocc_\pi,\ \dpi$ & action-occupancy / state-visitation of $\pi$ \\
$\Align(g)$ & prior--advantage alignment predictor (Sec.~\ref{sec:predictive}) \\
\bottomrule
\end{tabular}}
\end{table}

\section{Full Per-Game Results}
\label{app:full}

Table~\ref{tab:appendix-full} reports OPR scores on all $56$ ALE games we
evaluated (single seed, $10$M agent steps), alongside published baselines on the
$49$-game comparison subset.

\paragraph{Which game set is which.}
Two game counts appear in this paper and we state both explicitly to avoid any
ambiguity. The \textbf{$49$-game subset} is the set of games for which \emph{all} the
compared baselines (DQN, A2C, PPO, ACPER, SIL) publish per-game scores; it is the classic
DQN-era comparison set, and every aggregate in Table~\ref{tab:broad} is computed over it.
The \textbf{$56$-game set} is every ALE game we ran; over it, OPR's single-seed median HNS
is $127.5\%$ (median is lower because the extra seven games include several on which no
$10$M-agent-step policy is competitive). We report the $49$-game aggregate in the main text
purely because it is the only set on which a like-for-like comparison with the published
baselines is possible, not because it is the more favorable number---and we note that
the choice of game set moves the headline by $21$ pp, which is itself a reason to treat
single-seed cross-paper Atari aggregates with caution (\S\ref{sec:controlled}).

\begin{figure}[t]
\centering
\includegraphics[width=0.92\columnwidth]{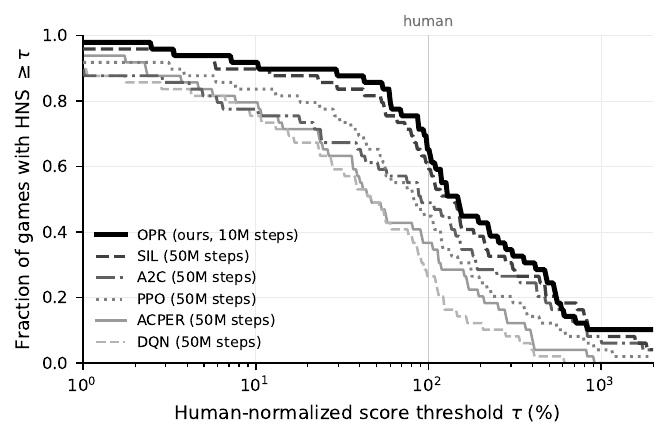}
\caption{Performance profiles on the $49$-game comparison set: fraction of games with HNS $\ge\tau$. OPR's profile is at or above the published baselines over essentially the whole threshold range. The methods use different backbones, action repeats, and codebases; this is descriptive context for the same-backbone comparison in the main paper.}
\label{fig:profile}
\end{figure}

\section{Frame-Budget Context}
\label{app:context}

Table~\ref{tab:context} places OPR against strong published agents purely as a function
of the environment-frame budget. We stress what this table is \emph{not}: the rows use
different backbones, different learners (V-trace vs.\ PPO), different codebases,
different game sets, and a single seed, so no row is an ablation of any other and no
causal claim can be read from it. In particular the deep IMPALA agent
\citep{espeholt2018impala}---whose \emph{encoder} we borrow, though not its off-policy
V-trace learner---reaches a higher median than OPR at $5\times$ the agent decisions, and
we do not claim otherwise. The one thing the table does support is a statement about
budget: from $10$M agent steps ($160$M raw frames at our action-repeat-$16$ protocol;
see the main paper's Setup) OPR is in the range of modern PPO and shallow IMPALA trained
to $200$M frames.
All claims of \emph{improvement} in this paper rest instead on the architecture-controlled,
seed-matched comparison of Table~\ref{tab:selected}.

\begin{table}[h]
\centering\small
\caption{Frame-budget context. Not architecture-controlled; different learners, codebases,
and game sets. Provided for calibration only.}
\label{tab:context}
\begin{tabular}{@{}llrr@{}}
\toprule
Method & Source & Steps & Median HNS \\
\midrule
IMPALA (deep)    & \citet{espeholt2018impala} & 50M & 191.8\% \\
CleanRL PPO      & \citet{huang2022cleanrl}   & 50M & $\approx$102\% \\
IMPALA (shallow) & \citet{espeholt2018impala} & 50M & \phantom{0}93.2\% \\
\midrule
OPR (ours), 49-game & this work & \textbf{10M} & 148.4\% \\
OPR (ours), 56-game & this work & \textbf{10M} & 127.5\% \\
\bottomrule
\end{tabular}
\end{table}

\begin{table*}[t]
\centering
\caption{Full per-game results. OPR uses the IMPALA ResNet backbone at 10M agent steps (160M ALE frames; see Setup), single seed (42). Published baselines (DQN, A2C, PPO, ACPER, SIL) are quoted from their original papers at 50M agent steps (200M frames). HNS = human-normalized score, computed before display rounding; \textbf{bold} highlights OPR when it reaches the best compared score. The first block is the 49-game comparison subset; the second block lists games without the full set of comparator scores.}
\label{tab:appendix-full}
\footnotesize
\setlength{\tabcolsep}{4pt}
\renewcommand{\arraystretch}{0.94}
\begin{tabular}{@{}lrrrrrrrr r@{}}
\toprule
Game & Random & Human & DQN & A2C & PPO & ACPER & SIL & \textbf{OPR} & HNS\% \\
\midrule
\multicolumn{10}{c}{\emph{49-game comparison subset}}\\
\midrule
Alien & 228 & 7{,}128 & 227 & 1{,}859 & 1{,}476 & 390 & 2{,}242 & \textbf{3{,}115} & 41.8 \\
Amidar & 5.8 & 1{,}720 & 164 & 740 & 757 & 425 & 1{,}362 & \textbf{3{,}795} & 221.1 \\
Assault & 222 & 742 & 538 & 1{,}981 & 4{,}971 & 818 & 1{,}812 & \textbf{11{,}454} & 2161.5 \\
Asterix & 210 & 8{,}503 & 855 & 16{,}083 & 4{,}531 & 3{,}533 & 17{,}984 & \textbf{44{,}300} & 531.6 \\
Asteroids & 719 & 47{,}389 & 1{,}008 & 2{,}056 & 2{,}195 & 1{,}780 & 2{,}259 & \textbf{2{,}292} & 3.4 \\
Atlantis & 12{,}850 & 29{,}028 & 58{,}012 & 3{,}032{,}444 & 3{,}010{,}196 & 58{,}012 & 3{,}084{,}782 & 869{,}070 & 5292.5 \\
BankHeist & 14 & 753 & 421 & 1{,}334 & 1{,}119 & 1{,}203 & 1{,}138 & 661 & 87.5 \\
BattleZone & 2{,}360 & 37{,}188 & 20{,}433 & 10{,}683 & 16{,}053 & 15{,}025 & 25{,}075 & \textbf{55{,}700} & 153.2 \\
BeamRider & 364 & 16{,}926 & 2{,}897 & 3{,}932 & 5{,}046 & 2{,}602 & 2{,}366 & \textbf{10{,}495} & 61.2 \\
Bowling & 23 & 161 & 46 & 31 & 41 & 59 & 31 & \textbf{63} & 29.2 \\
Boxing & 0.1 & 12 & 73 & 100 & 95 & 100 & 100 & 57 & 474.2 \\
Breakout & 1.7 & 30 & 120 & 502 & 339 & 118 & 452 & 82 & 278.8 \\
Centipede & 2{,}091 & 12{,}017 & 2{,}754 & 3{,}858 & 4{,}418 & 7{,}790 & 7{,}560 & \textbf{55{,}558} & 538.6 \\
ChopperCommand & 811 & 7{,}388 & 1{,}090 & 3{,}464 & 2{,}837 & 1{,}308 & 6{,}710 & \textbf{41{,}120} & 612.9 \\
CrazyClimber & 10{,}780 & 35{,}829 & 34{,}390 & 129{,}716 & 104{,}169 & 19{,}919 & 130{,}186 & \textbf{148{,}880} & 551.3 \\
DemonAttack & 152 & 1{,}971 & 6{,}310 & 18{,}331 & 11{,}370 & 4{,}778 & 10{,}140 & \textbf{78{,}400} & 4302.0 \\
DoubleDunk & -19 & -16 & -16 & -0.5 & -14 & -9.8 & 22 & -0.6 & 818.2 \\
Enduro & 0 & 860 & 277 & 0 & 694 & 3{,}113 & 1{,}205 & 872 & 101.4 \\
FishingDerby & -92 & -39 & -14 & 39 & 23 & 60 & 56 & 27 & 224.7 \\
Freeway & 0 & 30 & 23 & 0 & 32 & 31 & 32 & \textbf{32} & 109.8 \\
Frostbite & 65 & 4{,}335 & 110 & 340 & 314 & 2{,}342 & 6{,}290 & \textbf{24{,}926} & 582.3 \\
Gopher & 258 & 2{,}412 & 2{,}148 & 9{,}358 & 4{,}546 & 3{,}920 & 23{,}304 & 4{,}428 & 193.5 \\
Gravitar & 173 & 3{,}351 & 263 & 329 & 415 & 628 & 1{,}874 & 400 & 7.1 \\
Hero & 1{,}027 & 30{,}826 & 12{,}300 & 28{,}008 & 29{,}105 & 13{,}299 & 33{,}157 & \textbf{36{,}298} & 118.4 \\
IceHockey & -11 & 0.9 & -4.4 & -4.3 & -2.1 & 0 & -2.4 & \textbf{0.6} & 97.5 \\
Jamesbond & 29 & 303 & 365 & 399 & 511 & 598 & 311 & \textbf{61{,}710} & 22527.8 \\
Kangaroo & 52 & 3{,}035 & 3{,}340 & 1{,}563 & 5{,}318 & 5{,}875 & 2{,}888 & \textbf{14{,}470} & 483.3 \\
Krull & 1{,}598 & 2{,}666 & 3{,}800 & 8{,}884 & 7{,}488 & 11{,}323 & 10{,}615 & 5{,}379 & 354.2 \\
KungFuMaster & 258 & 22{,}736 & 20{,}500 & 32{,}508 & 26{,}640 & 20{,}485 & 34{,}449 & 13{,}710 & 59.8 \\
MontezumaRevenge & 0 & 4{,}753 & 0 & 5.8 & 0 & 0 & 1{,}100 & 0 & 0.0 \\
MsPacman & 307 & 6{,}952 & 1{,}016 & 2{,}843 & 2{,}650 & 1{,}016 & 4{,}025 & \textbf{4{,}257} & 59.4 \\
NameThisGame & 2{,}292 & 8{,}049 & 3{,}888 & 11{,}174 & 7{,}242 & 2{,}888 & 14{,}958 & 9{,}657 & 127.9 \\
Pong & -21 & 15 & 19 & 21 & 21 & 21 & 21 & -1.7 & 53.8 \\
PrivateEye & 25 & 69{,}571 & 100 & 211 & 100 & 100 & 661 & \textbf{1{,}743} & 2.5 \\
Qbert & 164 & 13{,}455 & 3{,}800 & 17{,}605 & 14{,}104 & 657 & 104{,}976 & 33{,}888 & 253.7 \\
Riverraid & 1{,}338 & 17{,}118 & 3{,}400 & 13{,}036 & 7{,}963 & 2{,}224 & 14{,}306 & 12{,}223 & 69.0 \\
RoadRunner & 12 & 7{,}845 & 9{,}562 & 39{,}874 & 34{,}400 & 8{,}925 & 57{,}072 & 32{,}250 & 411.5 \\
Robotank & 2.2 & 12 & 10 & 3.2 & 15 & 7.7 & 10 & \textbf{32} & 304.1 \\
Seaquest & 68 & 42{,}055 & 804 & 1{,}795 & 1{,}680 & 804 & 2{,}456 & \textbf{4{,}388} & 10.3 \\
SpaceInvaders & 148 & 1{,}669 & 730 & 2{,}466 & 1{,}000 & 730 & 2{,}952 & 1{,}468 & 86.8 \\
StarGunner & 664 & 10{,}250 & 10{,}000 & 57{,}372 & 35{,}000 & 1{,}108 & 31{,}309 & 12{,}840 & 127.0 \\
Tennis & -24 & -8.3 & -17 & -10 & -14 & -17 & -17 & \textbf{-0.8} & 148.4 \\
TimePilot & 3{,}568 & 5{,}229 & 3{,}952 & 5{,}347 & 4{,}500 & 3{,}952 & 10{,}812 & 5{,}530 & 118.1 \\
Tutankham & 11 & 168 & 271 & 306 & 200 & 271 & 340 & 159 & 94.6 \\
UpNDown & 533 & 11{,}693 & 9{,}562 & 48{,}132 & 35{,}000 & 9{,}562 & 53{,}315 & \textbf{79{,}770} & 710.0 \\
Venture & 0 & 1{,}188 & 0 & 0 & 0 & 0 & 0 & \textbf{1{,}230} & 103.6 \\
VideoPinball & 16{,}257 & 17{,}668 & 21{,}798 & 391{,}242 & 5{,}000 & 21{,}798 & 461{,}522 & 308{,}508 & 20712.3 \\
WizardOfWor & 564 & 4{,}756 & 1{,}550 & 4{,}197 & 4{,}000 & 1{,}550 & 7{,}088 & 4{,}680 & 98.2 \\
Zaxxon & 32 & 9{,}173 & 4{,}279 & 124 & 5{,}000 & 4{,}279 & 9{,}164 & \textbf{13{,}790} & 150.5 \\
\midrule
\multicolumn{10}{c}{\emph{OPR-only games (no published baselines)}}\\
\midrule
Berzerk & 124 & 2{,}630 & -- & -- & -- & -- & -- & \textbf{1{,}184} & 42.3 \\
Phoenix & 761 & 7{,}243 & -- & -- & -- & -- & -- & \textbf{5{,}236} & 69.0 \\
Pitfall & -229 & 6{,}464 & -- & -- & -- & -- & -- & \textbf{0} & 3.4 \\
Pooyan & 474 & 4{,}765 & -- & -- & -- & -- & -- & \textbf{14{,}620} & 329.7 \\
Skiing & -17{,}098 & -4{,}337 & -- & -- & -- & -- & -- & \textbf{-8{,}896} & 64.3 \\
Solaris & 1{,}236 & 12{,}327 & -- & -- & -- & -- & -- & \textbf{2{,}084} & 7.6 \\
YarsRevenge & 3{,}093 & 54{,}577 & -- & -- & -- & -- & -- & \textbf{94{,}617} & 177.8 \\
\bottomrule
\end{tabular}
\end{table*}


\section{Score Distributions and Theory-Aligned Splits}
\label{app:profiles}

The performance-profile figure is included in this supplement (Fig.~\ref{fig:profile}); at $\tau{=}100$ the
per-method fractions are OPR $0.65$, SIL $0.59$, A2C $0.49$, PPO $0.45$, ACPER $0.37$,
DQN $0.27$. Against SIL the sign test is inconclusive ($26/49$, $p=0.78$); this does
not establish equivalence. OPR's descriptive performance profile is nevertheless at
or above SIL's over the plotted range.

\paragraph{Theory-aligned split of the benchmark.}
The state-aliasing construction motivated the hypothesis that OPR's advantage would
concentrate on sparse, skill-structured games and its losses on reactive control.
Splitting the $49$ games by
the standard exploration taxonomy \citep{bellemare2016unifying}, the paired OPR$-$SIL
median difference is $+2.5$ pp on the $14$ hard-exploration games ($+10.5$ on the
hard/dense-reward subset) versus $-13.2$ pp on the $35$ easy-exploration games, and the
easy-category games where SIL scores higher are predominantly reactive-control titles (Pong,
Breakout, Boxing, Enduro, FishingDerby, SpaceInvaders). The direction matches the hypothesis,
but at $n{=}14$ hard games the subset sign test is far from significant
($9/14$, $p{=}0.42$); we report this split as \emph{descriptive} evidence only. By
contrast, OPR's advantage over $200$M-frame PPO is significant \emph{within both}
categories (hard/dense $8/9$, $p{=}0.039$; easy $25/35$, $p{=}0.017$), so the headline
paired result is not driven by any one game type.


\section{Complete Theoretical Analysis}
\label{app:theory}

This appendix is self-contained. Section~\ref{app:setup} fixes notation and the
standing assumptions; Section~\ref{app:prelim-lemmas} collects elementary facts used
repeatedly (Gibbs' inequality, a variance bound, McDiarmid's inequality, the
Performance Difference Lemma), each stated precisely and, where short, proved. The
remaining sections prove the results referenced in the main paper. We prioritize
explicitness over brevity: intermediate steps that a reader could reconstruct are
nonetheless written out, so that verification requires no external lookups.

\subsection{Notation and standing assumptions}
\label{app:setup}

\paragraph{Markov decision process.}
We work with an infinite-horizon discounted Markov decision process (MDP)
$\mathcal M=(\St,\A,P,\rew,\gamma,\rho_0)$ where $\St$ is a finite state space,
$\A$ is a \emph{finite} action set with cardinality $|\A|$, $P(\cdot\mid s,a)$ is the
transition kernel, $\rew:\St\times\A\to[-\Rmax,\Rmax]$ is a bounded reward function
with $\Rmax<\infty$, $\gamma\in(0,1)$ is the discount factor, and $\rho_0$ is the
initial-state distribution. A (stationary, Markov) policy $\pi(\cdot\mid s)$ is a
probability distribution on $\A$ for each $s$. We write $\tau=(s_0,a_0,s_1,a_1,\dots)$
for a trajectory generated by $s_0\sim\rho_0$, $a_t\sim\pi(\cdot\mid s_t)$,
$s_{t+1}\sim P(\cdot\mid s_t,a_t)$, and $\E_\pi[\cdot]$ for expectation over this law.

\paragraph{Value, action-value, advantage.}
For a policy $\pi$ define
\begin{align}
V^\pi(s)&=\E_\pi\Big[\textstyle\sum_{k\ge0}\gamma^k
\rew_{t+k}\,\Big|\,s_t=s\Big],\label{app:eq:V}\\
Q^\pi(s,a)&=\rew(s,a)+\gamma
\E_{s'\sim P(\cdot\mid s,a)}[V^\pi(s')],\label{app:eq:Q}\\
\Adv(s,a)&=Q^\pi(s,a)-V^\pi(s).\label{app:eq:A}
\end{align}
Both $V^\pi$ and $Q^\pi$ are well defined and bounded in $[-\Rmax/(1-\gamma),\,\Rmax/(1-\gamma)]$
because the reward is bounded and $\gamma<1$. The objective is
$J(\pi)=\E_{s_0\sim\rho_0}[V^\pi(s_0)]$.

\begin{remark}[Advantages are centered]\label{app:rem:centered}
For every state $s$,
$\sum_a\pi(a\mid s)\Adv(s,a)=\sum_a\pi(a\mid s)Q^\pi(s,a)-V^\pi(s)=V^\pi(s)-V^\pi(s)=0$,
using $\sum_a\pi(a\mid s)Q^\pi(s,a)=V^\pi(s)$ from \eqref{app:eq:V}--\eqref{app:eq:Q}.
We use this centering repeatedly.
\end{remark}

\paragraph{Occupancy and visitation measures.}
The (normalized) discounted state-visitation distribution of $\pi$ is
\begin{equation}\label{app:eq:dpi}
\dpi(s)=(1-\gamma)\sum_{t\ge0}\gamma^t\Pr_\pi(s_t=s),
\end{equation}
a probability distribution on $\St$ (the factor $1-\gamma$ normalizes it). The
discounted \emph{action-occupancy marginal} is the induced distribution on $\A$,
\begin{equation}\label{app:eq:rocc}
\rocc_\pi(a)=(1-\gamma)\sum_{t\ge0}\gamma^t\Pr_\pi(a_t=a)
=\E_{s\sim\dpi}\big[\pi(a\mid s)\big].
\end{equation}
The second equality holds because $\Pr_\pi(a_t=a)=\E[\pi(a\mid s_t)]$ and the
$\gamma$-weighted average of $\Pr_\pi(s_t=\cdot)$ is exactly $\dpi$.

\paragraph{Generalized Advantage Estimation.}
Given a (possibly imperfect) critic $V:\St\to\mathbb R$ and a rollout, the
GAE($\gamma,\lambda$) advantage with $\lambda\in[0,1]$ is
\begin{equation}\label{app:eq:gae}
\begin{aligned}
\hat A_t&=\sum_{l=0}^{\infty}(\gamma\lambda)^l
\delta^{\mathrm{TD}}_{t+l},\\
\delta^{\mathrm{TD}}_t&=\rew_t+\gamma V(s_{t+1})-V(s_t).
\end{aligned}
\end{equation}
where $\rew_t=\rew(s_t,a_t)$. For $\|V\|_\infty\le V_{\max}$, $|\delta^{\mathrm{TD}}_t|\le\Rmax+(1+\gamma)V_{\max}$. Truncating after $L$ terms has error at most $[\Rmax+(1+\gamma)V_{\max}](\gamma\lambda)^L/(1-\gamma\lambda)$. The infinite sum describes a continuing trajectory; in an episodic rollout the sum stops at termination, with terminal bootstrap value zero, and does not propagate across a reset. The finite-segment identities follow by truncating the same sums.

\paragraph{OPR objects (a conditional elite prior).}
Let $\Buf$ be the good-episode buffer: a FIFO ring of the episodes whose undiscounted
return $R(\tau)=\sum_t\rew_t$ lies at or above the $75$th percentile of their
collection batch. We allow the elite prior to be \emph{conditioned on a coarsening of
the state}: fix a map $z:\St\to\{1,\dots,K\}$ assigning each state to one of $K$
clusters (in practice, $k$-means on frozen encoder features), where $K\ge1$ is a
hyperparameter. Writing $N_{\mathrm{good}}(a\mid k)$ for the number of timesteps in the $\Kep{=}10$ highest-return episodes of $\Buf$ in cluster $k$ at which action $a$ was taken, and $N_{\mathrm{buf}}(k)=\sum_a
N_{\mathrm{good}}(a\mid k)$, the \emph{elite conditional prior} is the Laplace-smoothed
empirical distribution
\begin{equation}\label{app:eq:pgood}
\pgood(a\mid k)=
\frac{N_{\mathrm{good}}(a\mid k)+\eta}
{N_{\mathrm{buf}}(k)+\eta|\A|},\quad\eta=0.1 .
\end{equation}
with a matching current prior $\pcurr(a\mid k)$ (an exponential-moving-average estimate
of the policy's per-cluster action frequencies). The \textbf{default and deployed
setting is $K=1$} (a single cluster), for which $z\equiv1$ and \eqref{app:eq:pgood}
reduces to the plain \emph{action marginal} $\pgood(a)$. Theorem~\ref{app:thm:l1}
quantifies the increasing variance cost of $K>1$ but does not make $K=1$ optimal:
Theorem~\ref{app:thm:insuff} characterizes the opposing state-aliasing bias, and
Table~\ref{app:tab:k-bias-variance} measures both effects. Unless a result explicitly varies $K$, statements are written for the marginal
$K=1$ and we drop the ``$\mid k$''. The \emph{raw log-ratio}, its bounded transform, and
the \emph{bonus} at a transition in cluster $k=z(s_t)$ are
\begin{align}
\dratio(a\mid k)&=\log\pgood(a\mid k)-\log\pcurr(a\mid k),\label{app:eq:delta}\\
\sdratio(a\mid k)&=2\tanh(\dratio(a\mid k)/2)\in(-2,2),\label{app:eq:sdelta}\\
\bonus_t&=\ssc\dec\rstd\sdratio(a_t\mid z(s_t)),\label{app:eq:bonus}\\
\dec&=\max(0,1-e/\Edec).\notag
\end{align}
with shaping scale $\ssc>0$, epoch-$e$ decay factor $\dec$, decay horizon $\Edec$, and
$\rstd=\max(B^{-1}\sum_{t=1}^B|r_t|,10^{-4})$ the batch reward scale. The \emph{shaped reward}
is $\srew_t=\rew_t+\bonus_t$, and GAE~\eqref{app:eq:gae} is computed on $\srew_t$.

\begin{lemma}[Closed form: a scaled centered Bayes posterior]
\label{app:lem:posterior}
Write $p=\pgood(a\mid k)$, $q=\pcurr(a\mid k)$, and let
$\mu(a)=p/(p+q)$ be the posterior probability that $a$ was drawn from the elite prior
under a balanced mixture $\tfrac12\pgood+\tfrac12\pcurr$. Then
\begin{equation}\label{app:eq:posterior}
2\tanh\!\big(\dratio(a\mid k)/2\big)
=\frac{2(p-q)}{p+q}=2\big(2\mu(a)-1\big).
\end{equation}
\end{lemma}
\begin{proof}
With $x=p/q>0$ (S1),
\[
\begin{aligned}
\tanh(\tfrac12\log x)
&=\frac{\sqrt{x}-1/\sqrt{x}}{\sqrt{x}+1/\sqrt{x}}\\
&=\frac{x-1}{x+1}=\frac{p-q}{p+q}.
\end{aligned}
\]
Also $(p-q)/(p+q)=2\mu(a)-1$ by definition of $\mu$.
\end{proof}

\begin{remark}[Consequences]\label{app:rem:bbound}
(i) \emph{No explicit clip is needed}: $|2(p-q)/(p+q)|<2$ always, so
the bound $|\sdratio|<2$ --- and hence
\begin{equation}\label{app:eq:bbound}
|\bonus_t|\le 2\ssc\,\dec\,\rstd\quad\text{for all }t
\end{equation}
--- is intrinsic to the transform, not imposed for stability.
(ii) The map $\delta\mapsto2\tanh(\delta/2)$ is not merely an ad-hoc squashing:
$\sdratio(a)=2(2\mu(a)-1)$ is twice the Bayes-optimal
\emph{least-squares} regressor of the $\pm1$ elite label under a balanced class prior
--- the optimal discriminator of least-squares GAN, and the witness function of the
triangular discrimination between $\pgood$ and $\pcurr$ (Eq.~\ref{app:eq:triid}). Together with
Proposition~\ref{app:prop:airl} this says the raw log-ratio $\dratio$ is the optimal
\emph{logistic} discriminator's logit and its smoothed form $\sdratio$ the optimal
\emph{least-squares} discriminator's scaled output: the two scores
are the same Bayes posterior read through two link functions.
(iii) Since $\mu$ is a monotone bijection of $p/q$, pointwise rankings and thresholds on $\sdratio$ correspond to thresholds on the density ratio. This does not equate alignment gates after averaging against signed advantages.
\end{remark}

\paragraph{Standing assumptions.}
Unless a result says otherwise we assume: (S1) $\pcurr(a)>0$ and $\pgood(a)>0$ for all
$a$ (Laplace smoothing gives this for $\pgood$; the empirical-count EMA for $\pcurr$ must be initialized with positive mass and use a rate strictly below one); (S2) rewards are bounded by $\Rmax$; (S3) the discount satisfies
$\gamma\in(0,1)$ and $\lambda\in[0,1]$.

\subsection{Elementary facts used repeatedly}
\label{app:prelim-lemmas}

For self-containedness we first recall three classical inequalities used below, stated
in the finite-dimensional forms we need. \textbf{Jensen's inequality}: for a convex
function $\varphi$ and a random variable $X$, $\E[\varphi(X)]\ge\varphi(\E X)$, with
equality iff $X$ is a.s.\ constant when $\varphi$ is strictly convex.
\textbf{Hölder's inequality} (the case we use): for real vectors $u,v\in\mathbb R^{|\A|}$,
$|\sum_a u_a v_a|\le\|u\|_1\,\|v\|_\infty$ where $\|u\|_1=\sum_a|u_a|$ and
$\|v\|_\infty=\max_a|v_a|$. \textbf{Cauchy--Schwarz} (the case we use):
$\sum_a x_a\le\sqrt{|\A|\sum_a x_a^2}$ for nonnegative $x_a$ (apply the usual
Cauchy--Schwarz to $x$ and the all-ones vector).

\begin{lemma}[Gibbs' inequality]\label{app:lem:gibbs}
For probability distributions $p,q$ on the finite set $\A$ with $q(a)>0$ whenever
$p(a)>0$,
\[
\KL(p\|q)=\sum_a p(a)\log\frac{p(a)}{q(a)}\ge 0,
\]
with equality iff $p=q$.
\end{lemma}
\begin{proof}
Let $f(x)=-\log x$, which is strictly convex on $(0,\infty)$. By Jensen's inequality
applied to the random variable $q(a)/p(a)$ under $a\sim p$ (restricting to the support
of $p$, where $p(a)>0$),
\[
\begin{aligned}
\KL(p\|q)
&=\E_{a\sim p}\Big[-\log\tfrac{q(a)}{p(a)}\Big]\\
&\ge-\log\E_{a\sim p}\Big[\tfrac{q(a)}{p(a)}\Big]\\
&=-\log\!\Big(\sum_{a:\,p(a)>0}q(a)\Big)\ge0 .
\end{aligned}
\]
where the last inequality uses $\sum_{a:\,p(a)>0}q(a)\le\sum_a q(a)=1$. Strict
convexity of $f$ makes Jensen an equality iff $q(a)/p(a)$ is $p$-almost-surely
constant, i.e.\ $q\propto p$ on $\mathrm{supp}(p)$; combined with
$\sum_{a:p(a)>0}q(a)=1$ (needed for the second inequality to be tight) this forces
$q=p$.
\end{proof}

\begin{lemma}[Strict convexity of KL in its first argument]\label{app:lem:klconvex}
The map $p\mapsto\KL(p\|q)$ is strictly convex on the probability simplex
$\Delta(\A)$ for fixed $q$ with full support.
\end{lemma}
\begin{proof}
Write $\KL(p\|q)=\sum_a p(a)\log p(a)-\sum_a p(a)\log q(a)$. The second term is linear
in $p$. The first term is $\sum_a g(p(a))$ with $g(x)=x\log x$, whose second
derivative $g''(x)=1/x>0$ on $(0,1]$; hence $\sum_a g(p(a))$ is strictly convex on the
(relative interior of the) simplex, and adding a linear term preserves strict
convexity.
\end{proof}

\begin{lemma}[Per-coordinate variance of an empirical distribution]\label{app:lem:var}
Let $a_1,\dots,a_n$ be i.i.d.\ draws from a distribution $p$ on $\A$ and let
$\hat p(a)=\frac1n\sum_{i}\mathbf 1[a_i=a]$. Then $\E\hat p(a)=p(a)$ and
$\Var\hat p(a)=p(a)(1-p(a))/n$.
\end{lemma}
\begin{proof}
$n\hat p(a)\sim\mathrm{Binomial}(n,p(a))$, which has mean $np(a)$ and variance
$np(a)(1-p(a))$; divide by $n$ and $n^2$ respectively.
\end{proof}

\begin{lemma}[McDiarmid's bounded-differences inequality]\label{app:lem:mcdiarmid}
Let $X_1,\dots,X_n$ be independent and $f$ a function such that changing a single
coordinate changes $f$ by at most $c_i$: $|f(x)-f(x')|\le c_i$ whenever $x,x'$ differ
only in coordinate $i$. Then for all $\epsilon>0$,
$\Pr\big(f-\E f\ge\epsilon\big)\le\exp\!\big(-2\epsilon^2/\sum_i c_i^2\big)$.
\end{lemma}
\noindent We use Lemma~\ref{app:lem:mcdiarmid} as a standard black box (see, e.g.,
McDiarmid 1989).

\begin{lemma}[Performance Difference Lemma]\label{app:lem:pdl}
For any two policies $\pi,\pi'$,
\begin{equation}\label{app:eq:pdl}
J(\pi')-J(\pi)=\frac{1}{1-\gamma}\,\E_{s\sim d^{\pi'}}\,\E_{a\sim\pi'(\cdot\mid s)}\big[\Adv(s,a)\big].
\end{equation}
\end{lemma}
\begin{proof}
Let $\rew_t=\rew(s_t,a_t)$ along a trajectory of $\pi'$. Using the telescoping identity
$\Adv(s_t,a_t)=\rew_t+\gamma V^\pi(s_{t+1})-V^\pi(s_t)$ (from
\eqref{app:eq:Q}--\eqref{app:eq:A} in expectation over $s_{t+1}$),
\[
\begin{aligned}
&\E_{\pi'}\Big[\sum_{t\ge0}\gamma^t\Adv(s_t,a_t)\Big]\\
&\quad=\E_{\pi'}\Big[\sum_{t\ge0}\gamma^t
\big(\rew_t+\gamma V^\pi(s_{t+1})-V^\pi(s_t)\big)\Big].
\end{aligned}
\]
The $V^\pi$ terms telescope to $-V^\pi(s_0)$; the series converges because
$V^\pi$ is bounded and $\gamma<1$. Hence the right-hand side
equals $\E_{\pi'}[\sum_t\gamma^t\rew_t]-\E_{\pi'}[V^\pi(s_0)]=J(\pi')-J(\pi)$, using
$s_0\sim\rho_0$ for both policies. The left-hand side can be grouped by state:
\[
\E_{\pi'}\Big[\sum_t\gamma^t\Adv(s_t,a_t)\Big]
=\frac{\E_{s\sim d^{\pi'},a\sim\pi'}[\Adv(s,a)]}{1-\gamma},
\]
by the definition \eqref{app:eq:dpi} of $d^{\pi'}$. Equating the two gives
\eqref{app:eq:pdl}.
\end{proof}

\begin{lemma}[Data-processing / contraction of total variation]\label{app:lem:dpi}
Let $\nu_1,\nu_2$ be distributions on a space $\mathcal X$ and let $K$ be a Markov
kernel (channel) from $\mathcal X$ to $\mathcal Y$. Writing $K\nu$ for the pushforward,
$\TV(K\nu_1,K\nu_2)\le\TV(\nu_1,\nu_2)$. In particular, if $\mathcal X=\St\times\A$ and
$K$ is the deterministic projection $(s,a)\mapsto a$, then the action-marginals satisfy
$\TV(\nu_{1,A},\nu_{2,A})\le\TV(\nu_1,\nu_2)$.
\end{lemma}
\begin{proof}
Using the variational form
\[
\TV(\mu_1,\mu_2)=\sup_{\|g\|_\infty\le1}
\frac12\Big|\int g\,d\mu_1-\int g\,d\mu_2\Big|,
\]
for any test function $h$ on $\mathcal Y$ with $\|h\|_\infty\le1$ the pullback
$g=h\circ K$ (more precisely $g(x)=\E_{y\sim K(\cdot\mid x)}[h(y)]$) satisfies
$\|g\|_\infty\le1$ and $\int h\,d(K\nu_i)=\int g\,d\nu_i$. Taking the supremum over $h$
gives $\TV(K\nu_1,K\nu_2)\le\TV(\nu_1,\nu_2)$. The projection is the special case
$K(\cdot\mid s,a)=\delta_a$.
\end{proof}

\subsection{Result 1: raw-KL and deployed bounded-score identities}
\label{app:r1}

\begin{proposition}[Ordered identities for the raw and deployed scores]\label{app:prop:kl}
Fix any cluster $k$ and suppress ``$\mid k$''. Under (S1), the raw log-ratio
\eqref{app:eq:delta} satisfies
\begin{equation}\label{app:eq:klid}
\begin{aligned}
\E_{a\sim\pcurr}[\dratio(a)]
&=\E_{a\sim\pcurr}[\log\pgood(a)-\log\pcurr(a)]\\
&=-\KL(\pcurr\|\pgood)\le0 .
\end{aligned}
\end{equation}
whereas the bounded score actually deployed in \eqref{app:eq:sdelta} satisfies
\begin{equation}\label{app:eq:triid}
\begin{aligned}
\E_{a\sim\pcurr}[\sdratio(a)]
&=-\sum_{a\in\A}
\frac{(\pgood(a)-\pcurr(a))^2}{\pgood(a)+\pcurr(a)}\\
&\le0 .
\end{aligned}
\end{equation}
Both expectations equal zero iff $\pcurr=\pgood$. Equation~\eqref{app:eq:triid} is the
negative triangular discrimination, not a KL divergence.
\end{proposition}
\begin{proof}
By the definition of KL divergence on the finite alphabet $\A$,
\[
\begin{aligned}
\KL(\pcurr\|\pgood)
&=\sum_a\pcurr(a)\log\frac{\pcurr(a)}{\pgood(a)}\\
&=\sum_a\pcurr(a)(\log\pcurr(a)-\log\pgood(a))\\
&=-\E_{a\sim\pcurr}[\dratio(a)] .
\end{aligned}
\]
All logarithms are finite by (S1). The inequality and equality condition are
Lemma~\ref{app:lem:gibbs} with $p=\pcurr$, $q=\pgood$.
For the bounded score, abbreviate $p=\pgood$ and $q=\pcurr$. Pointwise,
\[
2q\frac{p-q}{p+q}=-\frac{(p-q)^2}{p+q}+(p-q).
\]
Summing over $a$ eliminates the second term because $\sum_a(p(a)-q(a))=0$, yielding
\eqref{app:eq:triid}. Every denominator is positive under (S1); hence equality holds
exactly when $p=q$ coordinatewise.
\end{proof}

\begin{corollary}[Ascent on the bonus is descent on the KL; unique fixed point]
\label{app:cor:kl}
Consider the idealized reduced problem in which the policy's action marginal
$\mu\in\Delta(\A)$ is the free variable and the objective is the expected raw bonus
$G(\mu):=\E_{a\sim\mu}[\log\pgood(a)-\log\mu(a)]$ (i.e.\ $\pcurr$ set to the current
$\mu$). Then $G(\mu)=-\KL(\mu\|\pgood)$, so $G$ is strictly concave, is maximized
uniquely at $\mu=\pgood$ with $G(\pgood)=0$, and any step that strictly increases $G$ strictly decreases $\KL(\mu\|\pgood)$.
\end{corollary}
\begin{proof}
$G(\mu)=-\KL(\mu\|\pgood)$ is exactly \eqref{app:eq:klid} with $\pcurr=\mu$.
By Lemma~\ref{app:lem:klconvex} $\mu\mapsto\KL(\mu\|\pgood)$ is strictly convex, so
$G$ is strictly concave; a strictly concave function on the simplex has a unique
maximizer, which by Lemma~\ref{app:lem:gibbs} is $\mu=\pgood$ (value $0$). Since
$G=-\KL$, increasing $G$ decreases $\KL(\mu\|\pgood)$; strict concavity rules out
stationary points other than the maximizer.
\end{proof}

\begin{remark}[Interpretation]
Corollary~\ref{app:cor:kl} concerns the \emph{raw-score reduced problem}. The deployed
bounded score instead has identity \eqref{app:eq:triid}; it shares the unique zero
$\pgood$ but is not an exact KL regularizer. Neither population identity by itself
proves descent for a finite clipped neural PPO update. Thus ``regularization'' denotes
a learning bias toward the elite action distribution, not an exact implemented
regularized-MDP objective.
\end{remark}

\paragraph{Justification of the logarithmic form.}
A natural question is whether the log-ratio $\dratio(a)=\log\pgood(a)-\log\pcurr(a)$ is
an arbitrary choice. Within the separable family below, the logarithm is the unique form up to scale when scores add over independent products. This information-additivity axiom is distinct from merely splitting an action category.

\begin{proposition}[The log-ratio form is forced, within a scoped family]
\label{app:prop:cauchy}
Suppose the per-action bonus is \emph{separable}, $b(a)=g(\pgood(a))-g(\pcurr(a))$ for
some measurable $g:(0,1]\to\mathbb R$, and satisfies \emph{information additivity}: the
score $g$ assigned to the probability of a conjunction of two independent events equals
the sum of the individual scores, $g(xy)=g(x)+g(y)$ for all $x,y\in(0,1]$ (the standard
Shannon-type axiom singling out self-information). Then $g(x)=c\log x$ for some constant
$c\in\mathbb R$.
\end{proposition}
\begin{proof}
Define $h:[0,\infty)\to\mathbb R$ by $h(u)=g(e^{-u})$ (so $u\ge0$ corresponds to
$x=e^{-u}\in(0,1]$). The hypothesis $g(xy)=g(x)+g(y)$ becomes, with $x=e^{-u}$,
$y=e^{-v}$, the additive Cauchy equation $h(u+v)=h(u)+h(v)$ for all $u,v\ge0$. Since
$g$ is measurable, so is $h$. We use the classical fact that \emph{every measurable
solution of the additive Cauchy equation is linear}, $h(u)=-c\,u$ for a constant $c$
(this is Fréchet's theorem; without a regularity condition such as measurability the
equation also admits everywhere-discontinuous solutions built from a Hamel basis of
$\mathbb R$ over $\mathbb Q$, which measurability rules out). Therefore
$g(x)=h(-\log x)=-c(-\log x)=c\log x$.
\end{proof}
\begin{remark}
Proposition~\ref{app:prop:cauchy} justifies the functional \emph{form} of the bonus
within the separable, scale-consistent family; it is not a claim that no non-separable
bonus could be useful. OPR chooses the positive orientation $c>0$, whose magnitude is absorbed into $\ssc$; the axiom alone does not fix that sign.
\end{remark}
\begin{remark}[Why a ratio rather than a difference]
There is also a practical reason to prefer the log-\emph{ratio} over a probability
\emph{difference} $\pgood(a)-\pcurr(a)$. At a fixed probability ratio, the difference shrinks with the absolute probabilities, whereas the log-ratio does not. The latter therefore gives the same score to common and rare actions with the same relative elite enrichment. This scale invariance does not remove estimation noise, which can be substantial for rare actions; smoothing and the bounded transform limit its effect on the reward.
\end{remark}

\subsection{Result 2: pre-advantage injection propagates densely}
\label{app:r2}

This is the mechanism that distinguishes OPR from post-hoc imitation losses. We first
give the exact statement under a fixed critic, then bound the effect of critic drift.

\begin{assumption}[Fixed critic within one GAE computation]\label{app:ass:fixedV}
When we compare the shaped advantage $\hat A_t^{\mathrm{OPR}}$ to the unshaped
$\hat A_t$, both use the same critic $V$; i.e.\ we measure the instantaneous effect of
adding $\{\bonus_t\}$ to the reward stream inside a single evaluation of
\eqref{app:eq:gae}.
\end{assumption}

\begin{lemma}[Exact shaped-advantage decomposition]\label{app:lem:gae}
Under Assumption~\ref{app:ass:fixedV}, with shaped reward $\srew_t=\rew_t+\bonus_t$ and fixed $\dec,\rstd$ over the compared segment,
\begin{equation}\label{app:eq:decomp}
\hat A_t^{\mathrm{OPR}}=\hat A_t+B_t,\qquad
B_t:=\sum_{l=0}^{\infty}(\gamma\lambda)^l\,\bonus_{t+l},
\end{equation}
and $B_t$ converges absolutely with $|B_t|\le 2\ssc\dec\rstd/(1-\gamma\lambda)$.
\end{lemma}
\begin{proof}
With $V$ fixed, the shaped TD residual is
\[
\tilde\delta^{\mathrm{TD}}_t=\srew_t+\gamma V(s_{t+1})-V(s_t)
=\big(\rew_t+\gamma V(s_{t+1})-V(s_t)\big)+\bonus_t=\delta^{\mathrm{TD}}_t+\bonus_t.
\]
Substituting into \eqref{app:eq:gae} and using linearity of the (absolutely
convergent) sum,
\[
\hat A_t^{\mathrm{OPR}}=\sum_{l\ge0}(\gamma\lambda)^l\tilde\delta^{\mathrm{TD}}_{t+l}
=\sum_{l\ge0}(\gamma\lambda)^l\delta^{\mathrm{TD}}_{t+l}
+\sum_{l\ge0}(\gamma\lambda)^l\bonus_{t+l}=\hat A_t+B_t.
\]
Absolute convergence of $B_t$ and its bound follow from \eqref{app:eq:bbound}:
$|B_t|\le\sum_{l\ge0}(\gamma\lambda)^l\,2\ssc\dec\rstd=2\ssc\dec\rstd/(1-\gamma\lambda)$.
\end{proof}

\begin{corollary}[Density: one bonus enters many advantages]\label{app:cor:density}
Reindexing \eqref{app:eq:decomp}, a single bonus $\bonus_\tau$ appears in the shaped
advantage of every earlier step $t\le\tau$ of the same rollout, with weight
$(\gamma\lambda)^{\tau-t}$. Its total downstream influence is
\begin{equation}
\sum_{t=0}^{\tau}(\gamma\lambda)^{\tau-t}=\frac{1-(\gamma\lambda)^{\tau+1}}{1-\gamma\lambda}
\xrightarrow[\tau\to\infty]{}\frac{1}{1-\gamma\lambda}.
\end{equation}
\end{corollary}
\begin{proof}
By \eqref{app:eq:decomp}, $\bonus_\tau$ is the $l=\tau-t$ term of $B_t$, present iff
$\tau\ge t$, with coefficient $(\gamma\lambda)^{\tau-t}$. Summing this coefficient over
$t=0,\dots,\tau$ is a finite geometric series with ratio $\gamma\lambda\in[0,1)$, giving
the stated value; the limit is the geometric-series sum.
\end{proof}

\begin{lemma}[Robustness to critic drift]\label{app:lem:drift}
Let $V,V'$ be two critics with $\|V-V'\|_\infty:=\sup_s|V(s)-V'(s)|=\epsilon_V$, and
let $\hat A_t^{\mathrm{OPR}}[V]$ denote the shaped advantage computed with critic $V$.
Then
\begin{equation}
\big|\hat A_t^{\mathrm{OPR}}[V']-\big(\hat A_t[V]+B_t\big)\big|\le\frac{(1+\gamma)\,\epsilon_V}{1-\gamma\lambda}.
\end{equation}
\end{lemma}
\begin{proof}
By Lemma~\ref{app:lem:gae}, $\hat A_t^{\mathrm{OPR}}[V']=\hat A_t[V']+B_t$ (the bonus
term $B_t$ does not depend on the critic). Thus the left-hand side equals
$|\hat A_t[V']-\hat A_t[V]|$. Each TD residual differs by
$\delta^{\mathrm{TD}}_t[V']-\delta^{\mathrm{TD}}_t[V]
=\gamma\big(V'(s_{t+1})-V(s_{t+1})\big)-\big(V'(s_t)-V(s_t)\big)$,
so $|\delta^{\mathrm{TD}}_t[V']-\delta^{\mathrm{TD}}_t[V]|\le(1+\gamma)\epsilon_V$.
Summing with weights $(\gamma\lambda)^l$ multiplies by $\sum_l(\gamma\lambda)^l=1/(1-\gamma\lambda)$.
\end{proof}

\begin{remark}[Contrast with SIL, quantified]\label{app:rem:silcount}
Self-imitation applies the gradient of
$\mathcal L_{\mathrm{SIL}}=-\E[\log\pi(a\mid s)(R-V(s))_+]$ to \emph{sampled replayed}
transitions only, with no analogue of $B_t$: the replayed signal is not propagated
through \eqref{app:eq:gae}. Per collection, OPR can shape all $T\times N$ rollout transitions; a bonus enters earlier advantages with total weight approaching $1/(1-\gamma\lambda)$ on long uninterrupted segments. SIL instead samples $n_{\mathrm{steps}}\times|\text{batch}|$ replay items across its inner updates, which need not be distinct. The controlled comparison in main-paper \S\ref{sec:controlled} tests reward-channel placement with matched information; the decomposition itself does not predict a numerical performance gap.
\end{remark}

\subsection{Result 3: statistical cost of estimating the prior}
\label{app:r3}

\begin{assumption}[Effective i.i.d.\ samples]\label{app:ass:iid}
The $\Nbuf$ action samples in the selected elite episodes are treated as i.i.d.\ draws from the population
elite marginal $\pgood$. Within an episode actions are temporally correlated;
Remark~\ref{app:rem:eff} explains how to read $\Nbuf$ as an effective sample size.
\end{assumption}

\begin{theorem}[Expected $L_1$ error of the marginal]\label{app:thm:l1}
Under Assumption~\ref{app:ass:iid}, the unsmoothed empirical marginal
$\hat\pgood(a)=N_{\mathrm{good}}(a)/\Nbuf$ satisfies
\begin{equation}\label{app:eq:l1}
\E\big\|\hat\pgood-\pgood\big\|_1\le\sqrt{\frac{|\A|-1}{\Nbuf}}=O\!\Big(\sqrt{|\A|/\Nbuf}\Big).
\end{equation}
For a state-clustered prior with $K$ clusters of comparable occupancy
($\approx\Nbuf/K$ samples each), the per-cluster estimate obeys
$\E\|\hat\pgood(\cdot\mid z)-\pgood(\cdot\mid z)\|_1=O(\sqrt{K|\A|/\Nbuf})$.
\end{theorem}
\begin{proof}
Write $n=\Nbuf$. By Jensen's inequality ($\E|X|\le\sqrt{\E X^2}$ applied to the
mean-zero $\hat\pgood(a)-\pgood(a)$) and Lemma~\ref{app:lem:var},
\[
\begin{aligned}
\E\|\hat\pgood-\pgood\|_1
&=\sum_a\E|\hat\pgood(a)-\pgood(a)|\\
&\le\sum_a\sqrt{\Var\hat\pgood(a)}\\
&=\frac{1}{\sqrt n}\sum_a\sqrt{\pgood(a)(1-\pgood(a))}.
\end{aligned}
\]
By the Cauchy--Schwarz inequality with the all-ones vector,
\[
\begin{aligned}
\sum_a\sqrt{\pgood(a)(1-\pgood(a))}
&\le\sqrt{|\A|\sum_a\pgood(a)(1-\pgood(a))}\\
&=\sqrt{|\A|\big(1-\textstyle\sum_a\pgood(a)^2\big)} .
\end{aligned}
\]
Finally $\sum_a\pgood(a)^2\ge 1/|\A|$ (by Cauchy--Schwarz, or because the uniform
distribution minimizes the $\ell_2$ norm on the simplex), so
$|\A|(1-\sum_a\pgood(a)^2)\le|\A|-1$, giving \eqref{app:eq:l1}. The clustered bound
replaces $n$ by $\Nbuf/K$ in the per-cluster application.
\end{proof}

\begin{theorem}[High-probability version]\label{app:thm:bhc}
Under Assumption~\ref{app:ass:iid}, for every $\epsilon>0$,
\[
\Pr\Big(\|\hat\pgood-\pgood\|_1\ge\E\|\hat\pgood-\pgood\|_1+\epsilon\Big)\le e^{-\Nbuf\epsilon^2/2}.
\]
\end{theorem}
\begin{proof}
View $f(a_1,\dots,a_n)=\|\hat\pgood-\pgood\|_1$ as a function of the $n=\Nbuf$ i.i.d.\
samples. Changing one sample $a_i$ moves one unit of empirical mass from one bin to
another, changing each of two coordinates of $\hat\pgood$ by $1/n$, hence changing
$f$ by at most $2/n$ in $\ell_1$; so the bounded-difference constants are $c_i=2/n$.
Lemma~\ref{app:lem:mcdiarmid} gives
$\Pr(f\ge\E f+\epsilon)\le\exp\!\big(-2\epsilon^2/\sum_i(2/n)^2\big)=\exp(-2\epsilon^2/(4/n))=\exp(-n\epsilon^2/2)$.
\end{proof}

\begin{remark}[Illustrative sample-size calculation]\label{app:rem:l1num}
At the illustrative i.i.d.\ sample size $\Nbuf=3\times10^5$ and $|\A|=18$, \eqref{app:eq:l1} gives an upper bound
$\approx\sqrt{17/(3\times10^5)}\approx0.0075$ for the marginal ($K=1$); the $K=16$
clustered prior inflates this bound by $\sqrt{16}=4\times$ under comparable occupancy. This quantifies a possible estimation cost, not a measured Atari error: $\Nbuf$ counts the selected top episodes, not all buffered transitions. It does not predict return or prove that $K{=}1$ is
optimal: conditioning can also reduce structural bias, as
Table~\ref{app:tab:k-bias-variance} demonstrates.
Laplace smoothing changes $\hat\pgood$ by $O(|\A|/\Nbuf)$ in $\ell_1$, which is
negligible relative to \eqref{app:eq:l1} and does not affect the rate.
\end{remark}

\begin{remark}[On Assumption~\ref{app:ass:iid}]\label{app:rem:eff}
If within-episode actions have integrated autocorrelation time $\kappa$, a heuristic
effective-sample-size correction replaces $\Nbuf$ by $\Nbuf/\kappa$ and inflates the
scale in \eqref{app:eq:l1} by $\sqrt{\kappa}$. This preserves the displayed
$\sqrt K$ comparison only under a common-$\kappa$ model; cluster-specific dependence
can change the relative ordering. The i.i.d. concentration statement itself does not
automatically extend to the replay-buffer samples.
\end{remark}

\paragraph{Controlled $K$ protocol.}
For each cell below, states are sampled uniformly and the elite action is correct
with probability $0.95$. The marginal and conditional estimators use the \emph{same}
sampled state--action pairs and Laplace smoothing $0.1$. From a uniform policy we
apply eight statewise KL-proximal updates with $\eta=0.2$ using only the bounded OPR
score, then report the mean probability of the correct action across states. This
deliberately isolates prior aliasing and estimation variance; it is not an MDP
learning curve with a task-reward gradient.

\begin{table}[t]
\centering
\small
\begin{tabular}{rcc}
\toprule
Elite samples & State-aliased & Shared action \\
 & $K{=}S-K{=}1$ reward & $K{=}S-K{=}1$ reward \\
\midrule
$8$     & $+0.137\pm0.003$ & $-0.263\pm0.006$ \\
$32$    & $+0.320\pm0.004$ & $-0.088\pm0.003$ \\
$128$   & $+0.400\pm0.002$ & $-0.0082\pm0.0007$ \\
$1{,}024$ & $+0.408\pm0.001$ & $-0.00048\pm0.00009$ \\
\bottomrule
\end{tabular}
\caption{Paired bias--variance check for $K{=}1$ (marginal) versus $K{=}S$
(true-state conditioning). Each cell is the paired reward difference after eight
KL-proximal bonus updates, mean $\pm$ 95\% CI over 400 independently sampled elite
datasets. In 16 equally likely states, the correct action either alternates by state
(state-aliased) or is shared. Conditioning removes structural bias in the former but
fragments scarce data in the latter. Eight sample sizes from 8 to 1,024 were run; four
are shown. This controlled result demonstrates a tradeoff, not that true-state labels
are available to Atari's learned clusters.}
\label{app:tab:k-bias-variance}
\end{table}

\subsection{Result 4: a value-gate comparison with SIL}
\label{app:r4}

\begin{corollary}[Marginalization contracts policy drift]\label{app:cor:drift}
Let $\mu^{(t)},\mu^{(t')}$ be the state--action occupancy measures of the policy at two
training times, and $\mu^{(t)}_A,\mu^{(t')}_A$ their action-marginals. Then
$\TV(\mu^{(t)}_A,\mu^{(t')}_A)\le\TV(\mu^{(t)},\mu^{(t')})$.
\end{corollary}
\begin{proof}
Immediate from Lemma~\ref{app:lem:dpi} with the projection $K:(s,a)\mapsto a$.
\end{proof}
\noindent Projecting the same two occupancy measures to action marginals cannot increase total variation. This does not compare elite-selected buffers with SIL replay distributions or bound the drift of each normalized conditional prior.

\begin{lemma}[Value-gate sign error in SIL]\label{app:lem:gate}
SIL includes a replayed transition with indicator $\mathbf 1\{R>\hat V(s)\}$. It differs from the exact-value indicator $\mathbf 1\{R>V^\pi(s)\}$ exactly when
\[
\min\{\hat V(s),V^\pi(s)\}<R\le\max\{\hat V(s),V^\pi(s)\}.
\]
If neither value equals $R$, this is equivalent to $R$ lying strictly between the two values and requires $|\hat V(s)-V^\pi(s)|>|R-V^\pi(s)|$. Including equality cases, disagreement requires the weak inequality $\ge$; neither magnitude condition alone is sufficient.
\end{lemma}
\begin{proof}
If $\hat V(s)<V^\pi(s)$, the indicators differ on $\hat V(s)<R\le V^\pi(s)$. If $\hat V(s)>V^\pi(s)$, they differ on $V^\pi(s)<R\le\hat V(s)$. They coincide when the values are equal. These two intervals give the result and its magnitude bounds. A realized return above $V^\pi(s)$ need not imply that the action has positive expected advantage.
\end{proof}

\noindent Ungated OPR has no value-based inclusion gate, so the particular sign-error
mechanism in Lemma~\ref{app:lem:gate} does not apply to it. This is not an overall
robustness guarantee: OPR still has prior-estimation, state-aliasing, selection, and
optimization errors, while the online gated extension deliberately uses a
learned-critic statistic and therefore reintroduces critic dependence.

\begin{proposition}[Regime comparison --- informal]\label{app:prop:regime}
Model the per-update policy-gradient error of each method by an upper bound of the form
$\mathcal E_{\mathrm{OPR}}\le C_1\sqrt{|\A|/\Nbuf}$ (from Theorem~\ref{app:thm:l1},
propagated through a Lipschitz policy-gradient map) and
$\mathcal E_{\mathrm{SIL}}\le C_2\,\TV(\mu^{(t)},\mu^{(t')})+C_3\,\epsilon_V$ (staleness,
controlled in the relevant part by Corollary~\ref{app:cor:drift}, plus gate error from
Lemma~\ref{app:lem:gate}). Then the OPR bound is smaller when
$C_1\sqrt{|\A|/\Nbuf}\ll C_2\TV(\mu^{(t)},\mu^{(t')})+C_3\epsilon_V$.
\end{proposition}
\begin{remark}
We label Proposition~\ref{app:prop:regime} \emph{informal}: it compares two upper
bounds, not the true errors, and the constants $C_1,C_2,C_3$ (absorbing the
policy-gradient Lipschitz constant) are not estimated. These error bounds are modeling assumptions, not consequences of the preceding lemmas alone: a linear critic-error bound on inclusion mistakes requires control of probability mass near $R=V^\pi(s)$, and translating prior error requires a suitably regular update map. The rigorous content is
Corollary~\ref{app:cor:drift} and Lemma~\ref{app:lem:gate}; the proposition only
organizes the qualitative reading (OPR favored under fast drift / noisy critic / large
buffer, SIL favored under an accurate critic and a near-stationary policy).
\end{remark}

\subsection{Result 5: decay removes the instantaneous perturbation}
\label{app:r5}

\begin{theorem}[Vanishing instantaneous perturbation]\label{app:thm:bias}
With $\dec=\max(0,1-e/\Edec)$ and $\Edec<\infty$, the per-step reward perturbation
obeys $|\srew_t-\rew_t|=|\bonus_t|\le2\ssc\dec\rstd$ and equals $0$ for all $e\ge\Edec$. The bound is nonincreasing if $\rstd$ is held fixed. With a varying batch scale, $\rstd\le\max\{\Rmax,10^{-4}\}$ instead gives the nonincreasing envelope $2\ssc\dec\max\{\Rmax,10^{-4}\}$. Consequently, for any fixed policy and
$e\ge\Edec$, its shaped return equals its true return. This concerns future updates
after decay completes; it does not undo earlier shaped updates or recover the
parameters of an independently trained unshaped agent.
\end{theorem}
\begin{proof}
The bound is Remark~\ref{app:rem:bbound}. The factor $\dec$ is nonincreasing in $e$ and
hits $0$ at $e=\Edec$. For $e\ge\Edec$, $\dec=0\Rightarrow\bonus_t\equiv0\Rightarrow
\srew_t=\rew_t$; hence the discounted shaped return $\sum_t\gamma^t\srew_t$ equals the
true return $\sum_t\gamma^t\rew_t$ sample-path-wise for the fixed policy. Nothing in
this calculation reverses parameter changes made before $e=\Edec$.
\end{proof}

\begin{proposition}[Constant-shaping bias]\label{app:prop:simlemma}
Fix the epoch, priors, and reward scale, making $\bonus(s,a)$ a common bounded reward
for all policies. Suppress $\dec$ (constant shaping, $\dec\equiv1$). Let $\pi^\star$ and
$\tilde\pi^\star$ maximize the true objective $J$ and the shaped objective $\tilde J$,
respectively. Then
\begin{equation}
\begin{aligned}
0\le J(\pi^\star)-J(\tilde\pi^\star)
&\le\frac{2\|\bonus\|_\infty}{1-\gamma}
\le\frac{4\ssc\rstd}{1-\gamma},\\
\|\bonus\|_\infty&:=\max_{s,a}|\bonus(s,a)|.
\end{aligned}
\end{equation}
\end{proposition}
\begin{proof}
For any fixed policy $\pi$, its true and shaped values differ by the discounted sum of
the bonus, so $|\tilde J(\pi)-J(\pi)|\le\sum_{k\ge0}\gamma^k\|\bonus\|_\infty=\|\bonus\|_\infty/(1-\gamma)$.
Decompose
\[
\begin{aligned}
J(\pi^\star)-J(\tilde\pi^\star)
={}&\big(J(\pi^\star)-\tilde J(\pi^\star)\big)\\
&+\big(\tilde J(\pi^\star)-\tilde J(\tilde\pi^\star)\big)\\
&+\big(\tilde J(\tilde\pi^\star)-J(\tilde\pi^\star)\big),
\end{aligned}
\]
where the middle term is $\le0$ because $\tilde\pi^\star$ maximizes $\tilde J$. This
gives the upper bound; the lower bound $0$ holds because $\pi^\star$ maximizes $J$.
Finally $\|\bonus\|_\infty\le2\ssc\rstd$ by Remark~\ref{app:rem:bbound} with $\dec=1$.
\end{proof}
\begin{remark}[OPR is not potential-based]
The OPR bonus is not constrained to have the potential form $\gamma\pot(s_{t+1})-\pot(s_t)$. For example, in a one-state self-loop MDP an action-dependent nonconstant bonus cannot have that form, since a state potential would give the same bonus to every action. Special cases such as the zero bonus can be potential-based; in general Ng et al.'s policy-invariance theorem does not apply, and constant shaping can shift the optimum. Proposition~\ref{app:prop:simlemma} bounds that shift, and
Theorem~\ref{app:thm:bias} removes the instantaneous perturbation once decay completes,
without erasing the earlier training path. The constant-shaping ablation is close to
the decayed run at the tested seed, but does not establish general invariance.
\end{remark}

\subsection{T1: action-ratio connections to adversarial imitation}
\label{app:t1}

We now relate the raw bonus to a Bayes classifier on action marginals. This connection
is weaker than state--action or trajectory-level adversarial imitation. Recall $\rocc_\pi$ from \eqref{app:eq:rocc};
write $\rocc_{\mathrm{good}}$ for an elite population action distribution under a specified sampling rule. The classifier identity below holds for any two such distributions. Undiscounted episode counts, top-episode selection, and a fixed-rate EMA need not consistently estimate discounted stationary-policy occupancies; Theorem~\ref{app:thm:l1} applies only under its stated i.i.d.\ sampling model.

\begin{proposition}[Action-marginal Bayes log odds]\label{app:prop:airl}
Consider the two-class problem of distinguishing an action drawn from the elite
occupancy $\rocc_{\mathrm{good}}$ (class $1$) from one drawn from the policy occupancy
$\rocc_\pi$ (class $0$), with equal class priors. The Bayes-optimal classifier is
\begin{equation}\label{app:eq:Dstar}
D^\star(a)=\frac{\rocc_{\mathrm{good}}(a)}{\rocc_{\mathrm{good}}(a)+\rocc_\pi(a)},
\end{equation}
and its log-odds equals the OPR raw bonus on the population marginals:
\begin{equation}\label{app:eq:airl}
\begin{aligned}
\log\frac{D^\star(a)}{1-D^\star(a)}
&=\log\frac{\rocc_{\mathrm{good}}(a)}{\rocc_\pi(a)}\\
&\xrightarrow{\ \text{estimators}\ }
\log\pgood(a)-\log\pcurr(a)=\dratio(a).
\end{aligned}
\end{equation}
Thus $\dratio$ has the same log-odds form used by adversarial imitation objectives,
after both elite and policy behavior are projected to action marginals. This does not
identify it with the state--action discriminator or learned reward of GAIL/AIRL
\citep{ho2016gail,fu2018airl}.
\end{proposition}
\begin{proof}
With equal priors $\tfrac12$ and class-conditional densities
$\rocc_{\mathrm{good}},\rocc_\pi$, Bayes' rule gives the posterior probability of class
$1$ as $D^\star(a)=\frac{\tfrac12\rocc_{\mathrm{good}}(a)}{\tfrac12\rocc_{\mathrm{good}}(a)+\tfrac12\rocc_\pi(a)}$,
which is \eqref{app:eq:Dstar}. Then $1-D^\star(a)=\rocc_\pi(a)/(\rocc_{\mathrm{good}}(a)+\rocc_\pi(a))$,
so $D^\star/(1-D^\star)=\rocc_{\mathrm{good}}(a)/\rocc_\pi(a)$ and its log is the stated
density ratio. 
Substituting the histograms $\pgood,\pcurr$ gives the corresponding empirical log ratio $\dratio$; convergence additionally requires consistent estimates of these same population distributions.
\end{proof}
\begin{remark}[Why this matters]
GAIL/AIRL train a discriminator by an inner adversarial optimization; for the much
coarser two-sample problem on action marginals, the Bayes classifier is available in
closed form as a histogram ratio. OPR therefore shares their density-ratio structure
without claiming trajectory-level imitation. The BC-shaping ablation adds a learned
prior, but its observed underperformance does not by itself identify optimization as
the cause.
\end{remark}

\begin{theorem}[Raw-score reduced objective and descent direction]\label{app:thm:fdiv}
Fix a full-support elite distribution $p=\pgood$ and let the action distribution $x$ vary freely over the simplex. The self-consistent objective $G(x)=\sum_a x(a)\log[p(a)/x(a)]=-\KL(x\|p)$ has unique maximizer $x=p$ and no other interior stationary point. At a full-support reference $q$, the frozen score $\log(p/q)$ is the negative gradient of $\KL(x\|p)$ at $x=q$, up to an action-independent constant. If $q\ne p$, projecting this vector onto the simplex tangent space gives a strict KL-descent direction: every sufficiently small feasible positive step, or an exact line search \emph{on KL} over a feasible segment in that direction, decreases KL. This statement concerns a free action distribution, not constrained MDP occupancies or a clipped neural PPO step.
\end{theorem}
\begin{proof}
Strict convexity and Gibbs' inequality give the unique maximizer. At $q$,
\[
\nabla_x\KL(x\|p)\big|_{q}=\log(q/p)+\mathbf 1.
\]
Let $\mathcal T=\{v:\sum_a v(a)=0\}$ and $v=\operatorname{proj}_{\mathcal T}\log(p/q)$. Then $\nabla\KL(q\|p)^\top v=-\|v\|_2^2$. This is strictly negative unless $\log(p/q)$ is constant across actions, which forces $p=q$ after normalization. Since $q$ is interior, $q+tv$ is feasible for sufficiently small $t>0$; differentiability then gives strict decrease. An exact minimization of KL on a feasible segment containing these small steps cannot do worse. Maximizing the frozen \emph{linear} score along that segment is not the same line search and can overshoot the KL minimum. If action distributions are constrained by an MDP or parametrization, $p$ may be unattainable and this free-simplex conclusion need not hold.
\end{proof}

\begin{remark}[Elite selection and advantage weighting]
Reward-weighted regression and related methods fit targets such as $q^\star(a\mid s)\propto\pi(a\mid s)\exp(\Adv(s,a)/\zeta)$ for temperature $\zeta$. OPR's return-quantile filter likewise favors high-return data, but it selects entire episodes and then aggregates actions. It is therefore a hard-selection analogue, not in general an estimator of that statewise advantage-weighted target, and varying a return quantile is not identical to varying $\zeta$. The useful connection is the shared use of reward-favored action distributions and density ratios, alongside the action-level adversarial interpretation of Proposition~\ref{app:prop:airl}.
\end{remark}

\subsection{T2: an assumed-direction checkpoint diagnostic}
\label{app:t2}

\begin{definition}[Prior--advantage alignment]\label{app:def:phi}
For a game (MDP) $g$ with current policy $\pi$, define
\begin{equation}\label{app:eq:phi}
\Align(g):=\E_{s\sim\dpi}\Big[\textstyle\sum_a\pgood(a)\,\Adv(s,a)\Big].
\end{equation}
By Remark~\ref{app:rem:centered}, $\sum_a\pi(a\mid s)\Adv(s,a)=0$, so equivalently
$\Align(g)=\E_{s\sim\dpi}\big[\sum_a(\pgood(a)-\pi(a\mid s))\Adv(s,a)\big]$: it measures
how much better, under the current advantage, the elite-marginal action mix is than the
policy's own mix.
\end{definition}

\begin{theorem}[Gain under an assumed policy direction]\label{app:thm:gain}
Let $\pi_\beta$ be the policy obtained from one OPR-shaped update at scale $\ssc$, and
suppose the update moves the action distribution proportionally toward $\pgood$:
\begin{equation}\label{app:eq:linstep}
\pi_\beta(\cdot\mid s)=\pi(\cdot\mid s)+\ssc\kappa\big(\pgood-\pi(\cdot\mid s)\big)+o(\ssc)
\end{equation}
for a step constant $\kappa>0$, with the $o(\ssc)$ remainder uniform over states in $L_1$. This direction is an explicit modeling assumption,
not a derived linearization of clipped PPO.
Assume the visitation map $\pi\mapsto\dpi$ is Lipschitz in total variation. Then
\begin{equation}\label{app:eq:gain}
\begin{aligned}
J(\pi_\beta)-J(\pi)
&=\frac{\ssc\kappa}{1-\gamma}\Align(g)+o(\ssc)
 +O(\ssc\varepsilon_{\mathrm{est}}),\\
\varepsilon_{\mathrm{est}}&=\|\hat\pgood-\pgood\|_1 .
\end{aligned}
\end{equation}
so, under this assumed direction and when the leading term dominates,
$\sign\big(J(\pi_\beta)-J(\pi)\big)=\sign\big(\Align(g)\big)$.
The same statement holds for a state-conditioned prior after replacing
$\pgood(a)$ by $\pgood(a\mid s)$ in \eqref{app:eq:linstep} and in the definition of
$\Align$; the proof is pointwise in $s$.
\end{theorem}
\begin{proof}
Apply the Performance Difference Lemma~\ref{app:lem:pdl} with $\pi'=\pi_\beta$:
\begin{equation}\label{app:eq:pdlstep}
\begin{aligned}
J(\pi_\beta)-J(\pi)
=\frac{1}{1-\gamma}\E_{s\sim d^{\pi_\beta}}
\Big[\E_{a\sim\pi_\beta(\cdot\mid s)}[\Adv(s,a)]\Big].
\end{aligned}
\end{equation}
We expand the right-hand side to first order in $\ssc$. First, by
\eqref{app:eq:linstep},
\[
\begin{aligned}
\E_{a\sim\pi_\beta(\cdot\mid s)}[\Adv(s,a)]
={}&\sum_a\big(\pi(a\mid s)\\
&+\ssc\kappa(\pgood(a)-\pi(a\mid s))\big)\Adv(s,a)
+o(\ssc).
\end{aligned}
\]
The zeroth-order term $\sum_a\pi(a\mid s)\Adv(s,a)=0$ by Remark~\ref{app:rem:centered}.
Hence the inner expectation is
$\ssc\kappa\sum_a(\pgood(a)-\pi(a\mid s))\Adv(s,a)+o(\ssc)$, an $O(\ssc)$ quantity.
Second, $d^{\pi_\beta}=\dpi+O(\ssc)$ in total variation (visitation Lipschitzness and
$\|\pi_\beta-\pi\|=O(\ssc)$). Because the inner expectation is already $O(\ssc)$,
replacing $d^{\pi_\beta}$ by $\dpi$ incurs only an $O(\ssc)\cdot O(\ssc)=O(\ssc^2)$
error. The assumed policy remainder still contributes $o(\ssc)$, so \eqref{app:eq:pdlstep} equals
\[
\begin{aligned}
&\frac{\ssc\kappa}{1-\gamma}\E_{s\sim\dpi}
\Big[\sum_a(\pgood(a)-\pi(a\mid s))\Adv(s,a)\Big]+o(\ssc)\\
&\hspace{2em}=\frac{\ssc\kappa}{1-\gamma}\Align(g)+o(\ssc),
\end{aligned}
\]
using Definition~\ref{app:def:phi}. Finally, the update actually uses the estimator
$\hat\pgood$; replacing $\pgood$ by $\hat\pgood$ changes the bracket by
$\sum_a(\hat\pgood(a)-\pgood(a))\Adv(s,a)$, bounded in absolute value by
$\|\Adv(s,\cdot)\|_\infty\|\hat\pgood-\pgood\|_1$ (Hölder), contributing the
$O(\ssc\,\varepsilon_{\mathrm{est}})$ term with $\|\Adv\|_\infty\le2\Rmax/(1-\gamma)$
absorbed into the constant. The sign statement follows because the leading term
dominates for small $\ssc$ when $\Align(g)\ne0$ and $\varepsilon_{\mathrm{est}}$ is
small (Theorem~\ref{app:thm:l1}). For a conditional prior, use $\varepsilon_{\mathrm{est}}=\max_k\|\hat\pgood(\cdot\mid k)-\pgood(\cdot\mid k)\|_1$. An $O(\ssc^2)$ return remainder requires an $O(\ssc^2)$ policy remainder in \eqref{app:eq:linstep}; it does not follow from $o(\ssc)$ alone.
\end{proof}

\begin{remark}[A zero bonus is distinct from an assumed prior-directed step]\label{app:rem:kappa}
If $\pgood=\pcurr$, the shaping score is zero, so shaped and raw updates from the same parameters and data coincide. The raw learning update may still change the policy. Equation~\eqref{app:eq:linstep} with $\kappa>0$ is not then a model of the bonus unless its prior-directed term also vanishes; using $\kappa=0$ removes that leading term but an $o(\ssc)$ remainder alone does not assert an exactly null update. In particular, $\Align$ may be nonzero even when the marginal shaping signal vanishes, as Theorem~\ref{app:thm:insuff} illustrates.
\end{remark}

\begin{corollary}[Three regimes within the assumed-direction model]\label{app:cor:regimes}
Within Theorem~\ref{app:thm:gain}'s assumed direction, the sign of $\Align(g)$ gives GAIN if
$\Align(g)>0$, HURT if $\Align(g)<0$, and FLAT if $\Align(g)\approx0$. Moreover
\begin{equation}\label{app:eq:phibound}
|\Align(g)|\le\|\Adv\|_\infty\;\E_{s\sim\dpi}\big\|\pgood-\pi(\cdot\mid s)\big\|_1,
\end{equation}
so a small average \emph{statewise} prior--policy gap makes the coefficient small for a fixed advantage bound. A small gap between action marginals alone does not bound this statewise distance.
\end{corollary}
\begin{proof}
The regime split is the sign statement of Theorem~\ref{app:thm:gain}. For
\eqref{app:eq:phibound}, use the centered form of $\Align(g)$
(Definition~\ref{app:def:phi}) and apply Hölder's inequality inside the expectation:
$|\sum_a(\pgood(a)-\pi(a\mid s))\Adv(s,a)|\le\|\Adv(s,\cdot)\|_\infty\|\pgood-\pi(\cdot\mid s)\|_1$,
then take expectations over $s\sim\dpi$ and bound $\|\Adv(s,\cdot)\|_\infty\le\|\Adv\|_\infty$.
\end{proof}

\begin{corollary}[Oracle gate for the idealized bonus-only update]\label{app:cor:gated}
Define the \emph{$\Align$-gated} local update that applies the idealized bonus step
of Theorem~\ref{app:thm:gain} when $\Align(g)>0$ and otherwise applies the identity
map (no bonus step). Under the hypotheses of that theorem, its change relative to
the current policy is
\begin{equation}\label{app:eq:gated}
\begin{aligned}
J(\pi_{\mathrm{gate}})-J(\pi)
&=\frac{\ssc\kappa}{1-\gamma}\Align(g)^{+}
+o(\ssc)+O(\ssc\varepsilon_{\mathrm{est}}),\\
\Align(g)^{+}&:=\max\{\Align(g),0\}\ge0 .
\end{aligned}
\end{equation}
Thus the leading term of this idealized, oracle-gated \emph{bonus contribution} is
nonnegative. This is not a comparison with the update or training path of an
independently trained PPO backbone. If the gate instead acts on an estimate
$\hat\Align$ satisfying the deterministic bound
$|\hat\Align-\Align|\le\varepsilon_\Align$, then
\begin{equation}\label{app:eq:gated-est}
\begin{aligned}
J(\pi_{\mathrm{gate}})-J(\pi)
&\ge\frac{\ssc\kappa}{1-\gamma}
\big(\Align(g)^{+}-\varepsilon_\Align\big)\\
&\quad-C\ssc\varepsilon_{\mathrm{est}}-r(\ssc),
\qquad r(\ssc)=o(\ssc).
\end{aligned}
\end{equation}
for a constant $C$ depending on the advantage bound and step scale and a nonnegative remainder $r$. Within the idealized model the gate can choose the wrong branch only when
$|\Align(g)|\le\varepsilon_\Align$. The bound does not hold automatically for a
same-batch random estimate; it requires the stated error event.
\end{corollary}
\begin{proof}
By construction the gate shapes iff $\Align(g)>0$. If $\Align(g)>0$,
Theorem~\ref{app:thm:gain} gives the gain
$\tfrac{\ssc\kappa}{1-\gamma}\Align(g)+o(\ssc)+O(\ssc\varepsilon_{\mathrm{est}})
=\tfrac{\ssc\kappa}{1-\gamma}\Align(g)^{+}+\dots$; if $\Align(g)\le0$ the update is the
identity, so the gain is $0=\tfrac{\ssc\kappa}{1-\gamma}\Align(g)^{+}$. Combining
the two cases yields \eqref{app:eq:gated}. For the estimated gate, its decision differs
from the oracle only when $\sign(\hat\Align)\ne\sign(\Align)$, which requires
$|\Align(g)|\le|\hat\Align-\Align|\le\varepsilon_\Align$. On such a game the wrong
choice---shaping when $\Align<0$, or not shaping when $\Align>0$---changes the
first-order gain by at most $\tfrac{\ssc\kappa}{1-\gamma}|\Align(g)|\le\tfrac{\ssc\kappa}{1-\gamma}\varepsilon_\Align$;
subtracting this from \eqref{app:eq:gated} and bounding both the prior-estimation term and the absolute Taylor remainder gives \eqref{app:eq:gated-est}.
\end{proof}

\begin{remark}[Scope of Corollary~\ref{app:cor:gated}]\label{app:rem:gated}
The corollary isolates the incremental direction of the idealized bonus step. It
does not establish that gated PPO is never worse than PPO: the implemented update
also contains the raw-reward gradient, clipping, a learned shared representation,
and a finite-sample sign estimate taken from the update batch. Moreover, after any
earlier shaped update, setting the current bonus to zero does not restore the
counterfactual parameters or data distribution of an independently trained
backbone. The retrospective per-game oracle and the end-to-end online gate are
therefore empirical diagnostics, not realizations of a safety theorem.
\end{remark}

\subsection{T2$'$: the induced update derived, not assumed}
\label{app:t2prime}

Theorem~\ref{app:thm:gain} assumes a policy direction. This subsection instead compares
a shaped and unshaped KL-proximal update from the same policy with the same raw score,
the standard local idealization of PPO's surrogate
\citep{geist2019regularized,neu2017unified}.

\begin{lemma}[A KL-proximal step is an exponential tilt]
\label{app:lem:tilt}
Fix a state $s$, a full-support reference policy $\pi(\cdot\mid s)$, a bounded score $u(s,\cdot)$, and step size $\eta>0$. The KL-proximal update
\[
\begin{aligned}
\pi_\eta(\cdot\mid s)
=\argmax_{\pi'\in\Delta(\A)}
\Big\{&\E_{a\sim\pi'}[u(s,a)]\\
&-\tfrac1\eta\KL(\pi'\|\pi(\cdot\mid s))\Big\},
\end{aligned}
\]
has the closed-form solution
\begin{equation}\label{app:eq:tilt}
\begin{aligned}
\pi_\eta(a\mid s)
&=\frac{\pi(a\mid s)e^{\eta u(s,a)}}
{\sum_b\pi(b\mid s)e^{\eta u(s,b)}}\\
&=\pi(a\mid s)\Big[1+\eta\big(u(s,a)
-\langle u\rangle_{\pi(\cdot\mid s)}\big)\Big]+O(\eta^2).
\end{aligned}
\end{equation}
where $\langle u\rangle_{\pi(\cdot\mid s)}=\sum_b\pi(b\mid s)u(s,b)$. The same
tilt is one natural-policy-gradient step on this score for the tabular-softmax
parametrization.
\end{lemma}
\begin{proof}
The objective is strictly concave on the simplex; stationarity of the Lagrangian gives
$\log\pi'(a)=\log\pi(a\mid s)+\eta u(s,a)-\mathrm{const}$, i.e.\ \eqref{app:eq:tilt};
expanding the normalizer to first order in $\eta$ gives the linearization. For softmax
$\pi_\theta$, the Fisher preconditioner of NPG cancels the $\pi(a\mid s)$ factor of the
vanilla gradient, moving each logit by $\eta(u(s,a)-\langle u\rangle)$ ---
the log of \eqref{app:eq:tilt}.
\end{proof}

\begin{theorem}[First-order incremental shaping effect]\label{app:thm:gainderived}
Let $\pi$ be a full-support policy on the finite discounted MDP, and let $h,f:\St\times\A\to\mathbb R$ be bounded frozen scores, where $h$ is the raw-update
score shared by both paths and $f$ is the shaping score. For fixed finite $\beta\ge0$,
define the two KL-proximal updates
\[
\begin{aligned}
\pi_\eta^0(a\mid s)&\propto\pi(a\mid s)e^{\eta h(s,a)},\\
\pi_\eta^b(a\mid s)&\propto
\pi(a\mid s)e^{\eta[h(s,a)+\beta f(s,a)]}.
\end{aligned}
\]
Then, as $\eta\to0$,
\begin{equation}\label{app:eq:gainderived}
\begin{aligned}
J(\pi_\eta^b)-J(\pi_\eta^0)
&=\frac{\eta\beta}{1-\gamma}\widetilde\Align[f]+O(\eta^2),\\
\widetilde\Align[f]
&=\E_{s\sim\dpi,a\sim\pi}[f(s,a)\Adv(s,a)] .
\end{aligned}
\end{equation}
The shared raw score $h$ cancels from the first-order difference.
\end{theorem}
\begin{proof}
Lemma~\ref{app:lem:tilt} applied to a general bounded score $u$ gives
$\pi_\eta^u(a\mid s)=\pi(a\mid s)\{1+\eta[u(s,a)-\E_\pi u(s,\cdot)]\}+O(\eta^2)$.
Subtracting the expansions for $u=h+\beta f$ and $u=h$ cancels the common raw term.
The performance-difference lemma then gives \eqref{app:eq:gainderived}; the centering
term vanishes because $\E_{a\sim\pi}[\Adv(s,a)]=0$, and replacing either perturbed
occupancy by $\dpi$ changes only the $O(\eta^2)$ remainder.
\end{proof}

\begin{corollary}[Oracle local gate]\label{app:cor:incrementalgate}
If $\beta_g=\beta\mathbf 1\{\widetilde\Align[f]>0\}$, then
\[
J(\pi_\eta^g)-J(\pi_\eta^0)
=\frac{\eta\beta}{1-\gamma}\max\{\widetilde\Align[f],0\}+O(\eta^2).
\]
Thus only the leading incremental effect is nonnegative, under the same idealized
update and at the common starting policy.
\end{corollary}

For the immediate OPR direction, set $f=\sdratio$. For the complete expected pre-GAE
direction, $f(s,a)=\E[\sum_{l\ge0}(\gamma\lambda)^l\sdratio_{t+l}\mid s_t=s,a_t=a]$.
The deployed covariance gate uses the immediate bounded score with raw-reward GAE and
is therefore a practical proxy for the propagated coefficient, not an exact estimator
of it in general.

\begin{proposition}[$\tilde\Align$ versus $\Align$]\label{app:prop:phirel}
In the state-blind regime $\pi(a\mid s)=\rocc(a)$ for all $s$ 
(a separate special case, not the state-dependent anti-optimal policy of Theorem~\ref{app:thm:insuff}), with $\pcurr=\rocc$, write
$u(a):=\big(\pgood(a)-\rocc(a)\big)\,\E_{s\sim\dpi}[\Adv(s,a)]$. Then
\[
\begin{aligned}
\Align(g)&=\sum_a u(a),\\
\tilde\Align(g)&=\sum_a w(a)u(a),\\
w(a)&=\frac{2\rocc(a)}{\pgood(a)+\rocc(a)}\in(0,2),
\end{aligned}
\]
so $\tilde\Align$ is a \emph{positive reweighting} of $\Align$'s summands with
$|w(a)-1|=|\sdratio(a)|/2$. Consequently
$|\tilde\Align-\Align|\le\tfrac12\max_a|\sdratio(a)|\cdot\sum_a|u(a)|$, and
\begin{equation}\label{app:eq:phirel}
\begin{aligned}
\sign\tilde\Align&=\sign\Align
\quad\text{whenever}\\
|\Align(g)|&>\tfrac12\max_a|\sdratio(a)|\sum_a|u(a)| .
\end{aligned}
\end{equation}
Thus the difference is second order near a fixed full-support $\pcurr$, but sign agreement still requires the displayed margin condition: cancellation can make $\Align$ second order or smaller. On
$4{,}000$ random instances the identity holds to machine precision and
\eqref{app:eq:phirel} exhibits no violation; unconditionally the two signs agree on
$\approx88\%$ of random instances, so the quantities are aligned but \emph{not}
interchangeable when $\Align$ is dominated by cancellation.
\end{proposition}
\begin{proof}
\[
\Align=\sum_a\pgood(a)\E_s[\Adv(s,a)]
=\sum_a(\pgood-\rocc)(a)\E_s[\Adv(s,a)],
\]
because $\sum_a\rocc(a)\E_s[\Adv(s,a)]=0$ in the state-blind regime. Inserting
$\pi(a\mid s)=\rocc(a)$ and the bounded-score identity gives
\[
\tilde\Align=\sum_a\rocc(a)\sdratio(a)\E_s[\Adv(s,a)]
=\sum_a w(a)u(a).
\]
The identity $|w-1|=|\sdratio|/2$ is algebra; the sign condition follows from
\[
|\tilde\Align-\Align|
\le\max_a|w(a)-1|\sum_a|u(a)|<|\Align|.
\]
\end{proof}

\begin{remark}[A weight-free gate estimator]\label{app:rem:covgate}
$\tilde\Align$ is estimable on-policy with \emph{no importance weights}:
$\hat{\tilde\Align}=\operatorname{mean}_t[\sdratio(a_t)\hat A_t]$, whose summands are
bounded by $2|\hat A_t|$ --- against $(W{-}1)|\hat A_t|$ for the truncated IS estimator
of $\Align$. Under the ideal held-out bounded-summand model of
Corollary~\ref{app:cor:heldoutradius}, replacing a $(W{-}1)$ range scale by $2$ would
shrink the squared sample-size term by
$\big(\tfrac{W-1}{2}\big)^2\approx20$ at $W{=}10$. This is a variance-scale
comparison, not a calibrated sample requirement for temporally dependent neural data.
Since Theorem~\ref{app:thm:gainderived} identifies $\tilde\Align$ --- not $\Align$ ---
as the exact first-order coefficient of the derived myopic update, the covariance is
the better-matched local statistic. We evaluate both an IS hard gate and this covariance
hard gate online; the retrospective oracle remains labeled by $\Align$.
\end{remark}

\begin{remark}[What remains assumed]\label{app:rem:t2scope}
Theorem~\ref{app:thm:gainderived} removes the assumed policy direction for the
incremental comparison of two tabular KL-proximal updates. Still idealized: (i) PPO's clipped surrogate approximates, but
is not identical to, the KL-proximal step; (ii) with shared function approximation the
tilt is distorted by the representation; (iii) the $l\ge1$ GAE terms of
Result~2 add a state-coupling correction, bounded via
$\|f_{\rm tail}\|_\infty\le2\gamma\lambda/(1-\gamma\lambda)$, where $f_{\rm tail}$ is the expected $l\ge1$ bounded-score tail. Including this tail in $f$ gives its exact first-order contribution in Theorem~\ref{app:thm:gainderived}; replacing it by the immediate score is an additional approximation. The finite clipped neural update remains outside that idealization.
\end{remark}

\subsection{T8: when the marginal is provably insufficient}
\label{app:t8}

\begin{theorem}[Marginal-insufficiency]\label{app:thm:insuff}
There is a finite discounted MDP with an optimal and an arbitrarily suboptimal policy sharing the same action marginal. A population elite prior can have that same marginal, so marginal OPR supplies no shaping signal at the suboptimal policy. At a full-support suboptimal policy in this family, a full-support state-conditioned elite prior instead has strictly positive alignment under Theorem~\ref{app:thm:gain} and a strictly positive bounded-score coefficient under Theorem~\ref{app:thm:gainderived}.
\end{theorem}
\begin{proof}[Construction and computation]
Take two actions $\{a_0,a_1\}$ and an action-independent cycle $s_0\to s_1\to s_0$, with $\rho_0(s_0)=\rho_0(s_1)=1/2$ and reward $\rew(s_i,a_j)=\mathbf 1\{i=j\}$. Uniform initialization makes the discounted visitation uniform for every policy. Let $\pi_q$ choose the correct action with probability $q$ in each state. Then $\rocc_{\pi_q}=(1/2,1/2)$ for every $q\in[0,1]$, while $J(\pi_q)=q/(1-\gamma)$. Thus the optimal $\pi_1$ and anti-optimal $\pi_0$ share a marginal, and their return gap grows without bound as $\gamma\uparrow1$.

For a population elite source, fix $0<q_0<q_1<1$ and take $P_{\rm good}(a_i\mid s_i)=q_1$. Both this source and $\pi_{q_0}$ have uniform action marginals, so at $K=1$, $\pgood=\pcurr$ and every shaping bonus is zero. The shaped update therefore equals the raw update from the same policy and data; it does not stop raw-reward learning. The marginal diagnostic itself is $\Align=1/2-q_0$, showing why a nonzero assumed-direction diagnostic need not imply a nonzero marginal bonus.

This population construction can also arise from elite selection: take even-length episodes with uniform initial phase under $\pi_{q_0}$ and select the top return quantile, breaking return ties independently of state/action labels. Simultaneously swapping the state and action labels preserves trajectory probability and return, so the selected population action marginal remains uniform. A nontrivial upper-return selection raises the correct-action rate above $q_0$; an appropriate quantile retains failures as well as successes and gives some $q_1<1$. Finite empirical buffers need not be exactly balanced; the zero-signal claim is a population statement.

At $K=2$, the correct- and incorrect-action advantages under $\pi_{q_0}$ are $1-q_0$ and $-q_0$, since the next-state distribution is action-independent. Hence
\[
\Align_{\rm state}=q_1(1-q_0)+(1-q_1)(-q_0)=q_1-q_0>0.
\]
The conditional bounded scores are
\[
f_{\rm correct}=\frac{2(q_1-q_0)}{q_1+q_0}>0,\qquad
f_{\rm incorrect}=-\frac{2(q_1-q_0)}{2-q_1-q_0}<0.
\]
Consequently $\widetilde\Align[f]=q_0(1-q_0)(f_{\rm correct}-f_{\rm incorrect})>0$. This proves the separation for both local models with full-support priors and well-defined logarithms, without claiming that empirical clustering always recovers the two states.
\end{proof}

\begin{remark}[Diagnostic reading]
The construction proves that state aliasing can erase a marginal signal; it does not
classify real games from their names or reward taxonomy. Paddle games such as Pong and
Breakout motivated the hypothesis, while sparse-reward games such as Frostbite and
Qbert motivated the contrasting hypothesis. Section~\ref{app:phi-est} reports a
retrospective in-sample association, not confirmation on held-out tasks.
\end{remark}

\subsection{Estimating $\Align(g)$ from a checkpoint}
\label{app:phi-est}

The predictor \eqref{app:eq:phi} involves a counterfactual sum over all actions, but is
estimable from on-policy rollouts of a saved checkpoint without retraining.

\begin{proposition}[Importance-sampling identity and estimator]\label{app:prop:isest}
For a full-support policy $\pi$ and a fixed prior $\pgood$,
\begin{equation}\label{app:eq:isphi}
\begin{aligned}
\Align(g)=\E_{\substack{s\sim\dpi\\a\sim\pi(\cdot\mid s)}}
\Big[\Big(\tfrac{\pgood(a)}{\pi(a\mid s)}-1\Big)\Adv(s,a)\Big].
\end{aligned}
\end{equation}
If samples have this state--action marginal and $\E[\hat A\mid s,a]=\Adv(s,a)$, their empirical average is unbiased. An independent, frozen prior can be treated conditionally in the same way. Consistency additionally requires a law of large numbers for the sampling process. Learned-critic bias, undiscounted rollout weighting, clipped ratios, or fitting the prior to the same evaluation samples need not preserve these conclusions.
\end{proposition}
\begin{proof}
By full support,
\[
\begin{aligned}
&\sum_a(\pgood(a)-\pi(a\mid s))\Adv(s,a)\\
&\quad=\sum_a\pi(a\mid s)
\Big(\tfrac{\pgood(a)}{\pi(a\mid s)}-1\Big)\Adv(s,a).
\end{aligned}
\]
Taking $\E_{s\sim\dpi}$ and using advantage centering gives \eqref{app:eq:isphi}. Conditional unbiasedness of $\hat A$ and linearity give the sample-average statement. The identity extends to a conditional prior by replacing $\pgood(a)$ with $\pgood(a\mid z(s))$ throughout.
\end{proof}

\begin{remark}[Control-variate variance reduction]
Both estimators $\operatorname{mean}_t[w_t\hat A_t]$ and
$\operatorname{mean}_t[(w_t-1)\hat A_t]$ (with $w_t=\pgood(a_t)/\pi(a_t\mid s_t)$) have
the same expectation under the assumptions of Proposition~\ref{app:prop:isest}, because the subtracted term $\operatorname{mean}_t[\hat A_t]$
estimates $\E_{a\sim\pi}[\Adv(s,a)]=0$ (Remark~\ref{app:rem:centered}); i.e.\ ``$-1$''
subtracts a control variate of known conditional mean zero. Writing $X=w\hat A$ and
$C=\hat A$, one has $\Var[X-C]=\Var[X]-\big(2\Cov[X,C]-\Var[C]\big)$, so the
$(w-1)$ form has smaller variance exactly when $\Cov[w\hat A,\hat A]>\tfrac12\Var[\hat A]$
; it is not an unconditional variance reduction, and learned advantages need not have zero conditional mean. Empirically this
control-variate form is what stabilizes the \emph{sign} of $\hat\Align$ at moderate
sample sizes (the naive form gave the wrong sign on \textsc{Frostbite} at low episode
counts).
\end{remark}

\begin{definition}[Local bonus coefficient]\label{app:def:localcoef}
At the current parameters $\theta$, let $U_0(\eta)$ and $U_1(\eta)$ be one optimizer
step of size $\eta$ on, respectively, the raw and bonus-shaped objectives, constructed
from the same current policy, with $U_0(0)=U_1(0)=\theta$. If both paths are
differentiable at zero and $J$ is Fr\'echet differentiable at $\theta$, define
\begin{equation}\label{app:eq:localcoef}
C_\lambda(\theta):=\nabla J(\theta)^\top
\left.\frac{\partial}{\partial\eta}\big(U_1(\eta)-U_0(\eta)\big)\right|_{\eta=0}.
\end{equation}
The subscript records that the shaped objective contains the propagated bonus
$B_t=\sum_{l\ge0}(\gamma\lambda)^l b_{t+l}$. Thus $C_\lambda$ is the true local effect
of adding that bonus direction, relative to the raw step; it need not equal either the
immediate covariance proxy or $\Align$.
\end{definition}

\begin{theorem}[Selective one-update guarantee]\label{app:thm:selectivegate}
Under Definition~\ref{app:def:localcoef}, let $\hat C_\lambda$ be any estimate with a valid radius $r_n(\delta)$ satisfying
\begin{equation}\label{app:eq:radius}
\Pr\!\left(|\hat C_\lambda-C_\lambda|\le r_n(\delta)\right)\ge1-\delta,
\end{equation}
and apply the bonus only under the selective rule
\begin{equation}\label{app:eq:trust}
g=\mathbf 1\!\left\{\hat C_\lambda>r_n(\delta)\right\}.
\end{equation}
Then, with probability at least $1-\delta$: (i) activation certifies
$C_\lambda>0$; (ii) if a beneficial bonus is rejected, then
$0<C_\lambda\le2r_n(\delta)$; and (iii), whenever the gate activates,
\begin{equation}\label{app:eq:selectivegain}
J(U_1(\eta))-J(U_0(\eta))=\eta C_\lambda+o(\eta)>0
\end{equation}
for sufficiently small positive $\eta$. In particular, the probability of activating
on a nonpositive local coefficient is at most $\delta$.
\end{theorem}
\begin{proof}
Work on the event in \eqref{app:eq:radius}. If the gate activates, then
$C_\lambda\ge\hat C_\lambda-r_n(\delta)>0$. If it rejects while $C_\lambda>0$, then
$C_\lambda\le\hat C_\lambda+r_n(\delta)\le2r_n(\delta)$. Finally, a first-order
Taylor expansion of the two update paths around their common starting point gives
\eqref{app:eq:selectivegain}; its sign is positive for sufficiently small $\eta$ when
$C_\lambda>0$. The complement of the radius event has probability at most $\delta$.
\end{proof}

\begin{corollary}[One valid idealized radius]\label{app:cor:heldoutradius}
Condition on a score and update direction constructed independently of a held-out
sample. If $X_1,\ldots,X_n$ are independent, $\E X_i=C_\lambda$, and every $X_i$ lies
in an interval of width $B$, then for
$\hat C_\lambda=n^{-1}\sum_iX_i$,
\begin{equation}\label{app:eq:heldoutradius}
r_n(\delta)=B\sqrt{\frac{\log(2/\delta)}{2n}}
\end{equation}
satisfies \eqref{app:eq:radius} by Hoeffding's inequality. The theorem therefore
applies to this held-out bounded setting.
\end{corollary}

\begin{remark}[What the theorem does and does not cover]\label{app:rem:selectivescope}
The result is conditional on a valid radius for the \emph{correct} propagated target.
In a neural implementation its error budget has distinct sampling, prior-estimation,
critic-bias, temporal-proxy, and dependence terms. Same-batch reuse, Markov samples,
a learned critic, EMA state, and substituting an immediate covariance for
$C_\lambda$ do not satisfy Corollary~\ref{app:cor:heldoutradius} automatically. Nor
does a sequence of locally certified steps reproduce the counterfactual data and
parameters of an independently trained backbone. The theorem is thus a selective,
one-update statement---not a PPO training-path safety guarantee.
\end{remark}

\begin{figure}[H]
\centering
\includegraphics[width=\columnwidth]{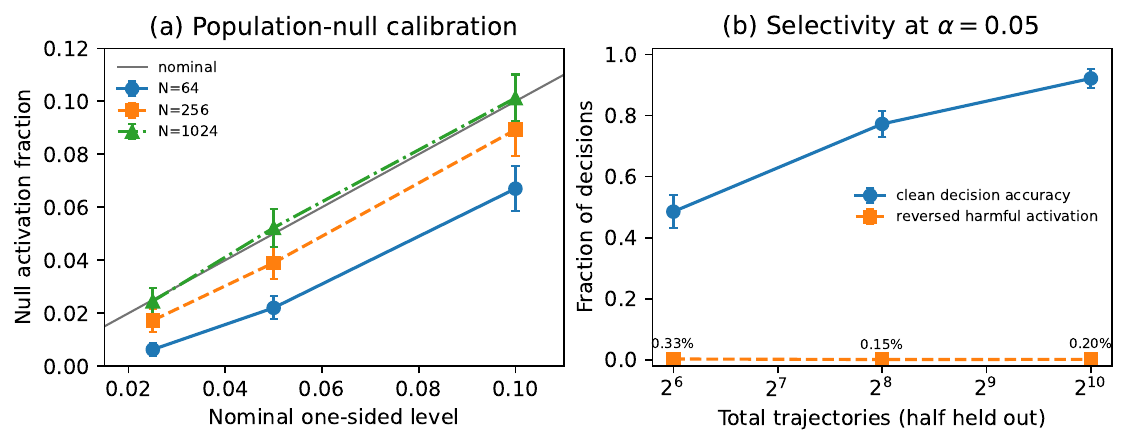}
\caption{Calibration diagnostic on 200 randomized delayed-credit environments (20
repetitions each). A score is estimated on half of each batch and the propagated
coefficient on the held-out half; four disjoint fold estimates form a $t_3$ abstention
rule. (a) Under the exact population null, observed activation tracks or is below the
nominal one-sided level. (b) At nominal $\alpha=0.05$, clean population-decision
accuracy rises from $48.6\%$ to $92.2\%$ as the total batch grows from 64 to 1,024,
while harmful activation under a reversed-prior stress test is $0.33\%$, $0.15\%$,
and $0.20\%$. Error bars are 95\% CIs clustered over environments. The $t$ rule is an
empirical calibration device, not the distribution-free radius of
Corollary~\ref{app:cor:heldoutradius}; these synthetic results do not certify neural
PPO.}
\label{app:fig:selective-calibration}
\end{figure}

\begin{remark}[Empirical validation]
Measuring $\hat\Align$ at a mid-training ($5$M-agent-step) checkpoint on the in-sample
$10$-game set (main paper, Fig.~\ref{fig:phi}) gives sign agreement on all seven games with
$|\Delta\mathrm{HNS}|>5$ and Spearman correlations $+0.517$ and $+0.500$ at nominal 30- and 100-episode budgets (neither significant at conventional levels). Breakout contributes 8 and 17 episodes, respectively. The signs are stable across these measurements, but both reuse the same games and checkpoint family. Moreover $\hat\Align$ is neither the bounded
myopic coefficient $\tilde\Align$ nor the propagated coefficient $C_\lambda$.
Accordingly this is a retrospective diagnostic association, not a direct theorem test
or held-out prediction.
\end{remark}

\subsection{Controlled tabular validation}\label{app:toy-validation}

Unless stated otherwise, $\pm$ denotes a normal 95\% Monte Carlo confidence half-width over the independent MDPs or paired replications specified below, not a standard deviation. Repeated estimates within an MDP are aggregated before forming across-MDP intervals; proportions use the stated Wilson intervals. Sensitivity grids are descriptive, without multiplicity-adjusted confirmatory claims.

The Atari results test the complete algorithm but do not isolate the
mechanisms in the analysis.  We therefore add controlled finite-MDP experiments
with exact value and occupancy calculations.  These experiments are diagnostic:
they test the local update model, state aggregation, temporal placement, and gate
estimation separately; they are not additional benchmark tasks or evidence that
gated PPO is globally safe.

\begin{table*}[t]
\centering
\small
\setlength{\tabcolsep}{4pt}
\caption{Evidence for each empirically testable theoretical prediction. Exact algebraic identities are audited numerically but do not require statistical confirmation.}
\label{tab:claim-evidence}
\begin{tabular}{@{}p{0.19\textwidth}p{0.43\textwidth}p{0.29\textwidth}@{}}
\toprule
Claim & Test & Result \\
\midrule
Score identities & Random action distributions for $|\A|\in\{2,4,18\}$ & Maximum identity error $3.6\times10^{-15}$ \\
Dense GAE propagation & Exact decomposition and sampled delayed-decision task & Error $1.1\times10^{-16}$; pre-GAE reaches $0.983$ versus $0.858$ raw \\
Prior-estimation rate & $4{,}000$ multinomial estimates for each $N,K,|\A|$, and prior shape & Log--log slopes $[-0.504,-0.481]$; all means below the bound \\
First-order effect & $5{,}000$ random finite MDPs & Residual scales as $\eta^{1.9995}$ \\
Elite alignment & Rollout-derived priors in $500$ random MDPs & Positive in $95.0\%$; stable across horizon and elite fraction \\
Selective gating & Oracle, noisy, cross-fitted, and held-out synthetic gates & Increasing samples improves sign recovery; abstention controls harmful activation \\
State conditioning & Aliased, shared-action, noisy-label, merging, and over-splitting controls & Conditioning helps only when it resolves action-relevant aliasing \\
Decay & $120$ MDPs with a persistent reversed prior & Decay recovers near-raw progress; constant shaping does not \\
Checkpoint diagnostic & Ten-game Atari set & Correct sign on all $7/7$ non-negligible effects; retrospective, not held out \\
\bottomrule
\end{tabular}
\end{table*}

\begin{figure*}[t]
\centering
\includegraphics[width=0.78\textwidth]{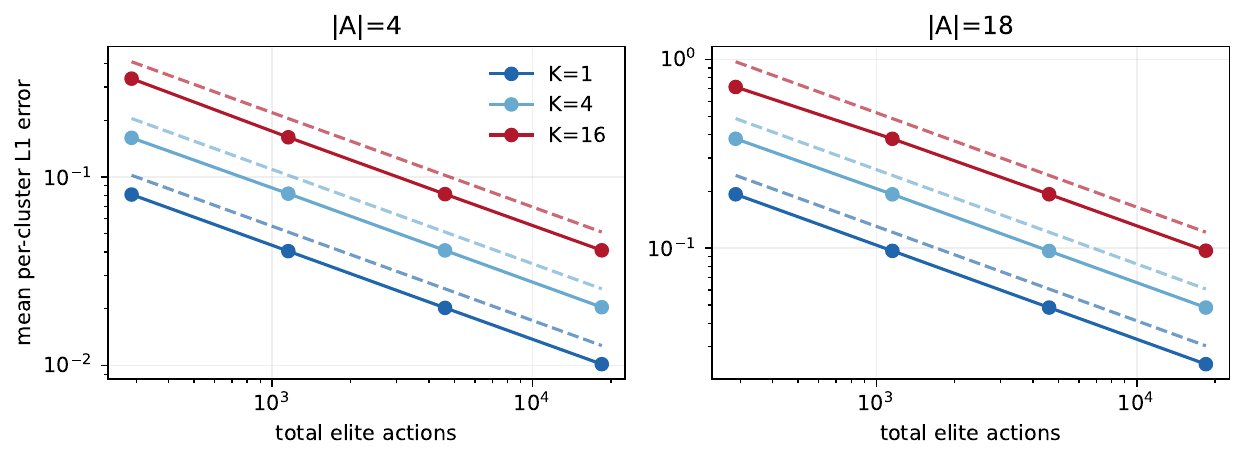}
\Description{Log-log plots of prior-estimation error against total elite actions for one, four, and sixteen balanced state clusters. Empirical errors follow parallel inverse-square-root slopes below the theoretical bounds.}
\caption{Direct validation of the prior-estimation prediction. Solid lines are empirical mean per-cluster $L_1$ errors; dashed lines are $\sqrt{(|\A|-1)/(N_{\rm elite}/K)}$. Each point uses $4{,}000$ independent estimates.}
\label{fig:prior-estimation}
\end{figure*}

\paragraph{Direct prior-estimation test.}
We draw $4{,}000$ independent multinomial estimates for every combination of $|\A|\in\{4,18\}$, $K\in\{1,4,16\}$, four total elite-sample counts, and uniform or skewed priors. Clusters are balanced, so each receives $N_{\rm elite}/K$ samples, matching Theorem~\ref{app:thm:l1}. Across the twelve fitted series, log--log slopes lie in $[-0.504,-0.481]$, matching the $N_{\rm elite}^{-1/2}$ rate. At fixed total $N_{\rm elite}$, the observed $K$-dependent increase is within $0.926$--$1.029$ of $\sqrt K$; every empirical mean remains below the theorem's bound (largest ratio $0.814$). Figure~\ref{fig:prior-estimation} shows the uniform-prior cases; skewed priors give the same rate. This experiment tests the theorem's i.i.d., comparably occupied setting, not trajectory dependence or learned state clusters.

\paragraph{Incremental theorem on random MDPs.}
We sample 5,000 six-state, three-action discounted MDPs
($\gamma=0.9$), with full-support transition kernels and policies.  For a frozen
bounded OPR score $f$ and the exact task advantage $h=A^\pi$, we compare the same
raw KL-proximal update with and without shaping,
\[
\begin{aligned}
\pi_\eta^0(\cdot\mid s)&\propto\pi(\cdot\mid s)e^{\eta h(s,\cdot)},\\
\pi_\eta^b(\cdot\mid s)&\propto
\pi(\cdot\mid s)e^{\eta[h(s,\cdot)+\beta f(\cdot)]}.
\end{aligned}
\]
The predicted increment is $\eta\beta\widetilde\Phi/(1-\gamma)$, where
\[
\widetilde\Phi=\E_{s\sim d^\pi,a\sim\pi}[f(a)A^\pi(s,a)].
\]
The median absolute residual follows $\eta^{1.9995}$ over the four smallest
step sizes (five disjoint 1,000-MDP blocks: exponents 1.995--2.008), matching the
$O(\eta^2)$ remainder.  Sign agreement is $99.96\%$ at $\eta=10^{-3}$ and
$94.98\%$ at $\eta=0.3$.  At the latter step size, agreement is $81.36\%$ in the
smallest-$|\widetilde\Phi|$ quartile and at least $98.56\%$ in every other
quartile.  This confirms the first-order coefficient while illustrating why its
sign is not a finite-step safety guarantee.
In a separate 1,000-MDP ensemble we replace $h=A^\pi$ by an independently sampled
bounded score $h(s,a)=\tanh Z(s,a)$, $Z\sim\mathcal N(0,1)$.  The residual exponent
is 2.004 and sign agreement remains $98.7\%$ at $\eta=0.1$, directly checking that
the common raw term cancels at first order rather than relying on the special
choice $h=A^\pi$.

\paragraph{Priors constructed by elite selection.}
The preceding stress test deliberately samples arbitrary priors to obtain both
alignment signs.  In a separate 500-MDP experiment, each current policy generates
2,000 length-25 rollouts and the action marginal of the top-return quartile forms
$P_{\rm good}$.  The resulting $\widetilde\Phi$ is positive in $475/500$ MDPs
($95.0\%$, Wilson $95\%$ CI $[92.7,96.6]\%$), and its sign agrees with the exact
increment on $99.8\%$, $99.6\%$, and $99.0\%$ of MDPs at
$\eta\in\{0.01,0.03,0.1\}$.  Elite selection therefore makes positive alignment
common in this family but does not guarantee it: 25 MDPs retain a negative
alignment; state aggregation is one possible source of this mismatch, but this experiment alone does not identify its cause.
These fixed-length rollouts act as pseudo-episodes in otherwise continuing MDPs;
their undiscounted ranking is therefore only an analogue of the episodic replay
buffer used in the benchmark experiments.

This conclusion is stable, but its magnitude is protocol-sensitive.  On 200
additional paired MDPs, positive alignment remains between $94.5\%$ and $96.5\%$
over rollout horizons $\{10,25,50\}$ and elite fractions
$\{0.1,0.25,0.5\}$, with $99$--$100\%$ finite-step sign agreement at
$\eta=0.03$.  The mean exact increment nevertheless falls from $0.0068$ at
horizon 10/top 10\% to $0.0013$ at horizon 50/top 50\%, as the mean marginal
gap shrinks from $0.193$ to $0.038$.  Elite selection is therefore usually
aligned in this generator, but broader/longer aggregation dilutes its strength.

\paragraph{Finite-sample gate for the idealized coefficient.}
On 250 further MDPs, we estimate $\widetilde\Phi$ from samples of the exact
discounted occupancy after adding zero-mean noise with standard deviation equal
to the MDP's advantage RMS\@.  Eighty estimates are drawn per MDP and sample size.
At 1,024 state-action samples, the hard rule
$\mathbb{1}\{\hat{\widetilde\Phi}>0\}$ selects the correct sign in $94.63\%$
of trials but activates a negative exact finite-step increment in $2.65\%$.
A conservative rule
$\mathbb{1}\{\hat{\widetilde\Phi}>1.645\,\widehat{\rm SE}\}$ reduces harmful
activations to $0.13\%$, while reducing positive-case recall from $94.56\%$ to
$75.99\%$.  Thus uncertainty-aware gating trades recall for fewer harmful
activations; the calculation still omits critic bias, temporal dependence, and
shared neural parameters.  Because within-batch centering and scaling make this
a self-normalized sample covariance, the threshold is a descriptive plug-in
rule rather than a calibrated 95\% test.

We therefore stress-test score-correlated critic bias separately on 250 MDPs.
With unbiased error, increasing the batch from 128 to 1,024 improves hard-gate
accuracy from $87.3\%$ to $95.5\%$ and reduces harmful activation from $6.8\%$
to $2.5\%$.  Adding bias of one-half advantage RMS in the shaping-score
direction instead makes $47.3\%$ and $49.2\%$ of hard-gate decisions harmful.
The 1.645-SE rule also rises from $31.5\%$ to $44.1\%$ harmful decisions because
more samples increase confidence in the biased mean.  Conversely, a bias of one
RMS in the opposite direction eliminates harmful activation by suppressing all
beneficial shaping.  Confidence thresholds therefore address variance, not
systematic critic error.

There is a second, distinct failure mode: the estimator can concentrate around
the wrong update target.  PPO normalizes the raw and shaped advantages
separately, so the matched coefficient is the task-gradient inner product with
the difference of those two \emph{normalized} actor updates
under the fully specified update-operator definition. Across 800 exact random MDPs at
$\beta=0.5$, the immediate covariance and this operator-matched coefficient
agree in sign on only $50.4\%$; the exact coefficient itself predicts a retained
$10^{-4}$ finite difference on all 800 MDPs.  In a 360,000-decision sampled-
minibatch analogue, covariance agreement remains $51.0$--$51.2\%$ as batch size
grows from 32 to 2,048 even though its mean absolute estimation error falls
eightfold.  A shared-scale control reaches $88.0\%$ agreement at batch 2,048,
localizing most of the discrepancy to separate normalization.

We therefore also test an operator-matched, cross-fitted gate: one batch forms
the normalized shaped-minus-raw actor direction, and an independent held-out
batch estimates its inner product with the raw task gradient.  At
$\beta=0.1$, increasing each batch from 32 to 2,048 raises sign agreement from
$72.2\%$ to $95.3\%$.  A one-sided lower-bound rule reduces conflict among
activations from $27.5\%$ to $0.76\%$, while coverage rises from $5.94\%$ to
$17.1\%$.  At $\beta=0.5$, conflict is $2.55\%$ but coverage only $0.78\%$ at
batch 2,048.  This is precisely the margin--radius tradeoff: a reliable gate is
selective when positive alignment is rare or small.  It is a controlled
development result using exact tabular advantages, not a retrospective
guarantee for the historical neural-PPO gate.

\paragraph{Marginal bias versus conditional variance.}
The representation and estimation analysis predicts two
opposing effects.  We test both in four-context/four-action contextual tasks,
varying action structure, context imbalance, 32--2,048 elite samples, label
noise, merging, and over-splitting over 300 paired repetitions per cell.  In the
balanced one-to-one task at $n=2{,}048$, clean conditioning improves correct-
action probability over $K=1$ by $0.5696\pm0.0008$ (95\% interval); the gain
falls to $0.3785$ with 20\% symmetric label noise, $0.1797$ with 40\% noise,
and zero with random labels.  Merging contexts retains only $0.1955$ when the
merged states require different actions, but retains the full oracle gain
($0.3742$) in a paired-action task where the merge is action-preserving.  These
controls show that action-relevant information, rather than $K$ itself, removes
aliasing bias.

The adverse controls expose the variance side.  In a shared-action task, where
$K=1$ is already sufficient, clean conditioning at $n=32$ is $0.0222\pm0.0020$
below $K=1$, and unnecessary over-splitting is $0.0562$ below.  Over-splitting
also costs roughly $0.028$--$0.034$ relative to clean conditioning across all
four action structures at $n=32$, while becoming nearly unbiased at
$n=2{,}048$.  Hence increasing $K$ helps only when the reduction in
action-relevant aliasing exceeds the cost of fewer reliable elite samples per
cluster.  Oracle or controlled labels isolate this mechanism; they do not show
that a learned representation will find the required partition.

\paragraph{Credit assignment through pre-GAE temporal propagation.}
Finally, a length-10 delayed-branch MDP makes the first action determine a later
action-sequence signature and terminal success.  From initial success probability
$0.1$, 20 exact updates reach $0.260$ with raw GAE, $0.018$ when the same marginal
score is attached only to the initial transition, and $0.981$ when the score is
inserted before GAE ($\gamma=0.99$, $\lambda=0.95$).  This demonstrates that a
future elite-sequence signature can reach the earlier choice; it does not imply
that propagation is always beneficial.  When the elite prior is fully
contaminated by failure trajectories, pre-GAE shaping instead reaches $0.038$
versus raw GAE's $0.260$, while the oracle local sign gate suppresses the bonus
and returns to $0.260$.  Across a sweep of $\lambda$ and contamination, ungated
pre-GAE shaping ranges from $0.369$ below to $0.778$ above the raw update.

The effect survives sampled learning.  In 200 stochastic replications with 256 trajectories per
update and a freshly estimated top-quartile prior, clean pre-GAE shaping reaches
success probability $0.983$ after 20 updates versus $0.858$ for raw GAE and
$0.102$ for the same score applied only locally.  When $75\%$ of the success
labels used for elite ranking are flipped, ungated pre-GAE shaping falls to
$0.053$.  A same-batch branch-gradient sign gate reaches $0.858$.  More
importantly, a second gate using the deployed immediate bounded-score/raw-GAE
covariance, hard sign decision, and EMA rate $0.02$ reaches $0.851\pm0.009$
(paired difference from raw: $-0.006\pm0.009$), while improving over ungated
pre-GAE by $0.798\pm0.009$.  Under clean ranks it preserves the complete early
gain ($0.983$), and under fully reversed ranks it never activates and exactly
matches the paired raw run.  Thus the deployed statistic suppresses a strongly
harmful shaping direction in this construction, but it does not establish an
improvement over raw learning under intermediate corruption.  The experiment
includes sampling, same-batch sign selection, and the deployed temporal EMA,
but still uses a one-parameter centered policy-gradient update without PPO
advantage normalization or clipping, zero-critic raw GAE, and a fixed suffix
controller.
Across the five corruption levels, the unsmoothed covariance statistic agrees
with the complete propagated-effect sign on $96.6$--$100\%$ of
replication-updates
(Pearson $r=0.95$--$0.99$).  At 50\% corruption, however, rate-0.02 EMA sign
agreement falls to $52.3\%$ because the direction fluctuates near zero.  This
separates an accurate instantaneous proxy in this construction from lag induced
by temporal smoothing.

We additionally sweep four batch sizes and four gate-EMA rates over 100 paired
stochastic replications per condition.  All 16 configurations preserve the
clean shaping gain and
exactly suppress fully reversed shaping, but intermediate corruption exposes a
hysteresis-variance tradeoff.  At batch size 256 and 50\% corruption, the
paired gated-minus-raw effect after 20 updates ranges from
$-0.009\pm0.012$ at the synthetic-study EMA rate 0.02 to $+0.067\pm0.003$ without
smoothing.  This sensitivity makes the EMA rate a reported algorithmic
hyperparameter rather than part of a universal safety claim.

To audit same-batch selection, we also perform two-fold cross-fitting: each half
of a 256-trajectory batch estimates a prior and gate used only on the other
half's gradient, and the two directions are averaged.  After 30 updates and 200
paired replications, cross-fitted gating reaches $0.986$ versus raw $0.941$ with clean
ranks, $0.944$ versus $0.941$ at 50\% corruption, and $0.930$ versus $0.941$ at
75\% corruption (paired difference $-0.010\pm0.010$), while cross-fitted
ungated shaping collapses to $0.051$.  Fully reversed ranks produce zero gate
activation and exactly match raw learning.  Thus update-sample reuse is not the
sole source of protection in this construction, although splitting increases
gate variance and is not the deployed algorithm.

To test breadth beyond that single suffix pattern, we generate 1,000
independently seeded exact binary-branch environments, stratified over horizons
$\{4,8,16,32\}$ and randomizing $\gamma$, $\lambda$, the initial policy, elite
fraction, shaping scale, and the strength and temporal coherence of the suffix
action signature.  Under clean rankings, the immediate-covariance gate improves
final success over raw by $0.1249\pm0.0110$ on average, but harms $20.8\%$ of
environments.  Under fully reversed rankings it reduces the harm rate from
$77.2\%$ for ungated shaping to $7.7\%$, with mean difference from raw
$-0.00135\pm0.00070$.  The immediate and complete propagated signs agree on
$71.7\%$ of clean environments (Wilson 95\% CI $[68.8,74.4]\%$), increasing
from $55.6\%$ for $0.5\leq\lambda<0.625$ to $92.0\%$ for
$0.875\leq\lambda\leq1$.  Agreement falls from $100\%$ in the lowest suffix-
coherence quartile to $45.8\%$ in the highest, where a consistent later action
signature aliases the oppositely coded causal first action.  A population gate
using the propagated-score target
never harms raw in this exact family and has zero harmful activations.  This
reference identifies a more faithful estimand for future cross-fitted
estimation; it is not an implemented PPO result or a general safety guarantee.

We next test such estimation without oracle access.  On the first 200
environments, one half of each batch estimates the smoothed elite/current score
and a disjoint half estimates either the immediate covariance or the covariance
of terminal success with the propagated bonus-to-go.  At total batch sizes
$64$, $256$, and $1{,}024$, propagated sign accuracy is respectively
$94.8\%$, $98.0\%$, and $99.0\%$, compared with $70.9\%$, $73.2\%$, and
$73.5\%$ for the immediate statistic on the same held-out trajectories.  Clean
harmful activation falls from $1.5\%$ to $0.5\%$ for the propagated estimator
but remains $18$--$20\%$ for the immediate proxy.  Larger batches therefore
remove sampling error around the propagated target but cannot repair the
immediate target's temporal mismatch.  These task-clustered results use 20
repetitions per environment; they do not include critic bias or PPO clipping.
At 50\% corruption the population elite and current marginals coincide, yet
finite elite selection generally constructs a nonzero score.  Its mean
absolute propagated target shrinks from $0.2440$ to $0.0626$ between total
batch sizes 64 and 1,024, while the propagated hard gate activates in
$47$--$50\%$ of repetitions.  Thus correcting the temporal target does not
control uncertainty in the prior itself; a practical extension also requires
prior-aware abstention.

We therefore test an explicitly abstaining prototype.  We partition both
score-construction and gate-evaluation halves into four disjoint fold pairs,
form one propagated estimate per pair, and activate only when their mean
exceeds $2.353$ estimated standard errors (the one-sided 5\% Student-$t$
cutoff with three degrees of freedom).  At total batch sizes 64, 256, and
$1{,}024$, population-null activation is respectively $2.20\%$, $3.90\%$,
and $5.23\%$; clean population-decision correctness is $48.6\%$, $77.3\%$,
and $92.2\%$; and reversed-prior harmful activation is $0.33\%$, $0.15\%$,
and $0.20\%$.  
Thus independent-fold abstention gives null activation near the nominal level in this construction, with a substantial low-batch power cost.  The cutoff is
not claimed calibrated for dependent neural-PPO data or learned-critic error.
At batch $1{,}024$, a fixed multiplier sweep $0$, $1.638$, $2.353$, and
$3.182$ yields monotone null activation $49.28\%$, $10.13\%$, $5.23\%$, and
$2.45\%$, with clean decision correctness $97.55\%$, $95.13\%$, $92.15\%$,
and $88.40\%$.  Hence $2.353$ is a named operating point on the reported
power-versus-safety curve rather than a threshold optimized on these outcomes.

\paragraph{From a local coefficient to repeated PPO updates.}
We separately stress the boundary between the local theorem and a finite clipped update on $800$ random MDPs. With eight PPO epochs, learning rate $1.0$, clipping radius $0.2$, and a positive population local coefficient, the shaped path is harmful on $31.3\%$, $52.0\%$, and $69.0\%$ of the positive-coefficient MDPs at $\beta=0.1,0.5,2.0$. At $\beta=2$ and learning rate $0.3$, applying shaping only in the first epoch lowers this rate from $27.5\%$ to $4.34\%$, but also lowers the mean incremental gain from $0.0865$ to $0.0324$. An independent $800$-MDP replication preserves every preselected direction, including the advantage-normalization mismatch and the improvement from first-epoch-only shaping. These experiments locate the local-to-nonlocal boundary; they do not select a new Atari training schedule.

\paragraph{Decay under persistent prior error.}
In $120$ exact multi-state MDPs with a reversed prior and $\beta=2$, constant shaping converges to normalized progress $-0.8045$ (95\% interval $[-0.8669,-0.7420]$), whereas decay over $40$ updates reaches $0.9898$ and raw learning reaches approximately $0.9952$. A slower $100$-update decay reaches $0.9322$. Thus decay removes a persistent prior-dependent force and can allow recovery, but it does not rewind earlier updates and may retain finite-budget lag.

\paragraph{Interpretation.}
Together these experiments support a deliberately local account: the
operator-matched alignment is the first-order incremental coefficient of adding
OPR to the same raw update.  Bonus--advantage covariance recovers that
coefficient only for the stated pre-normalization idealization; for separately
normalized PPO it is a task-dependent proxy.  A valid radius can control
estimation error around the correct target, but cannot correct target mismatch
or certify a nonlocal optimization path.  Increasing $K$ trades action-relevant
aliasing against per-cluster estimation error, and pre-GAE placement propagates
both useful and harmful signals to earlier decisions.  These experiments map
when each mechanism helps and fails; they do not compare the complete training
path of a gated policy with the counterfactual path of an independently trained
PPO backbone.

\section{Experimental Details and Reproducibility}
\label{app:repro}

\subsection{Environment and preprocessing}
The Atari experiments use the Arcade Learning Environment through the Gymnasium
\texttt{ALE/*-v5} interface with sticky actions (repeat-action probability $0.25$),
grayscale $84\times84$ observations, and one deliberate-to-disclose deviation from the
standard protocol: an \emph{effective} frameskip of $16$, from the v5 built-in
frameskip of $4$ composed with a skip-$4$ max-pooling wrapper. This is identical across
the Atari runs reported here, so their action-repeat settings match; see also the
protocol note in the main paper's Setup. The stack further uses
a $4$-frame stack, up to $30$ no-op resets, and full action sets. Rewards are clipped to
$\{-1,0,+1\}$ for the policy/critic loss; \emph{reported scores and all
human-normalized scores use the raw (unclipped) game score}. The reward scale
$\rstd$ used by the bonus (Eq.~\ref{app:eq:bonus}) is the mean absolute (clipped) reward
of the current batch, floored at $10^{-4}$; in sparse-reward games this automatically
keeps the bonus proportional to the ambient reward density.

\subsection{Architecture}
The shared encoder is the $15$-layer IMPALA ResNet with channel widths
$[64,128,128]$ and a $512$-unit fully-connected head, feeding a linear policy head and a
scalar value head. Observations are scaled to $[0,1]$ before the encoder. The
state-conditioned prior (\S\ref{app:t8}) clusters frozen random-encoder features by
mini-batch $k$-means; the BC-shaping baseline trains a separate $3$-layer Nature-CNN
$\pi_{\mathrm{BC}}$ on good-episode states.

\subsection{Optimization and OPR hyperparameters}
Table~\ref{tab:hparams} lists all hyperparameters. Values are shared across all $56$
games (no per-game tuning). PPO settings follow the standard Atari recipe; the OPR block
lists the shaping-specific values, all held fixed except where a value is the subject of
an ablation.

\begin{table}[t]
\centering\small
\caption{Hyperparameters (shared across all games; no per-game tuning).}
\label{tab:hparams}
\begin{tabular}{@{}ll@{}}
\toprule
\multicolumn{2}{@{}l}{\emph{PPO / optimization}}\\
\midrule
Optimizer & Adam ($\epsilon=10^{-5}$) \\
Learning rate & $2.5\times10^{-4}$, linear decay \\
Discount $\gamma$ & $0.99$ \\
GAE $\lambda$ & $0.95$ \\
Clip $\epsilon$ & $0.1$; value clip on \\
Value / entropy coef. & $0.25$ / $0.01$ \\
Max grad norm & $0.5$; advantage norm.\ on \\
Parallel train / test envs & $8$ / $10$ \\
Minibatch size & $256$ \\
Agent steps per epoch & $100{,}000$; epochs $100$ \\
Total budget & $10$M agent steps ($160$M ALE frames) \\
Frame stack & $4$ \\
\midrule
\multicolumn{2}{@{}l}{\emph{OPR-specific}}\\
\midrule
Shaping scale $\ssc$ & $0.5$ \\
Good-episode threshold & $75$th percentile / batch \\
Buffer capacity & $1000$ episodes (FIFO) \\
Warmup & $20$ episodes \\
Laplace smoothing $\eta$ & $0.1$ \\
Prior source & top $\Kep{=}10$ episodes of $\Buf$ \\
Current-marginal EMA rate $\alpha$ & $0.01$ \\
Decay horizon $\Edec$ & $100$ epochs (linear, full run) \\
Bounded score & $(-2,2)$ intrinsically (Eq.~\ref{app:eq:sdelta}) \\
\bottomrule
\end{tabular}
\end{table}

\subsection{Online self-gated OPR}
\label{app:onlinegate}

The main table separates a retrospective checkpoint-oracle gate from online gates.
The online variants estimate alignment during training and multiply the bonus by a
scalar gate $g_t\in[g_{\min},1]$, so
that Eq.~\ref{app:eq:bonus} becomes $b_t=g_t\cdot\ssc\,\lambda_e\,\rstd\,\tilde\delta_t$.
With $g_t\equiv1$ the method is exactly un-gated OPR.

At each PPO batch we form the control-variate estimate of Proposition~\ref{app:prop:isest},
\begin{equation}\label{app:eq:onlinephi}
\hat\Align_{\mathrm{batch}}
=\operatorname{mean}_t\Big[\big(w_t-1\big)\,\hat A^{\mathrm{raw}}_t\Big],
\qquad
w_t=\mathrm{clip}\!\Big(\frac{\pgood(a_t)}{\pi(a_t\mid s_t)},0,10\Big),
\end{equation}
smooth it with an exponential moving average
$\hat\Align^{\mathrm{ema}}\!\leftarrow(1-\alpha_g)\hat\Align^{\mathrm{ema}}+\alpha_g\hat\Align_{\mathrm{batch}}$
($\alpha_g=0.05$), and set either a \emph{hard} (scale-free) or a \emph{soft} gate:
\[
g^{\mathrm{hard}}=\mathbb{1}\{\hat\Align^{\mathrm{ema}}>0\},
\qquad
g^{\mathrm{soft}}=\sigma\!\big(\hat\Align^{\mathrm{ema}}/\tau\big),\quad \tau=0.05 .
\]
The hard gate is a heuristic online analogue of the oracle rule
(App.~\ref{app:cor:gated}), but its zero threshold is not the selective rule of
Theorem~\ref{app:thm:selectivegate}. 
The soft gate smooths the discontinuity at $\Align=0$ but remains positive on negative estimates and has no corresponding sign-gate guarantee. The reported online results use hard gates.

For the covariance variant, replace $\hat\Align_{\mathrm{batch}}$ with $\operatorname{mean}_t[\sdratio_t\bar A_t^{\mathrm{raw}}]$, where $\bar A_t^{\mathrm{raw}}=(\hat A_t^{\mathrm{raw}}-\operatorname{mean}\hat A^{\mathrm{raw}})/(\operatorname{sd}(\hat A^{\mathrm{raw}})+10^{-8})$. The IS variant also uses these standardized raw advantages. Both use the same EMA and hard rule; finite-batch standardization makes these practical sample statistics, not automatically unbiased population coefficients.

\begin{figure}[t]
\centering
\includegraphics[width=\columnwidth]{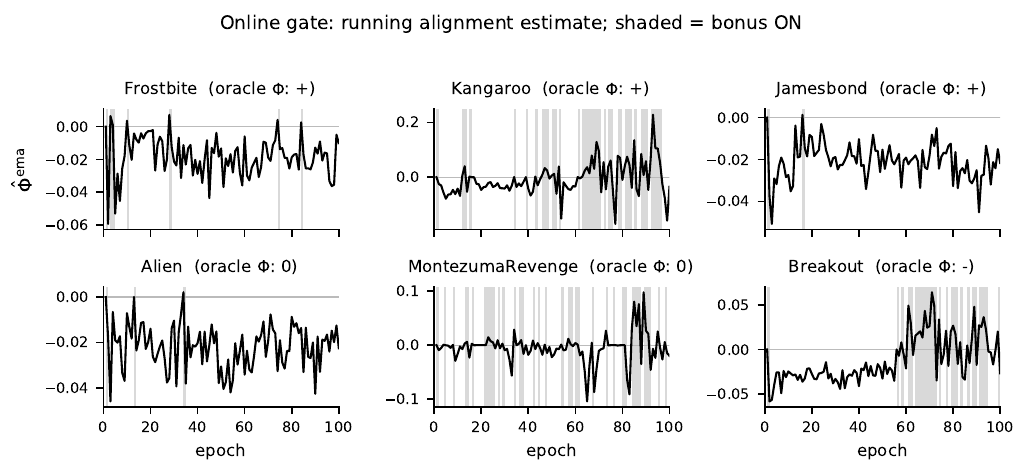}
\caption{Online (IS-estimator) hard gate over training on the six completed set runs:
running alignment estimate $\hat\Align^{\mathrm{ema}}$ (line; per-panel $y$-scales) with
bonus-active epochs shaded, oracle sign in each title. Latching (Kangaroo), starvation
(Frostbite, Jamesbond), and a late leak (Breakout) are the three regimes predicted by
the estimator's observed signal-to-noise behavior; the zero-threshold gate does not
satisfy the selective rule \eqref{app:eq:trust}.}
\label{fig:gatetraces}
\end{figure}

Figure~\ref{fig:gatetraces} shows the realized behavior of the deployed (IS-estimator)
hard gate on the completed set runs: the running estimate $\hat\Align^{\mathrm{ema}}$
across training with bonus-active spans shaded. Three regimes are visible, each
consistent with using a zero threshold without the valid radius required by
\eqref{app:eq:trust}. \emph{Latching} (Kangaroo):
$\hat\Align$ swings clearly positive mid-training and the gate opens repeatedly,
delivering shaping on an oracle-positive game. \emph{Starvation} (Frostbite, Jamesbond):
oracle-positive games whose online $|\hat\Align|\approx0.02$--$0.04$ remains small
relative to its observed fluctuations, so the estimate hovers noise-negative and the
upside is forfeited. \emph{Leak} (Breakout): a late noise
excursion holds the estimate positive on an oracle-negative game, applying harmful
shaping. This measured behavior is what motivates the covariance gate of
Remark~\ref{app:rem:covgate}, whose idealized bounded-summand range gives a
$\approx20\times$ smaller squared sample-size term and hence suggests higher
signal-to-noise. The factor is not a calibrated requirement for these neural traces.

\begin{figure}[t]
\centering
\includegraphics[width=\columnwidth]{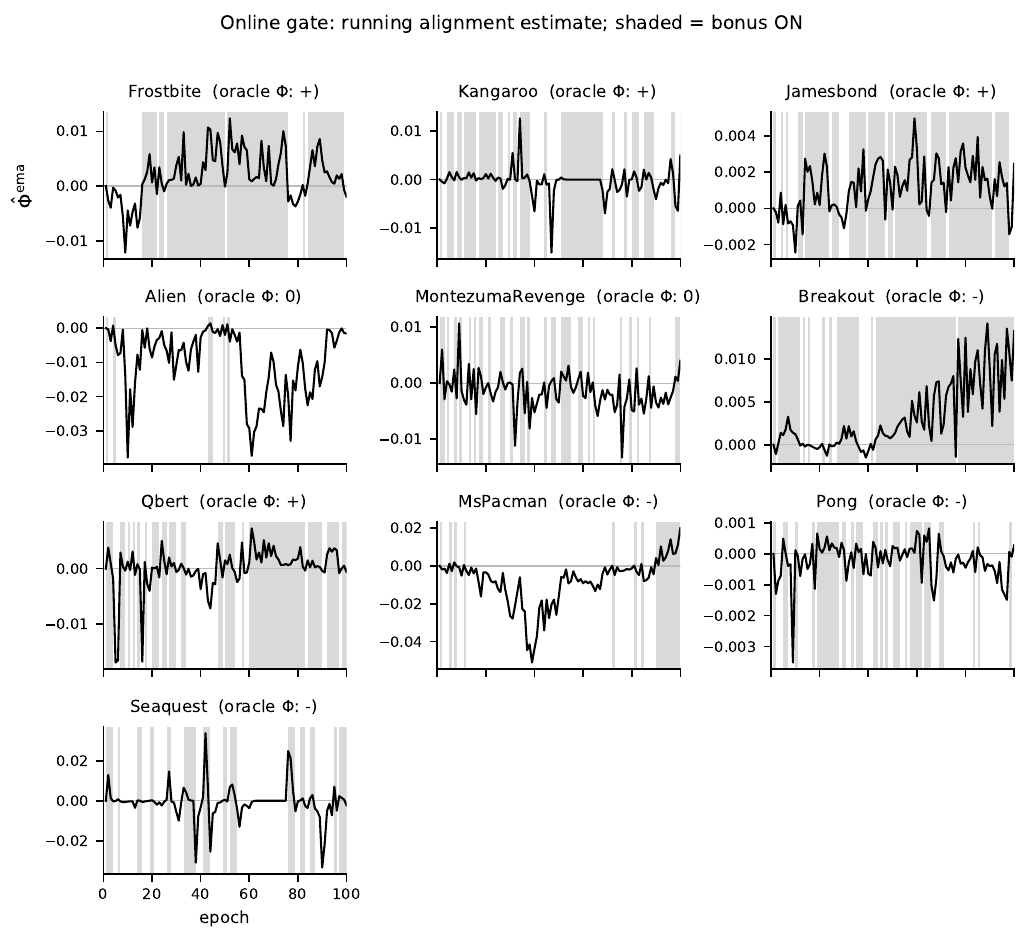}
\caption{The same trace for the \emph{covariance} gate (all ten games). Where the IS
gate starved, the covariance estimate latches: Frostbite opens and captures $655\%$
HNS (backbone $208$); on Breakout $\hat{\tilde\Align}$ stays positive against the
oracle-$\Align$ label and wins ($405$ vs $294$); Pong reaches $102.3$ versus $96.9$, while Kangaroo falls to $374.1$ versus $468.9$. Final medians: covariance gate $167.3\%$, IS gate $151.2\%$,
backbone $152.6\%$. This ordering is empirical; the selective theorem does not rank
the two full training runs.}
\label{fig:gatetracescov}
\end{figure}

Two details are not incidental. First, $\hat A^{\mathrm{raw}}$ is the GAE advantage
computed on the \emph{unshaped} reward. Using the shaped advantage would make the gate
self-confirming: a positive bonus inflates $\hat A$ on exactly the actions the bonus
favors, so $\hat\Align$ would be driven positive by the bonus's own presence rather than
by the environment reward. Second, the denominator of $w_t$ is the \emph{state-conditioned}
$\pi(a_t\mid s_t)$, not the action marginal $\pcurr(a_t)$. 
The state-conditioned denominator gives the population identity of Proposition~\ref{app:prop:isest} with exact advantages and unclipped weights. The implemented clipping, learned critic, batch standardization, and sample reuse can introduce bias. Substituting the marginal $\pcurr$ changes even the ideal population target. Online and offline estimates share the control-variate ratio form, but the online estimate additionally uses EMA and a changing policy/prior; they are not identical estimators. The archived implementation used for these historical traces fails open on non-finite estimates or exceptions ($g_t=1$), reverting to ungated OPR for that update; this fallback has no sign guarantee.

\begin{figure}[t]
\centering
\includegraphics[width=0.8\columnwidth]{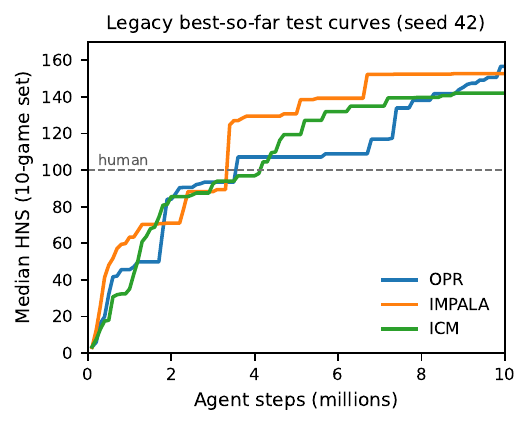}
\caption{Recorded test curves: median HNS over the 10-game set versus
agent steps (seed 42). The horizontal coordinate was previously mislabeled as raw
frames; $10$M agent steps equal $160$M raw ALE frames under our effective repeat 16.
At this seed the methods track early, IMPALA leads through most of the middle of the
run, and OPR overtakes only near the end; ICM finishes below both.
Table~\ref{tab:selected} shows that this ordering varies across seeds.}
\label{fig:curves}
\end{figure}

\subsection{Compute and evaluation}
Each $10$M-agent-step run takes $\approx\!20$ GPU-hours on a single accelerator; the full
study (main method, ablations, baselines, multiple seeds, $56$ games) was scheduled
across a small GPU pool. Runs used a single NVIDIA RTX PRO 6000 Blackwell (96GB) GPU
each, Python 3.11, PyTorch (CUDA 12.8 build), and Ubuntu 24.04 LTS. Test return is
evaluated over $10$ episodes at the end of each epoch; the tables report the original
recorded test results from the completed runs. The broad benchmark uses seed $42$;
the controlled $10$-game set uses seeds $\{42,43,44\}$.

\subsection{Reproducibility checklist highlights}
Code, per-game logs, and the exact command lines are released with the paper. All
hyperparameters are in Table~\ref{tab:hparams}; there is no per-game tuning. Randomness
is controlled by a single integer seed per run (environment, NumPy, and PyTorch).
Baselines quoted in the main table are taken from the cited papers at their original
frame budgets and are \emph{not} re-run in our framework; our same-backbone baselines
(IMPALA, ICM, SIL, BC-shaping) are run in our
framework with the identical encoder and budget.

\section{Complete Ablation Results}
\label{app:ablations}

Tables~\ref{tab:abl-beta} and~\ref{tab:abl-variants} give the full per-game raw scores
underlying the aggregate ablation numbers in the main text, regenerated directly from the
run logs. \textbf{Seed provenance:} the shaping-scale, decay, and $K_z{=}4$ conditions
(Table~\ref{tab:abl-beta}) were run at seed $42$; the $K_z{=}16$, return-quantile ($Q$),
ICM, and BC conditions (Table~\ref{tab:abl-variants}) are shown at seed $43$. Other available seeds, including seed $42$ for these conditions, are listed in App.~\ref{app:allresults}. The bottom row of each table is the median human-normalized score.
Two readings stand out. (i) The shaping scale is robust across $\ssc\in[0.1,0.5]$ and
degrades only for $\ssc\ge1$, consistent with the bonus overwhelming the task reward.
(ii) Both refinements of the marginal --- state-conditioning ($K_z$) and return-quantile
binning ($Q$) --- do not improve over the plain marginal in these single-seed settings.
The sample-complexity analysis (Theorem~\ref{app:thm:l1}) quantifies one plausible
variance cost but does not identify the return-optimal $K$ or establish that variance
caused this ordering. The learned BC prior likewise
underperforms the closed-form marginal. Paired Wilcoxon tests across the ten games find
\emph{none} of the variant-vs-main differences individually significant at $n{=}10$
(the BC prior comes closest, median $-22.7$ pp, $p{\approx}0.11$); the ablation
orderings should be read as consistent with the theory's predictions, not as
individually significant effects. Self-imitation baselines (corrected
implementation) are reported seed-matched in the main paper (Table~\ref{tab:selected}).
Given the seed sensitivity documented in Sec.~\ref{sec:controlled}, each single-seed row below
should be read as indicative rather than conclusive.

\begin{table*}[t]\centering\footnotesize\setlength{\tabcolsep}{4pt}
\caption{Per-game raw scores (10-game set, \textbf{seed 42}) for the shaping-scale, decay, and $K{=}4$ ablations. IMPALA is $\beta{=}0$. Bottom row: median HNS. Regenerated directly from run logs.}
\label{tab:abl-beta}
\begin{tabular}{@{}lrrrrrrrr@{}}
\toprule
Game & IMPALA & $\beta$.1 & $\beta$.25 & $\beta$.5 & $\beta$1 & $\beta$2 & no-dec & $K$4 \\
\midrule
Alien & 3{,}061 & 3{,}783 & 3{,}786 & 3{,}115 & 4{,}491 & 3{,}636 & 3{,}939 & 2{,}993 \\
Breakout & 86 & 122 & 107 & 82 & 138 & 25 & 95 & 29 \\
Frostbite & 8{,}963 & 22{,}879 & 20{,}044 & 24{,}926 & 9{,}137 & 10{,}505 & 9{,}433 & 8{,}636 \\
Jamesbond & 73{,}040 & 62{,}745 & 56{,}770 & 61{,}710 & 92{,}420 & 17{,}875 & 24{,}360 & 21{,}650 \\
Kangaroo & 14{,}040 & 14{,}740 & 11{,}570 & 14{,}470 & 11{,}220 & 14{,}400 & 11{,}240 & 14{,}240 \\
MontezumaRevenge & 0.0 & 0.0 & 80 & 150 & 0.0 & 0.0 & 0.0 & 0.0 \\
MsPacman & 5{,}022 & 5{,}825 & 5{,}313 & 4{,}257 & 5{,}469 & 8{,}079 & 5{,}002 & 6{,}999 \\
Pong & 14 & 7.2 & 10 & -1.7 & 8.4 & 7.6 & 11 & 12 \\
Qbert & 28{,}818 & 30{,}020 & 29{,}540 & 33{,}888 & 27{,}460 & 24{,}812 & 30{,}418 & 30{,}712 \\
Seaquest & 4{,}438 & 1{,}748 & 1{,}728 & 1{,}696 & 4{,}738 & 4{,}878 & 4{,}190 & 7{,}206 \\
\midrule
\emph{median HNS} & 153\% & 154\% & 154\% & 157\% & 144\% & 99\% & 155\% & 98\% \\
\bottomrule\end{tabular}\end{table*}

\begin{table*}[t]\centering\footnotesize\setlength{\tabcolsep}{4pt}
\caption{Per-game raw scores (10-game set, \textbf{seed 43}) for the $K{=}16$, return-quantile ($Q$), ICM, and BC-shaping conditions. Seed-42 results are also available in Table~\ref{tab:allagg}; numbers shown here are from seed-43 run logs. Bottom row: median HNS.}
\label{tab:abl-variants}
\begin{tabular}{@{}lrrrrrr@{}}
\toprule
Game & $K$16 & Q2 & Q4 & ICM & BC & BC-nd \\
\midrule
Alien & 3{,}079 & 3{,}530 & 3{,}318 & 3{,}424 & 2{,}863 & 4{,}282 \\
Breakout & 36 & 40 & 94 & 66 & 33 & 22 \\
Frostbite & 20{,}840 & 19{,}265 & 9{,}278 & 11{,}213 & 3{,}739 & 8{,}895 \\
Jamesbond & 21{,}110 & 28{,}140 & 78{,}945 & 26{,}025 & 82{,}470 & 27{,}200 \\
Kangaroo & 11{,}040 & 14{,}300 & 11{,}400 & 10{,}930 & 10{,}840 & 1{,}720 \\
MontezumaRevenge & 0.0 & 400 & 0.0 & 120 & 140 & 0.0 \\
MsPacman & 6{,}575 & 5{,}529 & 3{,}950 & 3{,}495 & 2{,}680 & 3{,}824 \\
Pong & 8.8 & 8.9 & 13 & 4.4 & -20 & -19 \\
Qbert & 5{,}520 & 32{,}765 & 35{,}038 & 28{,}290 & 9{,}908 & 26{,}982 \\
Seaquest & 5{,}954 & 4{,}154 & 4{,}540 & 1{,}658 & 1{,}648 & 1{,}534 \\
\midrule
\emph{median HNS} & 89\% & 108\% & 156\% & 141\% & 56\% & 57\% \\
\bottomrule\end{tabular}\end{table*}

\section{Predictor Measurement Details}
\label{app:predictor}

\subsection{Protocol}
For each set game we load the epoch-$50$ ($5$M-agent-step) checkpoint and roll out the
saved policy $
\pi$ (stochastic sampling) for $n_{\rm ep}$ episodes, reserving $K$ for state clusters. We form $\pgood$ from the
actions of the top-$25\%$ episodes by return and $\pcurr$ from all episodes, and estimate
the alignment by the control-variate importance-sampling estimator of
Proposition~\ref{app:prop:isest},
\[
\hat\Align=\operatorname{mean}_t[(w_t-1)\hat A_t],\qquad
w_t=\clip\!\left(\tfrac{\pgood(a_t)}{\pi(a_t\mid s_t)},0,10\right),
\]
where $\hat A_t$ is batch-standardized GAE from the saved value head on the \emph{raw} reward. The nominal budgets are 30 and 100 episodes; the saved Breakout records contain only 8 and 17 completed episodes, respectively. These clipped, same-sample estimates are retrospective diagnostics, not unbiased estimates under Proposition~\ref{app:prop:isest}.
No training or gradient steps are performed. Table~\ref{tab:predictor} reports the
estimate at $n_{\rm ep}=30$ and $n_{\rm ep}=100$ episodes together with $\KL(\pcurr\|\pgood)$ and the
realized OPR$-$IMPALA gap.

\subsection{Findings}
\begin{enumerate}[leftmargin=1.6em]
\item \textbf{Measurement point.} We precompute the diagnostic at epoch 50 to avoid
coupling its sign to the final reported evaluation. The raw-score reduced objective has
$\pcurr=\pgood$ as its unique optimum (Theorem~\ref{app:thm:fdiv}), but this does not
prove that bounded-score neural PPO converges there; epoch 50 is simply the disclosed
mid-training checkpoint used for this retrospective analysis.
\item \textbf{Sign stability.} On all seven games with a non-negligible effect
($|\Delta\mathrm{HNS}|>5$), $\operatorname{sign}(\hat\Align)$ matches
$\operatorname{sign}(\Delta\mathrm{HNS})$ at both $n_{\rm ep}=30$ and $n_{\rm ep}=100$; the two
``$\approx0$'' games (Alien, MontezumaRevenge) have $\Delta\mathrm{HNS}\approx0$ and an
uninformative sign. Over the nine non-outlier games, Spearman correlation is $+0.517$ at the nominal 30-episode budget and $+0.500$ at the nominal 100-episode budget. Breakout has the smaller realized counts stated above.
\item \textbf{Negative result (fitted-$Q$).} We also tried a fitted-$Q$ estimate of
$\Align$ that regresses Monte-Carlo returns on frozen encoder features per action
(giving counterfactual advantages for \emph{all} actions, not just the taken one). With
$512$-d features and a few hundred samples per action the per-action ridge regression is
too high-variance to be usable --- it returned physically implausible magnitudes
(hundreds of HNS-units) and unstable signs on \textsc{Qbert}/\textsc{MsPacman}. We
therefore report only the control-variate IS estimator, and note that a lower-variance
counterfactual-advantage estimator is the natural way to tighten the predictor.
\item \textbf{Limitations.} The validation is in-sample (the same ten games), with
$n=9$; the correlation is suggestive rather than significant at conventional levels. A
held-out set of games and a pre-registered sign/threshold criterion are required to
establish generalization beyond this retrospective association.
\end{enumerate}

\begin{table}[t]\centering\small
\caption{Predictor measurement. $\hat\Phi$ (clipped, batch-standardized control-variate IS estimate) and $\KL(\pcurr\|\pgood)$ at the epoch-50 (5M-agent-step) checkpoint, with nominal budgets of 30 and 100 episodes, versus the realized OPR$-$IMPALA gap ($\Delta$HNS, seed 42). Breakout records contain 8 and 17 episodes, respectively. Jamesbond is omitted from correlations as an HNS outlier.}
\label{tab:predictor}
\begin{tabular}{@{}lrrrr@{}}
\toprule
Game & $\hat\Phi_{30}$ & $\hat\Phi_{100}$ & KL$_{100}$ & $\Delta$HNS \\
\midrule
Frostbite & +0.062 & +0.113 & 0.0737 & +374 \\
Qbert & +0.011 & +0.004 & 0.0030 & +38 \\
Kangaroo & +0.039 & +0.025 & 0.0140 & +14 \\
MontezumaRevenge & -0.038 & -0.049 & 0.0000 & +3 \\
Alien & -0.085 & -0.098 & 0.0011 & +1 \\
Seaquest & -0.061 & -0.054 & 0.0002 & -6 \\
MsPacman & -0.005 & -0.009 & 0.0329 & -12 \\
Breakout & -0.013 & -0.007 & 0.0005 & -16 \\
Pong & -0.036 & -0.034 & 0.0052 & -43 \\
Jamesbond & +0.004 & +0.009 & 0.0003 & outl. \\
\bottomrule\end{tabular}\end{table}

\section{Complete Results Inventory}
\label{app:allresults}

This inventory reports the historical ten-game configurations and seeds used in the Atari analyses. Table~\ref{tab:allagg} aggregates a configuration only when all ten games are available; Table~\ref{tab:allpergame} also shows completed cells from partial configurations, with missing games marked \texttt{--}. Scores are taken from the recorded run results. The separate broad comparison is reported in Table~\ref{tab:appendix-full}.

\subsection{Aggregate results, every configuration and seed}

The backbone's seedwise medians span $123.0$--$152.6$. We therefore also compare each method's margin over the same-seed backbone; pairing matches the seed label and game, not divergent trajectories. ICM's two observed margins are similar ($-10.6/-10.9$ pp), whereas OPR's span $94$ pp ($+4.0/-43.2/+51.1$). Several elite-reward variants also vary substantially. This is consistent with a policy--elite-buffer--reward feedback effect, but these runs do not establish its cause (\S\ref{sec:controlled}).

\begin{table}[h]
\centering\small
\caption{Median HNS on the $10$-game set for every configuration and seed.
$\mathrm{sd}$ is the standard deviation across all available seeds (computed only where
$\ge\!2$ seeds are present; blank otherwise). Configurations are grouped by reward source or replay-loss mechanism. Seed-$42$ SIL runs with the incorrect inner-update count are excluded.}
\label{tab:allagg}
\begin{tabular}{@{}lrrrr@{}}
\toprule
Configuration & s42 & s43 & s44 & sd\\
\midrule
\multicolumn{5}{@{}l}{\emph{No reward bonus}}\\
PPO, Nature-CNN backbone (no shaping) & 56.4 & -- & -- & --\\
PPO, IMPALA backbone (no shaping) & 152.6 & 152.3 & 123.0 & \textbf{17.0}\\
\midrule
\multicolumn{5}{@{}l}{\emph{Bonus not from own elite episodes}}\\
ICM (curiosity shaping) & 142.0 & 141.4 & -- & 0.4\\
\midrule
\multicolumn{5}{@{}l}{\emph{Bonus from the agent's own elite episodes}}\\
OPR marginal, $\ssc{=}0.5$ (main) & 156.6 & 109.1 & 174.1 & \textbf{33.6}\\
OPR, $\ssc{=}0.1$ & 153.8 & -- & -- & --\\
OPR, $\ssc{=}0.25$ & 154.1 & -- & -- & --\\
OPR, $\ssc{=}1.0$ & 143.9 & -- & -- & --\\
OPR, $\ssc{=}2.0$ & 99.5 & -- & -- & --\\
OPR, no decay & 154.9 & -- & -- & --\\
OPR state-conditioned $K{=}4$ & 98.3 & 153.4 & -- & \textbf{39.0}\\
OPR state-conditioned $K{=}16$ & 52.0 & 89.0 & -- & \textbf{26.1}\\
OPR return-quantile $Q{=}2$ & 147.3 & 107.7 & -- & \textbf{28.0}\\
OPR return-quantile $Q{=}4$ & 148.0 & 156.2 & -- & \textbf{5.8}\\
BC-prior shaping & 76.3 & 55.8 & 57.2 & \textbf{11.5}\\
BC-prior shaping, no decay & 84.8 & 57.3 & 63.1 & \textbf{14.5}\\
\midrule
\multicolumn{5}{@{}l}{\emph{Self-imitation replay loss}}\\
marginal SIL & \emph{(excl.)} & 98.5 & 100.0 & 1.0\\
full SIL & \emph{(excl.)} & 64.6 & 63.5 & 0.8\\
\midrule
\multicolumn{5}{@{}l}{\emph{Gated variants (extension)}}\\
self-gated OPR (IS estimator, hard) & 151.2 & -- & -- & --\\
self-gated OPR (covariance est., hard) & 167.3 & 151.1 & 161.9 & \textbf{8.2}\\
\bottomrule
\end{tabular}
\end{table}

\paragraph{Encoder and shaping contributions.}
At seed $42$, PPO with Nature-CNN reaches $56.4\%$ median HNS, PPO with IMPALA ResNet reaches $152.6\%$, and OPR with the same ResNet reaches $156.6\%$. At this seed the encoder change accounts for about $96$ pp and shaping for about $4$ pp. Table~\ref{tab:selected} therefore holds the encoder fixed; Table~\ref{tab:broad} is a cross-paper comparison and does not isolate the shaping contribution.

\subsection{Per-game results, every configuration and seed}

Per-game human-normalized scores underlying the controlled ten-game aggregates. Complete rows supply a median; the partial $K=4$, seed-$44$ row does not.

\begin{table*}[t]
\centering\footnotesize
\setlength{\tabcolsep}{3.5pt}
\caption{Per-game human-normalized score (\%) for every completed run. Median is
over the ten games in that row.}
\label{tab:allpergame}
\resizebox{\textwidth}{!}{%
\begin{tabular}{@{}llrrrrrrrrrrr@{}}
\toprule
Configuration & Seed & Alien & Brkout & Frost & JBond & Kang & Monte & MsPac & Pong & Qbert & Seaq & \textbf{Med.}\\
\midrule
PPO, Nature-CNN backbone (no shaping) & 42 & 28.1 & 157.6 & 210.0 & 2745.1 & 58.6 & 9.0 & 54.1 & 51.3 & 166.5 & 3.9 & \textbf{56.4}\\
PPO, IMPALA backbone (no shaping) & 42 & 41.1 & 294.4 & 208.4 & 26665.8 & 468.9 & 0.0 & 71.0 & 96.9 & 215.6 & 10.4 & \textbf{152.6}\\
PPO, IMPALA backbone (no shaping) & 43 & 49.7 & 315.6 & 213.5 & 32115.0 & 491.7 & 0.0 & 41.2 & 91.2 & 257.2 & 3.7 & \textbf{152.3}\\
PPO, IMPALA backbone (no shaping) & 44 & 57.7 & 434.0 & 155.7 & 25365.6 & 374.4 & 0.0 & 79.9 & 90.4 & 222.9 & 6.8 & \textbf{123.0}\\
\midrule
ICM (curiosity shaping) & 42 & 50.8 & 226.7 & 216.5 & 20493.4 & 471.6 & 2.1 & 85.4 & 85.0 & 198.6 & 3.8 & \textbf{142.0}\\
ICM (curiosity shaping) & 43 & 46.3 & 224.7 & 261.1 & 9494.5 & 364.7 & 2.5 & 48.0 & 71.1 & 211.6 & 3.8 & \textbf{141.4}\\
\midrule
OPR marginal, $\ssc{=}0.5$ (main) & 42 & 41.8 & 278.8 & 582.3 & 22527.8 & 483.3 & 3.2 & 59.4 & 53.8 & 253.7 & 3.9 & \textbf{156.6}\\
OPR marginal, $\ssc{=}0.5$ (main) & 43 & 58.2 & 126.7 & 220.6 & 26711.5 & 427.7 & 0.0 & 61.1 & 91.5 & 232.8 & 10.4 & \textbf{109.1}\\
OPR marginal, $\ssc{=}0.5$ (main) & 44 & 51.9 & 420.8 & 566.1 & 19934.6 & 455.5 & 0.0 & 56.9 & 84.1 & 264.1 & 9.8 & \textbf{174.1}\\
OPR, $\ssc{=}0.1$ & 42 & 51.5 & 416.3 & 534.3 & 22905.8 & 492.4 & 0.0 & 83.0 & 79.0 & 224.6 & 4.0 & \textbf{153.8}\\
OPR, $\ssc{=}0.25$ & 42 & 51.6 & 367.0 & 467.9 & 20723.5 & 386.1 & 1.7 & 75.3 & 87.3 & 221.0 & 4.0 & \textbf{154.1}\\
OPR, $\ssc{=}1.0$ & 42 & 61.8 & 474.0 & 212.5 & 33744.0 & 374.4 & 0.0 & 77.7 & 82.4 & 205.4 & 11.1 & \textbf{143.9}\\
OPR, $\ssc{=}2.0$ & 42 & 49.4 & 81.9 & 244.5 & 6517.9 & 481.0 & 0.0 & 117.0 & 80.2 & 185.5 & 11.5 & \textbf{99.5}\\
OPR, no decay & 42 & 53.8 & 325.3 & 219.4 & 8886.4 & 375.1 & 0.0 & 70.7 & 90.4 & 227.6 & 9.8 & \textbf{154.9}\\
OPR state-conditioned $K{=}4$ & 42 & 40.1 & 95.8 & 200.7 & 7896.6 & 475.6 & 0.0 & 100.7 & 91.2 & 229.8 & 17.0 & \textbf{98.3}\\
OPR state-conditioned $K{=}4$ & 43 & 45.9 & 242.4 & 239.9 & 13787.8 & 414.6 & 0.0 & 45.2 & 85.3 & 221.6 & 11.5 & \textbf{153.4}\\
OPR state-conditioned $K{=}4$ & 44 & 54.2 & 144.8 & -- & 27396.3 & 440.1 & -- & 83.1 & 89.5 & 66.0 & 3.7 & \emph{(partial)}\\
OPR state-conditioned $K{=}16$ & 42 & 27.8 & 97.6 & 210.7 & 56528.9 & 373.7 & 0.0 & 38.9 & 65.2 & 33.9 & 3.9 & \textbf{52.0}\\
OPR state-conditioned $K{=}16$ & 43 & 41.3 & 120.8 & 486.6 & 7699.4 & 368.4 & 0.0 & 94.3 & 83.6 & 40.3 & 14.0 & \textbf{89.0}\\
OPR return-quantile $Q{=}2$ & 42 & 41.8 & 216.0 & 206.4 & 28729.4 & 372.4 & 0.0 & 55.4 & 88.1 & 281.1 & 3.9 & \textbf{147.3}\\
OPR return-quantile $Q{=}2$ & 43 & 47.9 & 131.6 & 449.7 & 10267.0 & 477.6 & 8.4 & 78.6 & 83.9 & 245.3 & 9.7 & \textbf{107.7}\\
OPR return-quantile $Q{=}4$ & 42 & 51.3 & 431.6 & 213.3 & 22980.6 & 420.7 & 0.0 & 69.2 & 82.7 & 220.5 & 10.8 & \textbf{148.0}\\
OPR return-quantile $Q{=}4$ & 43 & 44.8 & 320.5 & 215.8 & 28822.5 & 380.4 & 0.0 & 54.8 & 96.6 & 262.4 & 10.7 & \textbf{156.2}\\
BC-prior shaping & 42 & 25.8 & 119.8 & 172.9 & 13108.5 & 453.5 & 21.7 & 62.2 & 90.4 & 37.4 & 9.5 & \textbf{76.3}\\
BC-prior shaping & 43 & 38.2 & 108.0 & 86.0 & 30109.9 & 361.6 & 2.9 & 35.7 & 0.6 & 73.3 & 3.8 & \textbf{55.8}\\
BC-prior shaping & 44 & 37.8 & 153.8 & 201.9 & 24899.9 & 352.9 & 0.0 & 71.0 & 43.3 & 38.2 & 3.8 & \textbf{57.2}\\
BC-prior shaping, no decay & 42 & 22.7 & 87.5 & 175.6 & 7383.5 & 367.0 & 0.0 & 93.2 & 82.2 & 34.3 & 3.3 & \textbf{84.8}\\
BC-prior shaping, no decay & 43 & 58.8 & 72.2 & 206.8 & 9923.7 & 55.9 & 0.0 & 52.9 & 5.9 & 201.8 & 3.5 & \textbf{57.3}\\
BC-prior shaping, no decay & 44 & 23.5 & 120.5 & 96.5 & 12746.9 & 408.6 & 0.0 & 63.9 & 62.3 & 52.9 & 2.9 & \textbf{63.1}\\
marginal SIL & 43 & 24.8 & 124.0 & 177.0 & 11158.1 & 273.8 & 6.5 & 32.0 & 73.1 & 189.0 & 7.8 & \textbf{98.5}\\
marginal SIL & 44 & 32.3 & 142.7 & 164.4 & 8154.1 & 396.5 & 5.9 & 36.2 & 77.1 & 122.9 & 6.7 & \textbf{100.0}\\
full SIL & 43 & 20.8 & 137.8 & 204.8 & 6057.7 & 373.0 & 4.2 & 52.9 & 76.2 & 4.3 & 7.8 & \textbf{64.6}\\
full SIL & 44 & 22.7 & 208.7 & 120.0 & 4487.2 & 437.7 & 7.8 & 20.1 & 90.9 & 36.0 & 3.9 & \textbf{63.5}\\
\midrule
self-gated OPR (IS estimator, hard) & 42 & 48.7 & 221.9 & 212.5 & 20635.9 & 490.4 & 1.7 & 89.8 & 86.4 & 222.0 & 3.9 & \textbf{151.2}\\
self-gated OPR (covariance est., hard) & 42 & 57.1 & 404.9 & 655.3 & 29269.9 & 374.1 & 0.0 & 91.2 & 102.3 & 232.3 & 3.8 & \textbf{167.3}\\
self-gated OPR (covariance est., hard) & 43 & 50.7 & 282.6 & 213.0 & 23981.4 & 483.7 & 0.0 & 37.6 & 89.2 & 274.3 & 9.5 & \textbf{151.1}\\
self-gated OPR (covariance est., hard) & 44 & 57.7 & 369.1 & 1013.8 & 36176.4 & 408.9 & 0.0 & 41.9 & 83.6 & 240.2 & 3.8 & \textbf{161.9}\\
\bottomrule
\end{tabular}
}
\end{table*}

\section{CAGE-2 Cross-Domain Results}
\label{app:cage}

This section reports the full CAGE-2 cross-domain probe summarized in \S\ref{sec:cage}. CAGE-2 is a network-defense benchmark (CybORG) in which a Blue agent defends a small enterprise network against a Red attacker; the winning agent from the official challenge is a hand-engineered PPO policy with a $36$-action space, decoy priming, and forced start actions. We add OPR to this agent \emph{unmodified in every other respect}: identical action space, network architecture, optimizer, and all PPO hyperparameters, changing only the shaping bonus and its scale $\ssc$.

\paragraph{Scale selection.} We explored $\ssc\in\{0.1,0.25,0.5,1.0\}$ and a no-decay variant at $\ssc=0.5$ on B\_line (three seeds each). This is development on the CAGE-2 test domain, not an independent validation-then-test split. We report the entire official-protocol sweep in Table~\ref{tab:cage-sweep} and use $\ssc=0.1$ for the comparisons below. The sweep's reporting rule requires disclosure of all configurations.

\begin{table}[h]
\centering\small
\caption{CAGE-2 shaping-scale sweep against B\_line, official protocol (1000 episodes), mean episode return across three seeds. Higher (less negative) is better.}
\label{tab:cage-sweep}
\begin{tabular}{@{}lrrr@{}}
\toprule
Config & $30$-step & $50$-step & $100$-step \\
\midrule
Baseline (winner, unmodified) & $-4.10$ & $-7.19$ & $-15.31$ \\
OPR, $\ssc=0.1$          & $-3.97$ & $-7.04$ & $\mathbf{-14.80}$ \\
OPR, $\ssc=0.25$         & $-3.98$ & $-7.07$ & $-15.03$ \\
OPR, $\ssc=0.5$ (decay)  & $-3.94$ & $-7.12$ & $-15.26$ \\
OPR, $\ssc=1.0$          & $-4.11$ & $-7.43$ & $-15.59$ \\
OPR, $\ssc=0.5$ (no decay) & $-4.14$ & $-7.51$ & $-15.66$ \\
\bottomrule
\end{tabular}
\end{table}

The $100$-step scale ordering favors smaller bonuses, although the $30$-step scores are not monotone: a smaller, decayed bonus ($\ssc=0.1$) improves over the baseline at every horizon, while a large or non-decaying bonus ($\ssc=1.0$, no-decay) degrades it. $\ssc=0.1$ is the config we report at \S\ref{sec:cage} and use below.

\begin{table}[h]
\centering\small
\caption{CAGE-2 official-protocol evaluation (mean episode return, three seeds; higher is better) for the selected scale $\ssc=0.1$, against both official attackers.}
\label{tab:cage-main}
\begin{tabular}{@{}llrrr@{}}
\toprule
Attacker & Agent & $30$-step & $50$-step & $100$-step \\
\midrule
\multirow{3}{*}{B\_line} & Baseline        & $-4.10$ & $-7.19$ & $-15.31$ \\
                         & OPR (ungated)   & $-3.97$ & $-7.04$ & $\mathbf{-14.80}$ \\
                         & OPR (gated)     & $-3.93$ & $-7.20$ & $-15.53$ \\
\midrule
\multirow{3}{*}{Meander} & Baseline        & $-5.69$ & $-8.97$ & $-16.75$ \\
                         & OPR (ungated)   & $-5.56$ & $-8.80$ & $-16.92$ \\
                         & OPR (gated)     & $-5.54$ & $-8.81$ & $-17.29$ \\
\bottomrule
\end{tabular}
\end{table}

\paragraph{Reading the result.} Against B\_line, ungated OPR improves over the baseline at every horizon, most clearly at $100$ steps ($-14.80$ vs.\ $-15.31$), and the sign is consistent across all three seeds. Against Meander, 
the ungated OPR mean differs from the baseline by less than $0.2$ at each horizon, with inconsistent signs across seeds at $100$ steps. The reported means alone do not establish equivalence. This pattern is consistent with the alignment hypothesis under a small elite-prior signal, but two CAGE scenarios do not constitute an independent test of the checkpoint diagnostic.

\paragraph{Why we do not recommend the CAGE-2 gate.} Table~\ref{tab:cage-main} also reports the online covariance-gated variant at $\ssc=0.1$: it does not improve on ungated OPR, and at $100$ steps it is worse than both the baseline and ungated OPR on both attackers. Its measured statistic is only $|\hat{\tilde\Align}|\approx7\times10^{-4}$ to $1.3\times10^{-3}$ across seeds and attackers (Table~\ref{tab:cage-gate}), while the deployed zero-threshold rule has no valid error radius for these dependent neural data. The statistic therefore cannot certify its sign here. This negative result is evidence against using that gate on CAGE-2, not confirmation of a safety theorem or proof that all calibrated gates must fail.

\begin{table}[h]
\centering\small
\caption{Online covariance-gate diagnostics at $\ssc=0.1$: mean absolute alignment, fraction of updates with the gate open, and the sign-flip rate of the alignment estimate, each over three seeds.}
\label{tab:cage-gate}
\begin{tabular}{@{}lrrr@{}}
\toprule
Attacker & mean $|\hat{\tilde\Align}|$ & gate open & sign flips \\
\midrule
B\_line  & $0.00084$--$0.00132$ & $70$--$87\%$ & $4$--$7\%$ \\
Meander  & $0.00077$--$0.00108$ & $84$--$90\%$ & $7$--$8\%$ \\
\bottomrule
\end{tabular}
\end{table}

\paragraph{Scope.} This experiment shows that the ungated reward-channel mechanism can
be attached, with only a shaping-scale change, to a hand-engineered agent in an
adversarial cyber-defense domain unlike Atari. It does not validate the selective-gate
assumptions: the online gate is negative here and has no calibrated error radius. This
is one benchmark family at one selected scale, not a second full benchmark comparable
to the $49$-game Atari evaluation.

\section{Crafter Transfer Study}
\label{app:crafter-transfer}

\paragraph{Held-out transfer result.}
As a held-out transfer test, we evaluate a frozen OPR configuration on
Crafter, an open-world survival environment with 22 semantically distinct
achievements.  Following
the benchmark protocol, each run receives one million environment interactions
and its primary score is $\exp[22^{-1}\sum_{j=1}^{22}\log(1+p_j)]-1$, where $p_j$ is the success percentage for achievement $j$ over all completed training episodes ending within the budget.  We use ten paired seeds and report every
prespecified cell.  Our PPO baseline obtains $4.775\pm0.282$, closely
reproducing the original Crafter PPO baseline of $4.6\pm0.3$~\citep{hafner2022crafter}.  Ungated OPR obtains
$4.873\pm0.157$ with the marginal $K{=}1$ prior (paired difference $+0.097$,
95\% bootstrap CI $[-0.129,+0.325]$, 6/10 wins) and $4.878\pm0.223$ with a
privileged four-stage progress label ($+0.103$ $[-0.072,+0.284]$, 7/10 wins).
The direct privileged-$K{=}4$ minus $K{=}1$ difference is $+0.006$
$[-0.116,+0.151]$, so extra oracle structure has no observed mean advantage,
although this interval does not establish equivalence.  The gated variants
show similarly unresolved differences (Table~\ref{tab:crafter-transfer}).
These data support a modest transfer trend and exercise the $K{=}1$ design in a
new domain; they do not establish a statistically resolved Crafter improvement.

\paragraph{Protocol and complete result.}
We use Crafter 1.8.3 with a Nature-CNN PPO agent, eight parallel environments,
256-step rollouts (batch size 2,048), ten PPO epochs, learning rate
$3\times10^{-4}$, $\gamma=0.99$, GAE $\lambda=0.95$, clipping parameter 0.2,
and no entropy bonus.  Each of ten seeds (300--309) is trained for one million
interactions.  The transfer configuration constructs its prior from the top
100 episodes in the 1,000-episode elite buffer and scales the bonus by the
current rollout's mean absolute sparse reward.  These choices differ from the
Atari top-10 setting in the prior's episode count; both use mean absolute reward scaling. We describe the top-100 configuration as a domain adaptation rather than OPR ``unmodified.''  The
bounded ratio, elite buffer, linear decay, and pre-GAE placement are retained.
The $K{=}4$ arm uses inventory and cumulative-achievement progress stages and
is explicitly privileged.  It reads simulator internals unavailable in the
standard pixel observation and is therefore a mechanism diagnostic, not a
standard Crafter leaderboard arm.  The configuration was fixed 
before the primary transfer outcomes were observed; within that frozen study no outcome was used to select a variant, seed, or checkpoint. The separately labeled follow-ups below were designed later.

\begin{table}[h]
\centering
\small
\setlength{\tabcolsep}{2.8pt}
\caption{Frozen Crafter transfer at one million interactions. Scores are
mean $\pm$ s.d. over ten training seeds.  Differences, confidence intervals,
and wins are paired by seed against PPO.  Every interval crosses zero.}
\label{tab:crafter-transfer}
\begin{tabular}{@{}lrrr@{}}
\toprule
Method & Score & Paired $\Delta$ [95\% CI] & Wins \\
\midrule
PPO & $4.775\pm0.282$ & -- & -- \\
OPR, $K{=}1$ & $4.873\pm0.157$ & $+0.097$ [$-0.129,+0.325$] & 6/10 \\
OPR, privileged $K{=}4$ & $4.878\pm0.223$ & $+0.103$ [$-0.072,+0.284$] & 7/10 \\
gated OPR, $K{=}1$ & $4.806\pm0.340$ & $+0.031$ [$-0.229,+0.317$] & 5/10 \\
gated OPR, privileged $K{=}4$ & $4.872\pm0.244$ & $+0.096$ [$-0.107,+0.305$] & 6/10 \\
\bottomrule
\end{tabular}
\end{table}

\begin{figure}[h]
\centering
\includegraphics[width=\columnwidth]{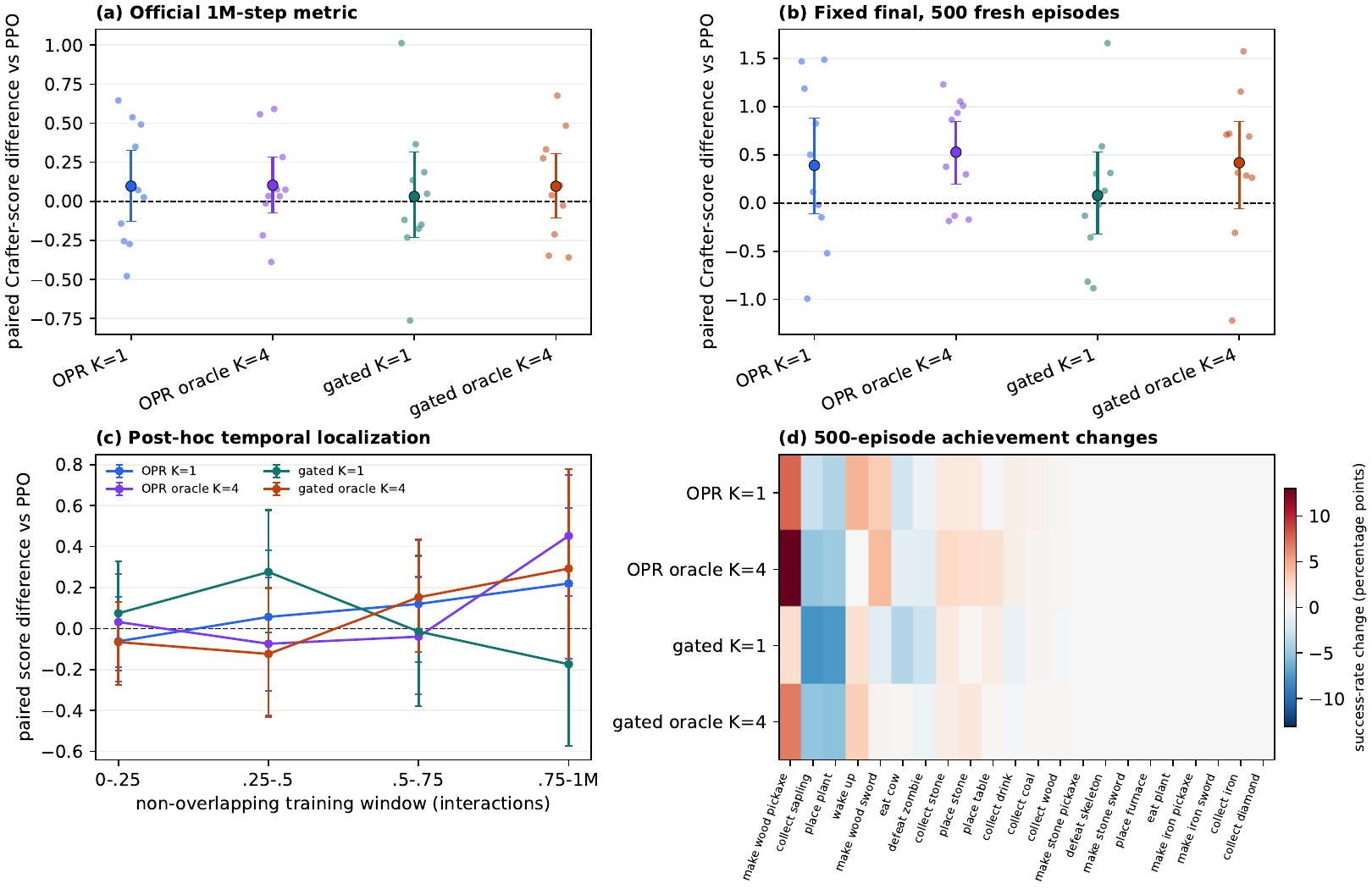}
\caption{Complete paired Crafter evidence. Points are the ten seedwise
differences from PPO; large points and bars are paired means and 95\% bootstrap
intervals.  Panel (a) is the official all-training-episodes score.  Panel (b)
uses the frozen 500-episode evaluation of every fixed final checkpoint.  Panel
(c) is a post-hoc reconstruction over non-overlapping training quarters; its
pointwise intervals are diagnostic, not multiplicity-corrected tests.  Panel
(d) reports every achievement-rate difference in the fixed-final endpoint,
ordered by the largest absolute change.}
\label{fig:crafter-transfer}
\end{figure}

Across the cumulative training metric, ungated $K{=}1$ OPR increases 13 of 22
mean achievement rates, decreases six, and leaves three unchanged or
unreached.  Its largest gain is on making a wood pickaxe.  This suggests that
the small aggregate change is distributed across several early crafting and
survival behaviors rather than one anomalous achievement, but no individual
achievement is treated as a separately powered hypothesis.  The gate is open
on 90.6\% of active $K{=}1$ updates and 82.0\% of active $K{=}4$ updates; this
experiment therefore supplies no evidence of a strong gate advantage.  The
controlled synthetic study remains the direct test of gate calibration.

The official score pools the entire learning history.  Consistently, the paired
$K{=}1$ difference grows from $-0.068$ at 0.25M interactions to $+0.097$ at
1M; the experiment does not support an early sample-efficiency claim.  A
fixed-final evaluation asks a complementary question about the learned policy
after the standard budget and can expose a late difference that the cumulative
metric dilutes.  The bonus has decayed to zero by the final update, but this
does not test persistence under additional unshaped training beyond 1M steps.
Consistently, a post-hoc reconstruction over non-overlapping training quarters
finds final-quarter differences of $+0.220$ $[-0.151,+0.590]$ for $K{=}1$ and
$+0.452$ $[+0.158,+0.749]$ for privileged $K{=}4$; earlier $K{=}4$ quarters
are near zero.  These pointwise, non-prespecified window estimates localize the
effect in time and are not additional primary tests.

\paragraph{Precision follow-up disclosure.}
After observing the primary study and its noisy 50-episode endpoint, but before
inspecting any 500-episode outcomes, we froze a 500-episode evaluation of every
fixed final checkpoint on fresh environment and policy-randomness seeds.
This tenfold increase over the original 50-episode endpoint reduces evaluation
noise without checkpoint or variant selection.  It remains secondary because
the official Crafter metric aggregates all training episodes.  All 50
evaluations pass checkpoint and protocol validation.  OPR obtains
$5.554\pm0.772$ at $K{=}1$ (paired difference $+0.390$,
$[-0.110,+0.886]$, 6/10 wins) and $5.691\pm0.442$ with privileged $K{=}4$
($+0.527$ $[+0.197,+0.847]$, 7/10 wins, exact sign-flip $p=0.025$), versus
$5.164\pm0.485$ for PPO.  The direct privileged-$K{=}4$ minus $K{=}1$
contrast remains unresolved ($+0.137$ $[-0.331,+0.579]$), so the result does
not show that privileged labels outperform the marginal prior.  Gated $K{=}1$
and privileged $K{=}4$ score $5.242\pm0.635$ and $5.582\pm0.558$, respectively;
their paired intervals against PPO cross zero.  Thus the higher-precision
endpoint strengthens the OPR transfer evidence, while its only pointwise
resolved variant-versus-PPO contrast uses privileged labels, is not the primary
all-training-episodes benchmark metric, and does not survive a four-comparison
Holm or Bonferroni family-wise correction.  The non-privileged $K{=}1$ arm is
the relevant standard-interface result and remains a positive but unresolved
trend.  Because the precision study was motivated after the original endpoint,
we treat it as a transparently designed follow-up rather than independent
confirmation.

Separately, a top-10 sensitivity (retaining Crafter's reward-scale adaptation)
was frozen after observing the top-100 results and is therefore a disclosed
development analysis rather than held-out confirmation; both settings will be
reported regardless of outcome.  All 20 runs reach the full budget.  Ungated
top-10 $K{=}1$ obtains $4.843\pm0.316$ (paired difference from PPO $+0.067$,
$[-0.098,+0.253]$) and differs from top-100 by $-0.030$
$[-0.257,+0.179]$.  Gated top-10 obtains $4.792\pm0.203$, differing from PPO
by $+0.017$ $[-0.245,+0.269]$ and from gated top-100 by $-0.015$
$[-0.192,+0.147]$.  Thus the modest trend is not visibly dependent on using
100 rather than 10 elite episodes, but neither setting resolves the $K{=}1$
contrast against PPO and the intervals do not establish equivalence.

The frozen 500-episode top-10 endpoint gives $5.180\pm0.544$ for ungated OPR
(difference from PPO $+0.016$ $[-0.374,+0.469]$, 4/10 wins) and
$5.491\pm0.413$ for gated OPR ($+0.327$ $[+0.057,+0.616]$, 7/10 wins).
The latter bootstrap interval excludes zero, but its exact sign-flip test gives
$p=0.0625$.  Directly, gated minus ungated is $+0.311$
$[-0.032,+0.643]$ (7/10 wins, $p=0.125$), and the official cumulative metric
shows no gated advantage.  We therefore treat this as suggestive evidence that
gating may help a smaller, noisier elite prior, not a confirmed improvement.

\begin{figure}[h]
\centering
\includegraphics[width=0.88\columnwidth]{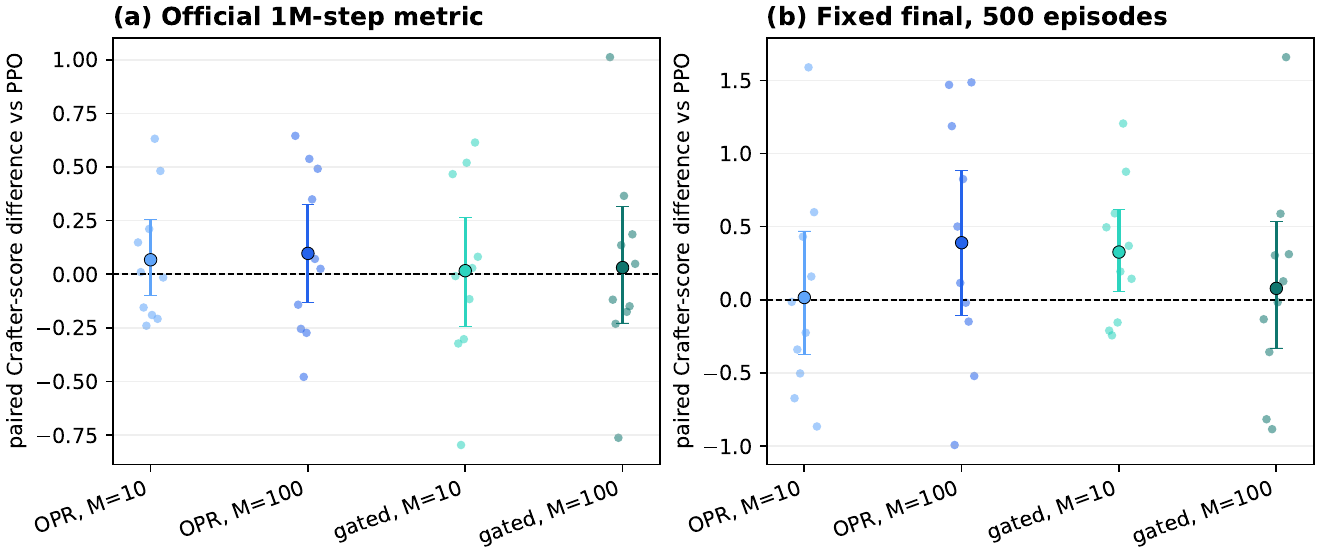}
\caption{Disclosed elite-set sensitivity with all ten paired seeds.
$M$ is the number of top buffered episodes used to estimate the marginal prior.
Bars are pointwise 95\% paired-bootstrap intervals.  For gated $M{=}10$ in
panel (b), that interval excludes zero but the exact sign-flip test gives
$p=0.0625$; we do not label it significant.}
\label{fig:crafter-topk}
\end{figure}


\end{document}